\documentclass[10pt,twocolumn,letterpaper]{article}

\usepackage[pagenumbers]{iccv} 

\definecolor{iccvblue}{rgb}{0.21,0.49,0.74}
\usepackage[pagebackref,breaklinks,colorlinks,allcolors=iccvblue]{hyperref}
\newcommand\blfootnote[1]{%
  \begingroup
  \renewcommand\thefootnote{}\footnote{#1}%
  \addtocounter{footnote}{-1}%
  \endgroup
}
\usepackage{multirow}
\usepackage{multicol}
\usepackage{dirtytalk}
\usepackage{breqn}
\usepackage{pifont}
\newcommand{\cmark}{\ding{51}}%

\def\paperID{*****} 
\def\confName{ICCV}
\def\confYear{2025}

\title{TransGUNet: Transformer Meets Graph-based Skip Connection for Medical Image Segmentation}

\author{Ju-Hyeon Nam \qquad Nur Suriza Syazwany \qquad Sang-Chul Lee$^{*}$ \\
Department of Electrical and Computer Engineering, Inha University\\
100, Inha-ro, Michuhol-gu, Incheon, Republic of Korea\\
{\tt\small \{jhnam0514, surizasyazwany, sclee\}@inha.edu}
}

\begin{document}
\maketitle
\begin{abstract}
Skip connection engineering is primarily employed to address the semantic gap between the encoder and decoder, while also integrating global dependencies to understand the relationships among complex anatomical structures in medical image segmentation. Although several models have proposed transformer-based approaches to incorporate global dependencies within skip connections, they often face limitations in capturing detailed local features with high computational complexity. In contrast, graph neural networks (GNNs) exploit graph structures to effectively capture local and global features. Leveraging these properties, we introduce an attentional cross-scale graph neural network (ACS-GNN), which enhances the skip connection framework by converting cross-scale feature maps into a graph structure and capturing complex anatomical structures through node attention. Additionally, we observed that deep learning models often produce uninformative feature maps, which degrades the quality of spatial attention maps. To address this problem, we integrated entropy-driven feature selection (EFS) with spatial attention, calculating an entropy score for each channel and filtering out high-entropy feature maps. Our innovative framework, \textbf{TransGUNet}, comprises \textit{ACS-GNN} and \textit{EFS-based spatial attention} to effectively enhance domain generalizability across various modalities by leveraging GNNs alongside a reliable spatial attention map, ensuring more robust features within the skip connection. Through comprehensive experiments and analysis, TransGUNet achieved superior segmentation performance on six seen and eight unseen datasets, demonstrating significantly higher efficiency compared to previous methods.
\end{abstract}   
\blfootnote{$*$ denotes the corresponding author.}
\section{Introduction}
\label{sec:intro}

\begin{figure}[t]
    \centering
    \includegraphics[width=0.5\textwidth]{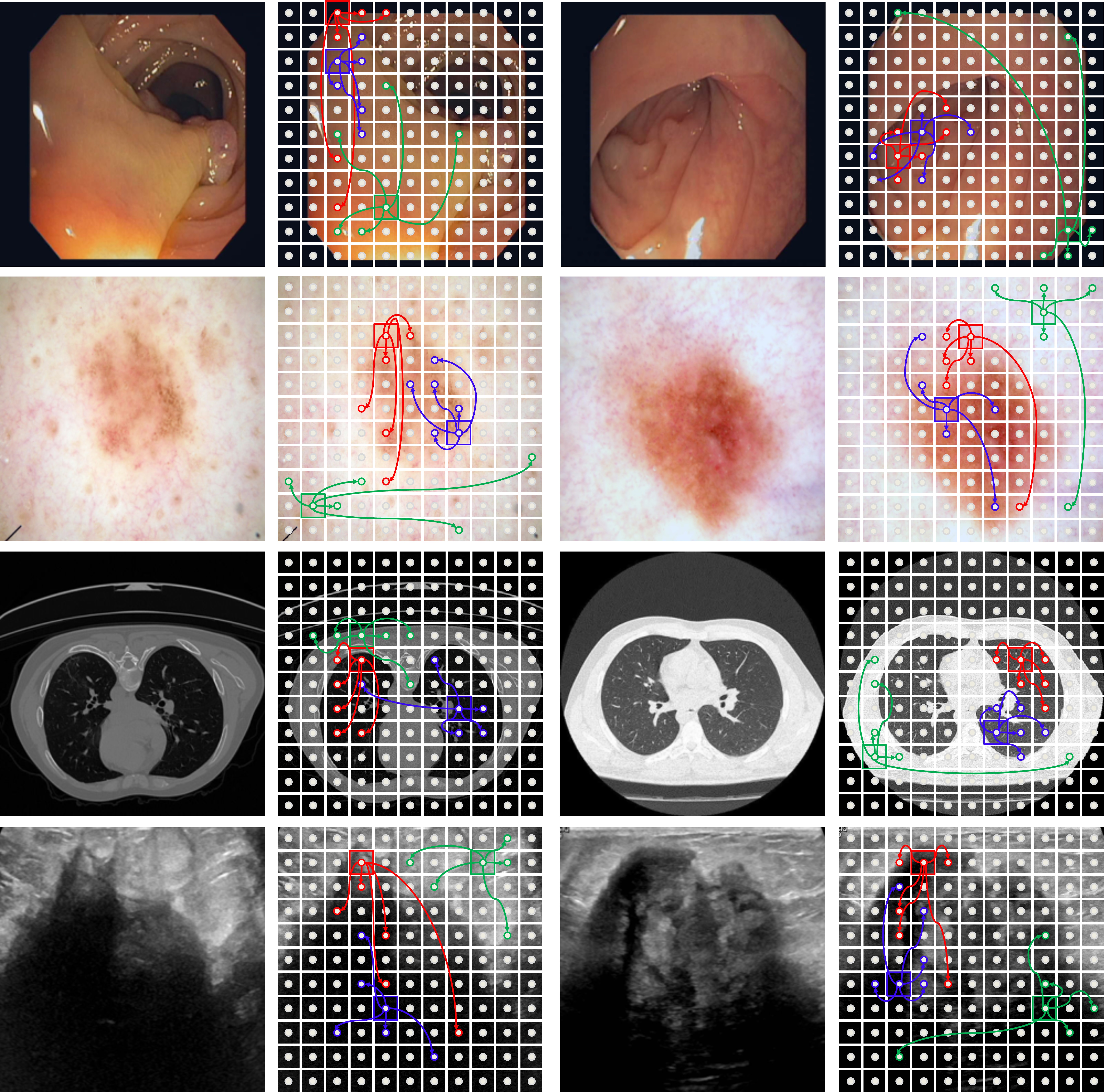}
        \caption{Graph Visualization of TransGUNet. We selected three patches (\textcolor{red}{\textbf{Red}}, \textcolor{blue}{\textbf{Blue}}, \textcolor{ForestGreen}{\textbf{Green}}) and found the five nearest patches for each, based on the adjacency matrix of ACS-GNN, connecting them with lines to visualize the relationships using each color. This figure reveals that lesion patches exhibit high similarity with other lesion patches, while non-lesion patches similarly cluster together. This result demonstrates the model’s effectiveness in distinguishing lesion from non-lesion regions and maintaining strong intra-class correlations.}
    \label{fig:GraphVisualization}
\end{figure}

Medical image segmentation is crucial for the early detection of abnormal tissues and the development of treatment plans \cite{coates2015tailoring}. Traditional segmentation algorithms have received  considerable attention from medical experts \cite{otsu1979threshold, haralick1987image, kass1988snakes, tizhoosh2005image}. However, these methods still lack generalizability owing to the severe noise, inhomogeneous intensity distribution, and various clinical settings in medical images \cite{riccio2018new}. Consequently, this issue has raised concerns about the reliability of computer-based diagnostic procedures \cite{gaube2021ai}.

Recently, convolutional neural networks (CNNs) have been widely employed for medical image segmentation owing to their robustness in capturing local and spatial hierarchical features \cite{ronneberger2015u, zhao2021automatic, nam2023m3fpolypsegnet}. However, CNN-based models struggle to capture the global dependencies necessary to understand the complex anatomical structures in medical images \cite{hatamizadeh2022unetr}. This limitation has expanded the use of transformers to extract global dependencies for medical image segmentation \cite{chen2021transunet, cao2022swin}. However, despite their strengths, transformer-based models often fail to bridge the semantic gap between the encoder and decoder, hindering their ability to fully leverage global dependencies and resulting in sub-optimal segmentation performance \cite{wang2022uctransnet}.

Several models have been actively employed to improve skip connections to reduce this semantic gap for medical image segmentation. The most representative of these attempts is UNet++ \cite{zhou2018unet++}, which uses cross-scale feature fusion through dense connectivity in skip connection. Similarly, UCTransNet \cite{wang2022uctransnet}  and CFATransUNet \cite{wang2024cfatransunet} adopted a transformer-based approach to capture local cross-channel interactions of feature maps from a channel-wise perspective. However, these models suffer from increased computational complexity and ambiguous attention owing to their dense connectivity, extensive use of transformer blocks, and complex background in medical images with severe noise. Consequently, addressing the question, \say{\textit{How can we efficiently leverage global dependency without ambiguity while reducing the semantic gap between the encoder and decoder?}} is critical to overcoming these challenges and improving performance for medical image segmentation.

To answer this question, we focused on graph neural networks (GNNs), which are particularly suitable for flexibly and effectively capturing local and global dependencies, making them ideal for complex visual perception tasks \cite{han2022vision}. By leveraging this capability, we propose an \textit{attentional cross-scale GNN (ACS-GNN)} that can efficiently reduce the semantic gap between the encoder and decoder. It transforms cross-scale feature maps into graphs and applies attention to each node to facilitate robust feature integration. Additionally, we observed that deep learning models often produce uninformative feature maps that degrade the quality of spatial attention maps \cite{chen2021lesion, shawn2024ct}. To address this issue, we introduce an \textit{entropy-driven feature selection (EFS)}, which calculates entropy per channel and filters out high-entropy channels. By integrating \textit{ACS-GNN} and \textit{EFS-based spatial attention}, we designed a new medical image segmentation model called \textbf{TransGUNet} which effectively captures the relationships between patches, regardless of the lesion size and distance between patches (Fig. \ref{fig:GraphVisualization}). Extensive experimental results demonstrate that our graph-based approach consistently outperforms transformer- and convolution-based methods. Thus, TransGUNet represents a significant advancement in skip connection frameworks for medical image segmentation and offers a robust and efficient solution to the existing challenges. The main contributions of this study can be summarized as follows:

\begin{itemize}
    \item We propose \textbf{TransGUNet}, a novel medical image segmentation model that leverages the cross-scale GNN-based skip connection framework without ambiguous spatial attention and is applicable to various modalities and clinical settings. To the best of our knowledge, our novel skip connection framework is the first study to successfully and effectively exploit attentional cross-scale GNN with non-ambiguous spatial attention for medical image segmentation.

    \item The proposed \textit{attentional cross-scale graph neural network (ACS-GNN)} allows the model to comprehend the complex anatomical structures within medical images. Additionally, we incorporated \textit{entropy-driven feature selection (EFS)} with spatial attention to produce more reliable spatial attention maps.

    \item Our experimental results demonstrate that TransGUNet significantly outperforms transformer- and convolution-based approaches employed for medical image segmentation with various clinical settings.
\end{itemize}

\begin{table*}[t]
    \centering
    \footnotesize
    \setlength\tabcolsep{3.5pt} 
    \begin{tabular}{c|c|ccccccc|c}
    \multicolumn{2}{c|}{\multirow{2}{*}{Skip Connection Properties}}                       & UNet   & UNet++ & M2SNet & ViGUNet & CFATUNet & PVT-GCAS & GSENet & \textbf{TransGUNet} \\
    \multicolumn{2}{c|}{ } & \scriptsize{(MICCAI2016)} & \scriptsize{(DLMIA2018)} & \scriptsize{(arxiv2023)} & \scriptsize{(ISBI2023)} & \scriptsize{(CBM2024)} & \scriptsize{(WACV2024)} & \scriptsize{(BSPC2025)} & \scriptsize{\textbf{(Ours)}} \\
    \hline
    \multicolumn{2}{c|}{Full Global Dependency}  & -      & -      & -      & \cmark   & \cmark   & -       & \cmark & \cmark \\ 
    \multicolumn{2}{c|}{Cross-Scale Fusion}      & -      & \cmark & \cmark & -        & \cmark   & -       & \cmark & \cmark \\ 
    \multicolumn{2}{c|}{Non-Ambiguous Spatial Attention} & -      & -      & -      & -      & -      & -        & -        & \cmark \\ 
    \hline
    \multicolumn{1}{c|}{\multirow{4}{*}{Efficiency}} & Params        (M) & 8.2  & 34.9  & 25.3 & 2.3 & 64.6  & 25.4 & 26.9 & 25.0 \\
                                                     & FLOPs         (G) & 23.7 & 197.8 & 12.8 & 5.0 & 32.9  & 7.9 & 17.4 & 10.0 \\
                                                     & Inference Speed    (ms) & 18.6    & 26.5     & 34.9    & 25.1     & 36.0    & 17.4     & 37.5    & 19.4    \\
                                                     & Required GPU Memory    (G) & 0.4    & 1.2     & 0.6    & 3.1     & 1.0    & 0.4     & 0.5  & 0.4    \\
    \hline
    \end{tabular} 
    \caption{Comparison of skip  connection frameworks characteristics among UNet, UNet++, M2SNet, ViGUNet, CFATransUNet (CFATUNet) , PVT-GCASCADE (PVT-GCAS), GSENet, and \textbf{TransGUNet (Ours)}.}
    \label{tab:SkipConnectionEngineeringProperties}
\end{table*}

\begin{figure*}[t]
    \centering
    \includegraphics[width=\textwidth]{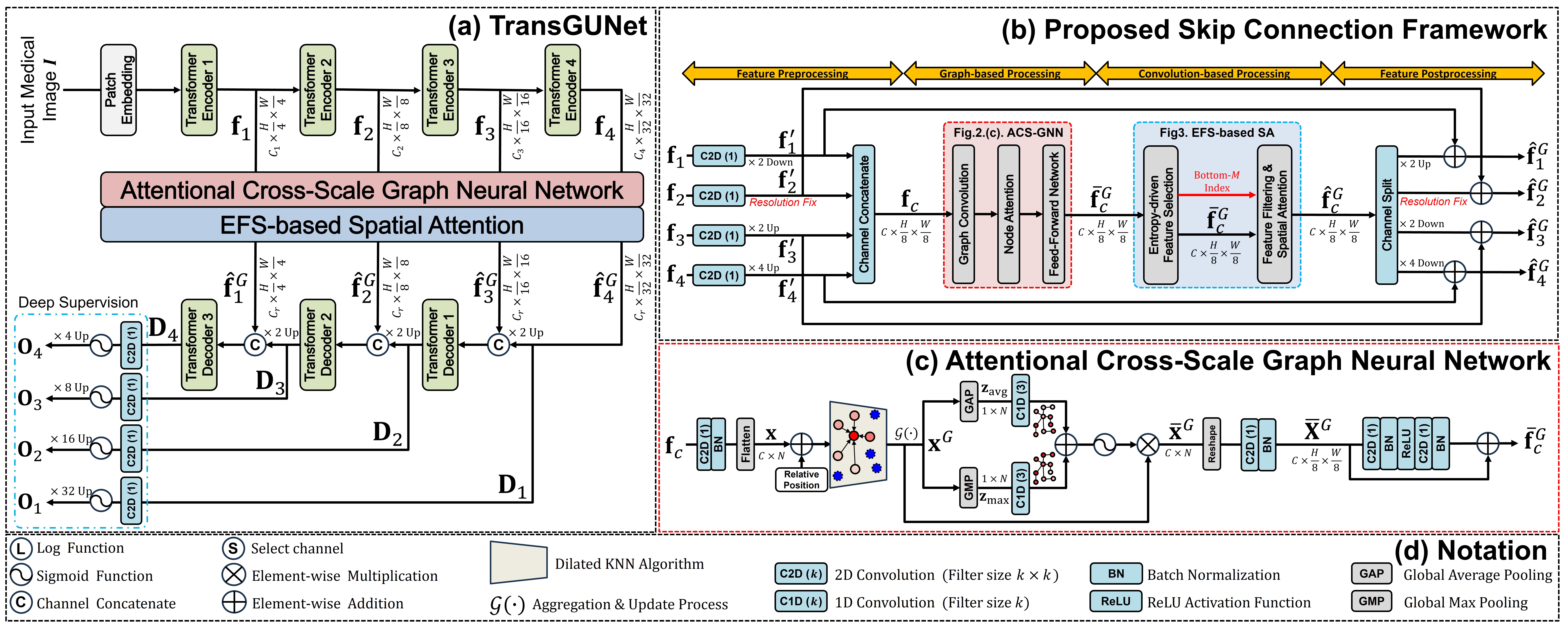}
    \caption{(a) The overall architecture of the proposed TransGUNet mainly comprises ACS-GNN and EFS-based spatial attention (See Fig. \ref{fig:ESA_based_spatial_attention}). (b) The proposed novel skip connection framework. In this figure, we set the target resolution as $(H_{t}, W_{t}) = (\frac{H}{8}, \frac{W}{8})$. And, for convenience, we assume that $C = 4C_{r}$. (c) Overall block diagram of the proposed ACS-GNN. (d) Notation description used in this paper. This notation is also used in Fig. \ref{fig:ESA_based_spatial_attention}.}
    \label{fig:TransGUNet}
\end{figure*}

\section{Related Works}

\noindent \textbf{Skip Connection Engineering for Medical Image Segmentation.} The introduction of skip connections in UNet marked a significant milestone and made it the most widely used baseline model in medical image segmentation. However, there is still a semantic gap between the encoder and decoder, which results in suboptimal performance \cite{mahmud2021covsegnet}. This problem has driven recent efforts to refine skip connections to minimize this semantic gap. UNet++ is a representative model incorporating dense connectivity and neighbor scale features in skip connections. Additionally, MSNet \cite{zhao2021automatic} and M2SNet \cite{zhao2023m} reduce redundant features using subtraction modules to design more efficient models. Recently, transformers have been employed as skip connection modules to capture the global dependencies in medical images \cite{gao2021utnet, heidari2023hiformer}. Notably, UCTransNet \cite{wang2022uctransnet}, FCT \cite{tragakis2023fully}, and CFATransUNet \cite{wang2024cfatransunet} maintain global dependency by leveraging transformer-based skip connection frameworks. However, these models have complex architectures with more than \textbf{60M parameters}, which can be computationally expensive and inefficient. Our innovative TransGUNet addresses these challenges by utilizing cross-scale GNN with node attention and reducing the semantic gap with a significantly more efficient architecture comprising \textbf{25M parameters}. We compare the schemes and properties of various skip connection frameworks in Tab. \ref{tab:SkipConnectionEngineeringProperties} and Appendix (Fig. \ref{fig:SkipConnectionEngineeringScheme}).

\noindent \textbf{GNNs for Computer Vision.} GNNs have been traditionally employed for natural language processing \cite{kipf2016semi} and recommendation systems \cite{ying2018graph} owing to their ability to comprehend intricate relationships within datasets. Recently, in computer vision, GNNs \cite{han2022vision, han2023vision} have been actively explored to flexibly extract global dependencies and local features based on graphs, which are generalized data structures encompassing grids (CNN) and sequences (Transformer). For example, SFDGNet \cite{wang2023dynamic} extracts content-specific manipulated frequency features using GNN and comprehends complex spatial and frequency relationships. Additionally, GazeGNN \cite{wang2024gazegnn} integrates raw eye-gaze data and images into a unified representation graph for real-time disease classification. In particular, ViGUNet \cite{jiang2023vig} and PVT-GCASCADE \cite{rahman2024g} utilize GNNs to handle complex anatomical structures in medical image segmentation. Additionally, GTBA-Net \cite{xu2023graph}, TSGCNet \cite{duan20233d}, MSAGAANet \cite{wang2024multi}, TGNet \cite{zhang2025transgraphnet}, and GSENet \cite{li2025gse} tried to combine transformer and GNN for medical image segmentation. However, these models do not consider cross-scale features and lack a reliable spatial attention map. Noting these limitations, we carefully designed TransGUNet, which incorporates cross-scale features through ACS-GNN with EFS-based reliable spatial attention and fully utilizes global dependency.  
\section{Method}

\subsection{Encoder and Decoder in TransGUNet}

We used a Pyramid Pooling Transformer (P2T) \cite{wu2022p2t} comprising multiple pooling-based multi-head self-attention. P2T incurs a significantly lower computational cost and higher representation power than Vision Transformer (ViT) and Pyramid Vision Transformer (PVT), which are adopted in various medical image segmentation models \cite{dong2108polyp, zhang2022hsnet, liu2024cafe}. Inspired by previous studies, we utilized the same encoder architecture as the decoder to fully leverage global dependency. Although we primarily present the experimental results using P2T, we also provide various CNN and transformer backbones to demonstrate the versatility and robustness of the proposed approach across different backbone architectures in the Appendix (Tab. \ref{tab:ablation_backbone_networks}).

\subsection{ACS-GNN with EFS-based spatial attention for Skip Connection}

\begin{figure}[t]
    \centering
    \includegraphics[width=0.48\textwidth]{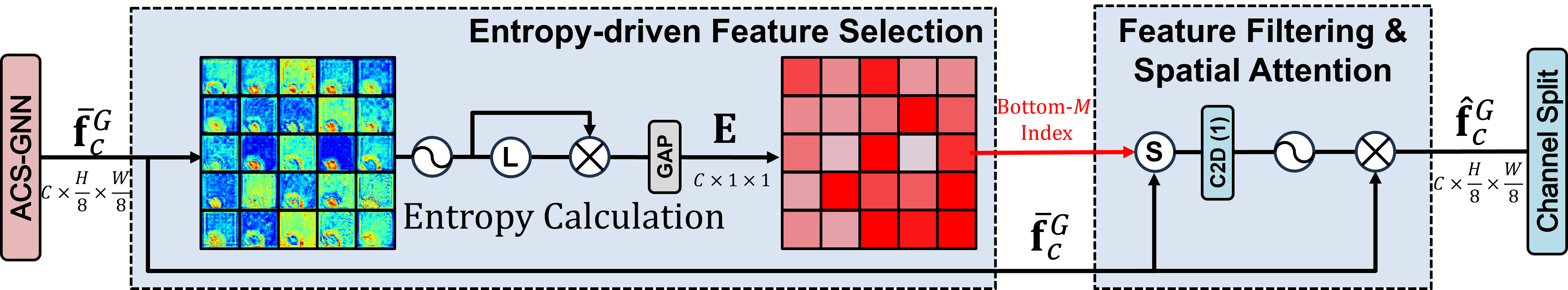}
    \caption{The overall block diagram of the Entropy-driven Feature Selection with spatial attention. Low and high transparency red indicates a high and low entropy score, respectively. We produce the spatial attention map using the indices corresponding to the $M$ channels with the lowest entropy scores, called Bottom-$M$ index.}
    \label{fig:ESA_based_spatial_attention}
\end{figure}

\noindent \textit{Motivation:} The human visual system (HVS) recognizes objects by dividing them into large parts and understanding them based on the connectivity strengths of each part \cite{majaj2015simple}. This process helps interpret complex scenes by identifying relationships between different parts of an image, leading to a holistic understanding of objects and their interactions \cite{palmer1999vision, marr2010vision}. Inspired by these principles, our approach employs a similar strategy of the HVS by transforming the cross-scale feature map into graphs to understand the complex anatomical structures in a high-dimensional feature space. However, significant noise and complex backgrounds create highly ambiguous visual signals that disturb the neural systems. The HVS mitigates this issue through signal filtering and attention processing \cite{posner1990attention, treue2001neural}. Therefore, we propose an entropy-based feature selection strategy that mimics these feature filtering and attention processes, called EFS-based spatial attention. The integration of these components enhances the preservation of global dependencies and the local details without ambiguity in attention mechanism. The overall architecture of the TransGUNet is illustrated in Fig. \ref{fig:TransGUNet}. The ACS-GNN with EFS-based spatial attention can be divided into four steps: \textit{1) Feature Preprocessing}, \textit{2) ACS-GNN (Fig. \ref{fig:TransGUNet} (c))}, \textit{3) EFS-based spatial attention (Fig. \ref{fig:ESA_based_spatial_attention})}, and \textit{4) Feature Postprocessing}.

\noindent \textbf{Feature Preprocessing.} Let $\mathbf{f}_{i} \in \mathbb{R}^{C_{i} \times \frac{H}{2^{i + 1}} \times \frac{W}{2^{i + 1}}}$ be the feature maps from the $i$-th encoder stage for $i = 1, 2, 3, 4$ where $(H, W)$ is the resolution of the input image. Because the number of channels in each stage primarily affects the complexity of the decoder, we employed a 2D convolution with a kernel size of $1 \times 1$ to reduce the number of channels to $C_{r}$. To obtain the cross-scale feature map $\mathbf{f}_{c}$, we resized them to the same resolution for $i = 1, 2, 3, 4$ as follows:
\begin{equation}
    \mathbf{f}^{'}_{i} = \textbf{Resize}_{(H_{t}, W_{t})} (\textbf{C2D}_{1 \times 1} (\mathbf{f}_{i})) \in \mathbb{R}^{C_{r} \times H_{t} \times W_{t}}
\end{equation}

\noindent where $\textbf{C2D}_{k \times k} ( \cdot )$, and $\textbf{Resize}_{(H_{t}, W_{t})} (\cdot)$ denote the 2D convolution with a kernel size of $k \times k$, and an operation to resize into the target spatial resolution $(H_{t}, W_{t})$, respectively. If the target and original resolutions of the input feature map differ, bilinear interpolation is used to upsample or downsample feature map to match the target resolution. Alternatively, if they have the same resolution, we do not apply resize, denoted as \say{\textit{Resolution Fix}} in Fig. \ref{fig:TransGUNet} (b). And then, we concatenate each resized feature map $\mathbf{f}_{c} = \left[ \mathbf{f}^{'}_{1}, \mathbf{f}^{'}_{2}, \mathbf{f}^{'}_{3}, \mathbf{f}^{'}_{4} \right] \in \mathbb{R}^{4C_{r} \times H_{t} \times W_{t}}$ where $\left[ \cdots \right]$ denotes concatenation between feature maps along the channel dimension. For convenience, we assume that $C = 4C_{r}$.

\noindent \textbf{Attentional Cross-Scale Graph Neural Network.} After obtaining the cross-scale feature $\mathbf{f}_{c} \in \mathbb{R}^{C \times H_{t} \times W_{t}}$, we apply 2D convolution with a kernel size of $1 \times 1$ and batch normalization (BN). Subsequently, the feature map is converted into a flattened vector $\mathbf{x} \in \mathbb{R}^{C \times N}$, where $N = H_{t}W_{t}$. Each pixel in $\mathbf{x}$ acts as a node in the graph. Additionally, a relative positional vector is added to each flattened vector element to preserve the position information. Next, we constructed the feature graph using the dilated $K$-nearest neighbors (KNN) algorithm. To implement the exchange of information between nodes, we adopted the Max-Relative graph convolution (MRConv) \cite{li2019deepgcns} owing to its simplicity and efficiency as it does not require learnable parameters for node aggregation. The MRConv and Update processes are implemented for graph convolution as $\mathbf{x}^{G} = \mathcal{G} (\mathbf{x})$. 

\noindent To improve feature aggregation by adaptively weighting the node importance, we applied node attention to prioritize critical features while learning their relevance. Based on ECANet \cite{wang2020eca}, we designed a node attention mechanism using a single 1D convolution operation with kernel size of $k$ and sigmoid function. Firstly, $\mathbf{x}^{G}$ is compressed into $\mathbf{z}_{\text{avg}}$ and $\mathbf{z}_{\text{max}}$ using Global Average Pooling and Global Max Pooling, respectively, and then each statistic is aggregated to produce a node attention map. Finally, such an attention mechanism is readily implemented as follows:
\begin{equation}
    \overline{\mathbf{x}}^{G} = \mathbf{x}^{G} \times \sigma \left( \sum_{d \in \{ \text{avg}, \text{max} \}} \textbf{C1D}_{k} (\mathbf{z}_{d}) \right)
\end{equation}

\noindent where $\textbf{C1D}_{k} ( \cdot )$ and $\sigma( \cdot )$ denote 1D convolution with a kernel size of $k$ and the sigmoid function, respectively. And, we reshape flattened vector $\overline{\mathbf{x}}^{G}$ into original feature map shape and apply 2D convolution with a kernel size of $1 \times 1$ and BN for more nonlinearity to obtain the refined feature map $\overline{\mathbf{X}}^{G} \in \mathbb{R}^{C \times H_{t} \times W_{t}}$. To address the oversmoothing problem \cite{li2018deeper, oono2020graph} in GNN, we also utilize Feed-Forward Networks with two consecutive convolutional layers and residual connections as follows:
\begin{equation}
    \overline{\mathbf{f}}^{G}_{c} = \textbf{BN} (\textbf{C2D}_{1 \times 1} (\delta(\textbf{BN} (\textbf{C2D}_{1 \times 1} (\overline{\mathbf{X}}^{G}))))) + \overline{\mathbf{X}}^{G}
\end{equation}

\noindent where $\delta( \cdot )$ and $\textbf{BN} (\cdot)$ denote the ReLU activation function for non-linearity and batch normalization, respectively.
\begin{figure}[t]
    \centering
    \includegraphics[width=0.48\textwidth]{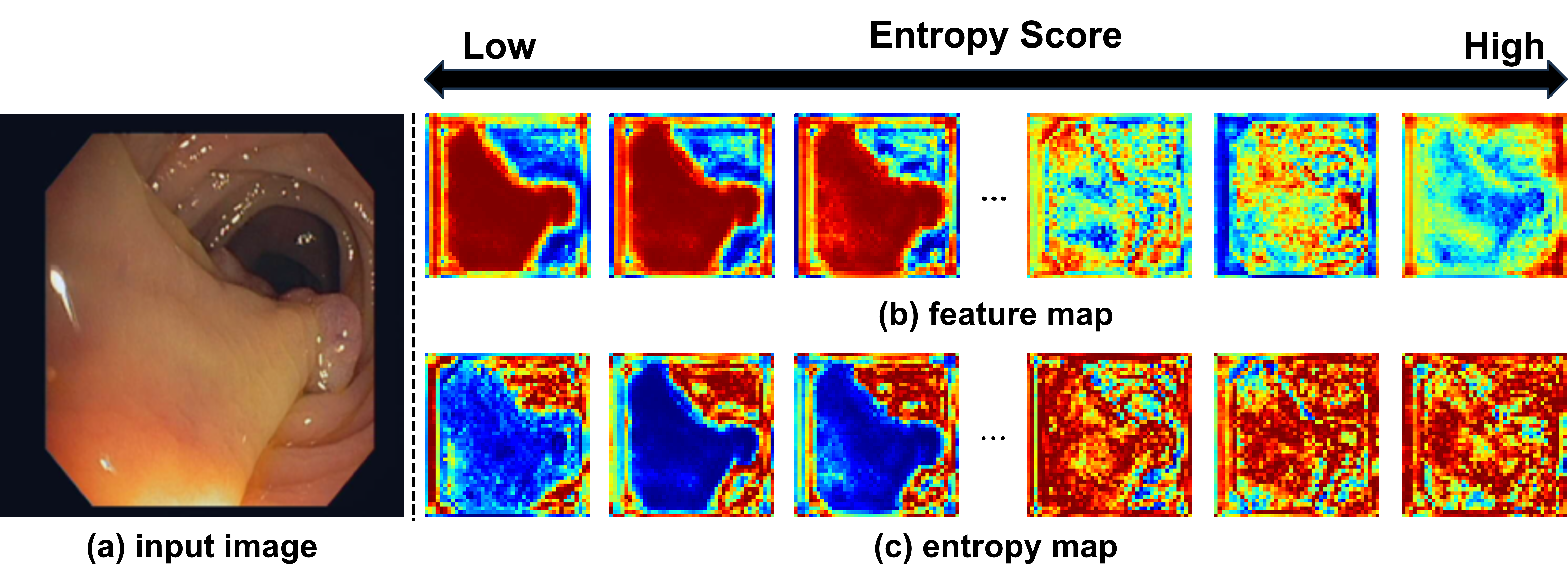}
    \caption{(a) Input image, (b) Feature map from ACS-GNN, (c) Entropy map according to each feature map. We calculated entropy map using Shannon Entropy at pixel level.}
    \label{fig:FeatureVisualization}
\end{figure}

\noindent \textbf{Spatial Attention with Entropy-driven Feature Selection.} Although a GNN can maintain global dependencies, various noise and complex backgrounds in medical images still result in an uninformative feature map containing a high-entropy score, leading to a poor spatial attention map \cite{chen2021lesion, shawn2024ct}. To address this issue, we propose \textit{entropy-driven feature selection (EFS)-based spatial attention}, which illustrated in Fig. \ref{fig:ESA_based_spatial_attention}, that filters uninformative feature maps and generates a more reliable spatial attention map. We observed that the channel with a high entropy is uniformly or noisily activated as shown in Fig. \ref{fig:FeatureVisualization}. Therefore, inspired by previous studies \cite{vu2019advent, wang2019boundary}, we calculated the entropy score $\mathbf{E} \in \mathbb{R}^{C}$ at the pixel level for each channel of the input feature map using Shannon Entropy as follows:
\begin{equation}
    \mathbf{E} = \frac{1}{HW} \sum_{h = 1}^{H} \sum_{w = 1}^{W} (-\sigma(\overline{\mathbf{f}}^{G}_{c})_{:, h, w} \log (\sigma(\overline{\mathbf{f}}^{G}_{c})_{:, h, w}))
\end{equation}

\noindent Subsequently, we only retained the feature map channels with the $M$ lowest entropy scores and used them for spatial attention as follows:
\begin{equation}
    \hat{\mathbf{f}}^{G}_{c} = \overline{\mathbf{f}}^{G}_{c} \times \sigma(\textbf{C2D}_{1 \times 1} ((\overline{\mathbf{f}}^{G}_{c})_{\textbf{sorting} \left(  \mathbf{E} \right)[: M]}))
\end{equation}

\noindent where $\textbf{sorting}(\cdot)$ denotes the sorting algorithm into ascending. Hence, $\textbf{sorting} \left(  \mathbf{E} \right)[: M]$ means the Bottom-$M$ index of feature map $\overline{\mathbf{f}}^{G}_{c}$ with low entropy score. We used the Introselect algorithm, the default sorting algorithm in PyTorch. This dual process of ACS-GNN and EFS-based spatial attention allows our model to comprehend the intricate anatomical structures in medical images and generate more reliable spatial attention maps.

\noindent \textbf{Feature Postprocessing.} We first divided $\hat{\mathbf{f}}^{G}_{c}$ along the channel dimension with an equal number of channels $C_{r}$. Subsequently, each feature map was resized and applied residual connection between the original feature map to enhance the training stability for $i = 1, 2, 3, 4$ as follows:
\begin{equation}
    \hat{\mathbf{f}}^{G}_{i} = \mathbf{f}^{'}_{i} + \textbf{Resize}_{(\frac{H}{2^{i + 1}}, \frac{W}{2^{i + 1}})} \left( (\hat{\mathbf{f}}^{G}_{c})_{C_{r} \cdot (i - 1):C_{r} \cdot i} \right)
\end{equation}

\noindent For convenience, let $\mathbf{D}_{1} = \hat{\mathbf{f}}^{G}_{4}$. Then, we produce the decoder block output $\mathbf{D}_{i + 1}$ for $i = 1, 2, 3$ as $\mathbf{D}_{i + 1} = \textbf{Decoder}_{i} \left( \left[ \hat{\mathbf{f}}^{G}_{4 - i}, \textbf{Up}_{2} ( \mathbf{D}_{i} ) \right] \right)$ where $\textbf{Up}_{n} ( \cdot )$ and $\textbf{Decoder}_{i} ( \cdot )$ denote a bilinear upsampling with scale factor $2^{n - 1}$ and $i$-th transformer decoder block, respectively. We summarized the technical novelty and impact of TransGUNet in the Appendix.

\subsection{Training Procedure}
We employed a multi-task learning with deep supervision to enhance the representation power of the model and mitigate the gradient vanishing problem. To implement this, we obtained four predictions $\mathbf{O}_{i}$ from $\mathbf{D}_{i}$ for $i = 1, 2, 3, 4$ by applying 2D convolution with kernel size of $1 \times 1$ following sigmoid function and upsampling at each stage. We denote $\mathbf{O}_{i} = \{ \mathbf{R}_{i}, \mathbf{B}_{i} \}$ as the multiple outputs representing region $\mathbf{R}_{i}$ and boundary $\mathbf{B}_{i}$ predictions, respectively. The total loss function is defined as:
\begin{equation}
    \mathcal{L}_{total} = \sum_{i=1}^{4} \left( \mathcal{L}_{R} (\mathbf{R}_{t}, \mathbf{R}_{i}) + \mathcal{L}_{B} (\mathbf{B}_{t}, \mathbf{B}_{i}) \right)  
\end{equation}

\noindent where $\mathbf{R}_{t}$ and $\mathbf{B}_{t}$ denote the ground truths of region and the boundary, respectively. We obtained $\mathbf{B}_{t}$  by applying an anisotropic Sobel edge detection filter \cite{kanopoulos1988design} to $\mathbf{R}_{t}$. The loss function for region prediction was defined as $\mathcal{L}_{R} = \mathcal{L}^{w}_{IoU} + \mathcal{L}^{w}_{bce}$, where $\mathcal{L}^{w}_{IoU}$ and $\mathcal{L}^{w}_{bce}$ are the weighted IoU and binary cross entropy (BCE) loss functions, respectively. This loss function is identically defined in previous studies \cite{fan2020pranet, zhao2023m, nam2024modality}. Additionally, we defined boundary loss function $\mathcal{L}_{B}$ as the BCE loss function. 

\begin{table*}[t]
    \centering
    \scriptsize
    \setlength\tabcolsep{5.0pt} 
    \begin{tabular}{c|cc|cc|cc|cc|cc|cc|c}
    \hline
    \multicolumn{1}{c|}{\multirow{2}{*}{Method}} & \multicolumn{2}{c|}{Synapse \cite{Synapse_dataset}} & \multicolumn{2}{c|}{ISIC2018 \cite{gutman2016skin}} & \multicolumn{2}{c|}{COVID19-1 \cite{ma_jun_2020_3757476}} & \multicolumn{2}{c|}{BUSI \cite{al2020dataset}} & \multicolumn{2}{c}{CVC-ClinicDB \cite{bernal2015wm}} & \multicolumn{2}{c|}{Kvasir-SEG \cite{jha2020kvasir}} & \multicolumn{1}{c}{\multirow{2}{*}{$P$-value}} \\ \cline{2-13}
     & DSC & mIoU & DSC & mIoU & DSC & mIoU & DSC & mIoU & DSC & mIoU & DSC & mIoU & \\
     \hline
     UNet \cite{ronneberger2015u}         & 69.8 \tiny{(0.6)} & 58.9 \tiny{(0.4)} & 86.9 \tiny{(0.8)} & 80.2 \tiny{(0.7)} & 47.7 \tiny{(0.6)} & 38.6 \tiny{(0.6)} & 69.5 \tiny{(0.3)} & 60.2 \tiny{(0.2)} & 76.5 \tiny{(0.8)} & 69.1 \tiny{(0.9)} & 80.5 \tiny{(0.3)} & 72.6 \tiny{(0.4)} &  9.3E-06 \\
     UNet++ \cite{zhou2018unet++}         & 79.3 \tiny{(0.2)} & 69.8 \tiny{(0.1)} & 87.3 \tiny{(0.2)} & 80.2 \tiny{(0.1)} & 65.6 \tiny{(0.7)} & 57.1 \tiny{(0.8)} & 72.4 \tiny{(0.1)} & 62.5 \tiny{(0.2)} & 79.7 \tiny{(0.2)} & 73.6 \tiny{(0.4)} & 84.3 \tiny{(0.3)} & 77.4 \tiny{(0.2)} &  4.0E-05 \\
     CENet \cite{gu2019net}            & 75.2 \tiny{(0.3)} & 64.6 \tiny{(0.4)} & 89.1 \tiny{(0.2)} & 82.1 \tiny{(0.1)} & 76.3 \tiny{(0.4)} & 69.2 \tiny{(0.5)} & 79.7 \tiny{(0.6)} & 65.0 \tiny{(0.5)} & 89.3 \tiny{(0.3)} & 83.4 \tiny{(0.2)} & 89.5 \tiny{(0.7)} & 83.9 \tiny{(0.7)} &  7.1E-07  \\
     TransUNet \cite{chen2021transunet}      & 77.5 \tiny{(0.2)} & 67.1 \tiny{(0.6)} & 87.3 \tiny{(0.2)} & 81.2 \tiny{(0.8)} & 75.6 \tiny{(0.4)} & 68.8 \tiny{(0.2)} & 75.5 \tiny{(0.5)} & 68.4 \tiny{(0.1)} & 87.4 \tiny{(0.2)} & 82.4 \tiny{(0.1)} & 86.4 \tiny{(0.4)} & 80.1 \tiny{(0.4)} & 4.5E-09 \\
     MSRFNet \cite{srivastava2021msrf}          & 77.2 \tiny{(0.5)} & 67.6 \tiny{(0.3)} & 88.2 \tiny{(0.2)} & 81.3 \tiny{(0.2)} & 75.2 \tiny{(0.4)} & 68.0 \tiny{(0.4)} & 76.6 \tiny{(0.7)} & 68.1 \tiny{(0.7)} & 83.2 \tiny{(0.9)} & 76.5 \tiny{(1.1)} & 86.1 \tiny{(0.5)} & 79.3 \tiny{(0.4)} & 1.0E-07 \\
     DCSAUNet \cite{xu2023dcsau}         & 71.0 \tiny{(0.3)} & 55.8 \tiny{(0.1)} & 89.0 \tiny{(0.3)} & 82.0 \tiny{(0.3)} & 75.3 \tiny{(0.4)} & 68.2 \tiny{(0.4)} & 73.7 \tiny{(0.5)} & 65.0 \tiny{(0.5)} & 80.6 \tiny{(1.2)} & 73.7 \tiny{(1.1)} & 82.6 \tiny{(0.5)} & 75.2 \tiny{(0.5)} & 5.4E-08 \\
     M2SNet \cite{zhao2023m}         & 77.1 \tiny{(0.1)} & 67.5 \tiny{(0.4)} & 89.2 \tiny{(0.2)} & 83.4 \tiny{(0.2)} & 81.7 \tiny{(0.4)} & 74.7 \tiny{(0.5)} & 80.4 \tiny{(0.8)} & 72.5 \tiny{(0.7)} & 92.2 \tiny{(0.8)} & \textit{88.0} \tiny{(0.8)} & 91.2 \tiny{(0.5)} & 86.1 \tiny{(0.6)} & 2.2E-05 \\
     ViGUNet \cite{jiang2023vig}     & 72.3 \tiny{(1.5)} & 67.7 \tiny{(1.2)} & 88.3 \tiny{(1.2)} & 81.3 \tiny{(0.1)} & 70.2 \tiny{(0.3)} & 62.1 \tiny{(0.3)} & 70.9 \tiny{(0.2)} & 61.3 \tiny{(0.1)} & 77.5 \tiny{(0.4)} & 69.7 \tiny{(0.7)} & 79.6 \tiny{(0.6)} & 71.3 \tiny{(0.6)} & 3.8E-08 \\
     PVT-GCAS \cite{rahman2024g}    & 78.4 \tiny{(0.3)} & 68.9 \tiny{(0.2)} & \textit{90.6} \tiny{(0.3)} & \textit{84.2} \tiny{(0.5)} & 82.3 \tiny{(0.4)} & 74.8 \tiny{(0.5)} & \textit{82.0} \tiny{(0.5)} & \textit{73.6} \tiny{(0.5)} & 92.2 \tiny{(0.6)} & 87.6 \tiny{(0.6)} & 91.6 \tiny{(0.2)} & 86.8 \tiny{(0.4)} & 1.8E-04 \\
     CFATUNet \cite{wang2024cfatransunet}     & \textit{80.5} \tiny{(1.4)} & \textit{70.4} \tiny{(1.8)} & 90.3 \tiny{(0.4)} & 83.6 \tiny{(0.7)} & 80.4 \tiny{(0.5)} & 73.6 \tiny{(0.2)} & 80.6 \tiny{(0.3)} & 72.8 \tiny{(0.4)} & 91.0 \tiny{(0.2)} & 86.2 \tiny{(0.1)} & \textit{92.1} \tiny{(0.5)} & \textit{87.2} \tiny{(0.5)} & 3.2E-05 \\
     MADGNet \cite{nam2024modality}         & 79.3 \tiny{(1.2)} & 69.8 \tiny{(1.3)} & 90.2 \tiny{(0.1)} & 83.7 \tiny{(0.2)} & \textit{83.7} \tiny{(0.2)} & \textit{76.8} \tiny{(0.2)} & 81.3 \tiny{(0.4)} & 73.4 \tiny{(0.5)} & \textbf{\underline{93.9}} \tiny{(0.6)} & \textbf{\underline{89.5}} \tiny{(0.5)} & 90.7 \tiny{(0.8)} & 85.3 \tiny{(0.8)}  & 6.5E-02 \\ 
     GSENet \cite{li2025gse} & 79.1 \tiny{(1.6)} & 69.2 \tiny{(1.7)} & 90.7 \tiny{(0.2)} & \textit{84.2} \tiny{(0.2)} & 80.7 \tiny{(0.3)} & 73.7 \tiny{(0.4)} & 80.6 \tiny{(0.4)} & 72.8 \tiny{(0.5)} & 91.5 \tiny{(0.2)} & 86.4 \tiny{(0.1)} & \textit{92.1} \tiny{(0.3)} & \textit{87.2} \tiny{(0.4)} & 3.2E-02 \\
     \hline
     \multicolumn{1}{c|}{\multirow{2}{*}{\textbf{TransGUNet}}} & \textbf{\underline{80.9}}  \tiny{(0.4)} & \textbf{\underline{71.4}}  \tiny{(0.2)} & \textbf{\underline{91.1}}  \tiny{(0.3)} & \textbf{\underline{84.8}}  \tiny{(0.4)} & \textbf{\underline{84.0}} \tiny{(0.2)} & \textbf{\underline{77.0}} \tiny{(0.5)} & \textbf{\underline{82.7}} \tiny{(0.3)} & \textbf{\underline{74.7}} \tiny{(0.4)} & \textit{92.3} \tiny{(0.2)} & 87.7 \tiny{(0.4)} & \textbf{\underline{93.1}} \tiny{(0.1)} & \textbf{\underline{88.4}} \tiny{(0.2)} & \multicolumn{1}{c}{\multirow{2}{*}{-}} \\ \cline{2-13}
     & \textbf{+0.4} & \textbf{+1.0} & \textbf{+0.5} & \textbf{+0.6} & \textbf{+0.3} & \textbf{+0.2} & \textbf{+0.7} & \textbf{+1.1} & \textbf{-1.6} & \textbf{-1.8} & \textbf{+1.0} & \textbf{+1.2} & \\
    \hline
    \end{tabular}
    \caption{Segmentation results on six different datasets with \textit{seen} clinical settings. $(\cdot)$ denotes the standard deviations of multiple experiment results. We also provide the Wilcoxon signed rank test results ($P$-value) compared to our method and other methods.}
    \label{tab:comparison_sota_in_domain}
\end{table*}

\begin{table*}[t]
    \centering
    \scriptsize
    \setlength\tabcolsep{1.25pt} 
    \begin{tabular}{c|cc|cc|cc|cc|cc|cc|cc|cc|c}
    \hline
    \multicolumn{1}{c|}{\multirow{2}{*}{Method}} & \multicolumn{2}{c|}{AMOS-CT \cite{ji2022amos}} & \multicolumn{2}{c|}{AMOS-MRI \cite{ji2022amos}} & \multicolumn{2}{c|}{PH2 \cite{mendoncca2013ph}} & \multicolumn{2}{c|}{COVID19-2 \cite{COVID19_2}} & \multicolumn{2}{c|}{STU \cite{zhuang2019rdau}} & \multicolumn{2}{c}{CVC-300 \cite{vazquez2017benchmark}} & \multicolumn{2}{c}{CVC-Colon \cite{tajbakhsh2015automated}} & \multicolumn{2}{c|}{ETIS \cite{silva2014toward}} & \multicolumn{1}{c}{\multirow{2}{*}{$P$-value}} \\ \cline{2-17}
     & DSC & mIoU & DSC & mIoU & DSC & mIoU & DSC & mIoU & DSC & mIoU & DSC & mIoU & DSC & mIoU & DSC & mIoU & \\
     \hline
     UNet \cite{ronneberger2015u}             & 56.3 \tiny{(2.1)} & 44.8 \tiny{(1.4)} & 8.3 \tiny{(2.5)} & 5.3 \tiny{(1.9)} & 90.3 \tiny{(0.1)}  & 83.5 \tiny{(0.1)} & 47.1 \tiny{(0.7)} & 37.7 \tiny{(0.6)} & 71.6 \tiny{(1.0)} & 61.6 \tiny{(0.7)} & 66.1 \tiny{(2.3)} & 58.5 \tiny{(2.1)} & 56.8 \tiny{(1.3)} & 49.0 \tiny{(1.2)} & 41.6 \tiny{(1.1)} & 35.4 \tiny{(1.0)} & 1.7E-08  \\
     UNet++ \cite{zhou2018unet++}             & 67.5 \tiny{(2.3)} & 56.6 \tiny{(2.8)} & 6.0 \tiny{(1.2)} & 3.8 \tiny{(1.5)} & 88.0 \tiny{(0.3)}  & 80.1 \tiny{(0.3)}  & 50.5 \tiny{(3.3)} & 40.9 \tiny{(3.7)} & 77.3 \tiny{(0.4)} & 67.8 \tiny{(0.3)} & 64.4 \tiny{(2.2)} & 58.4 \tiny{(2.0)} & 57.5 \tiny{(0.4)} & 50.2 \tiny{(0.4)} & 39.1 \tiny{(2.4)} & 34.0 \tiny{(2.1)} & 3.1E-07 \\
     CENet \cite{gu2019net}                   & 67.9 \tiny{(2.3)}  & 56.5 \tiny{(2.4)} & 14.5 \tiny{(1.4)} & 9.0 \tiny{(1.6)} & 90.5 \tiny{(0.1)} & 83.3 \tiny{(0.1)} & 60.1 \tiny{(0.3)} & 49.9 \tiny{(0.3)} & 86.0 \tiny{(0.7)} & 77.2 \tiny{(0.9)} & 85.4 \tiny{(1.6)} & 78.2 \tiny{(1.4)} & 65.9 \tiny{(1.6)} & 59.2 \tiny{(0.1)} & 57.0 \tiny{(3.4)} & 51.4 \tiny{(0.5)} & 1.8E-05 \\
     TransUNet \cite{chen2021transunet}       & 68.3 \tiny{(1.1)} & 57.7 \tiny{(1.2)} & 9.1 \tiny{(2.3)} & 5.8 \tiny{(2.5)} & 89.5 \tiny{(0.3)} & 82.1 \tiny{(0.4)} & 56.9 \tiny{(1.0)} & 48.0 \tiny{(0.7)} & 41.4 \tiny{(4.5)} & 32.1 \tiny{(4.2)} & 85.0 \tiny{(0.6)} & 77.3 \tiny{(0.3)} & 63.7 \tiny{(0.1)} & 58.4 \tiny{(0.3)} & 50.1 \tiny{(0.5)} & 44.0 \tiny{(2.3)} & 7.3E-06 \\
     MSRFNet \cite{srivastava2021msrf}        & 61.8 \tiny{(1.3)} & 51.3 \tiny{(1.7)} & 6.5 \tiny{(2.3)} & 4.2 \tiny{(1.9)} & 90.5 \tiny{(0.3)} & 83.5 \tiny{(0.3)} & 58.3 \tiny{(0.8)} & 48.4 \tiny{(0.6)} & 84.0 \tiny{(5.5)} & 75.2 \tiny{(5.2)} & 72.3 \tiny{(2.2)} & 65.4 \tiny{(2.2)} & 61.5 \tiny{(1.0)} & 54.8 \tiny{(0.8)} & 38.3 \tiny{(0.6)} & 33.7 \tiny{(0.7)} & 5.6E-06 \\
     DCSAUNet \cite{xu2023dcsau}              & 45.7 \tiny{(1.2)} & 36.3 \tiny{(1.5)} & 1.7 \tiny{(0.5)} & 1.1 \tiny{(0.2)} & 89.0 \tiny{(0.4)} & 81.5 \tiny{(0.3)} & 52.4 \tiny{(1.2)} & 44.0 \tiny{(0.7)} & 86.1 \tiny{(0.5)} & 76.5 \tiny{(0.8)} & 68.9 \tiny{(4.0)} & 59.8 \tiny{(3.9)} & 57.8 \tiny{(0.4)} & 49.3 \tiny{(0.4)} & 43.0 \tiny{(3.0)} & 36.1 \tiny{(2.9)} & 6.9E-07  \\
     M2SNet \cite{zhao2023m}                  & 69.6 \tiny{(1.3)} & 58.5 \tiny{(0.5)} & 22.0 \tiny{(0.6)} & 14.7 \tiny{(0.8)} & 90.7 \tiny{(0.3)} & 83.5 \tiny{(0.5)} & 68.6 \tiny{(0.1)} & 58.9 \tiny{(0.2)} & 79.4 \tiny{(0.7)} & 69.3 \tiny{(0.6)} & \textbf{\underline{90.0}} \tiny{(0.2)} & \textbf{\underline{83.2}} \tiny{(0.3)} & 75.8 \tiny{(0.7)} & 68.5 \tiny{(0.5)} & 74.9 \tiny{(1.3)} & 67.8 \tiny{(1.4)} & 1.3E-04 \\
     ViGUNet \cite{jiang2023vig}              & 50.8 \tiny{(2.7)} & 42.8 \tiny{(2.3)} & 6.5 \tiny{(0.2)} & 4.5 \tiny{(0.3)} & 90.1 \tiny{(0.7)} & 82.8 \tiny{(0.5)} & 49.7 \tiny{(0.3)} & 41.9 \tiny{(0.4)} & 75.5 \tiny{(1.2)} & 65.0 \tiny{(1.1)} & 72.9 \tiny{(0.4)} & 62.8 \tiny{(0.2)} & 53.7 \tiny{(0.6)} & 53.7 \tiny{(0.7)} & 38.6 \tiny{(3.1)} & 31.6 \tiny{(2.2)} & 3.9E-09 \\
     PVT-GCAS \cite{rahman2024g}          & 69.3 \tiny{(1.4)} & 58.5 \tiny{(1.6)} & 32.8 \tiny{(2.4)}  & 24.3 \tiny{(2.2)} & \textit{91.5} \tiny{(1.3)} & 84.9 \tiny{(1.5)} & 71.0 \tiny{(0.4)} & 60.4 \tiny{(0.4)} & 86.4 \tiny{(0.6)} & 76.6 \tiny{(0.2)} & 88.2 \tiny{(0.4)} & 81.0 \tiny{(0.5)} & 79.5 \tiny{(0.9)} & 71.6 \tiny{(0.9)} & \textit{79.5} \tiny{(0.7)} & \textit{71.6} \tiny{(1.0)} & 3.9E-04 \\
     CFATUNet \cite{wang2024cfatransunet} & 68.0 \tiny{(2.6)} & 56.7 \tiny{(2.1)} & 35.9 \tiny{(3.2)} & 25.9 \tiny{(3.6)} & \textit{91.5} \tiny{(0.6)} & \textit{85.0} \tiny{(0.7)} & 65.7 \tiny{(1.2)} & 56.2 \tiny{(1.1)} & \textit{87.9} \tiny{(0.1)} & \textit{79.2} \tiny{(0.2)} & \textit{89.1} \tiny{(0.6)} & 82.4 \tiny{(0.6)} & 78.0 \tiny{(0.9)} & 70.3 \tiny{(0.8)} & 77.0 \tiny{(0.6)} & 69.5 \tiny{(0.7)} & 8.8E-05 \\
     MADGNet \cite{nam2024modality}           & \textit{74.9} \tiny{(1.2)} & \textit{64.4} \tiny{(1.3)} & 14.8 \tiny{(2.5)} & 9.8 \tiny{(2.7)} & 91.3 \tiny{(0.1)}  & 84.6 \tiny{(0.1)} & \textit{72.2} \tiny{(0.1)} & \textbf{\underline{62.6}} \tiny{(0.3)} & \textbf{\underline{88.4}} \tiny{(1.0)} & \textbf{\underline{79.9}} \tiny{(1.5)} & 87.4 \tiny{(0.4)} & 79.9 \tiny{(0.4)} & 77.5 \tiny{(1.1)} & 69.7 \tiny{(1.2)} & 77.0 \tiny{(0.3)} & 69.7 \tiny{(0.5)} & 1.3E-02 \\ 
     GSENet \cite{li2025gse} & 74.2 \tiny{(0.2)} & 63.4 \tiny{(0.4)} & \textit{36.3} \tiny{(4.0)} & \textit{27.2} \tiny{(3.5)} & \textit{91.5} \tiny{(0.3)} & \textit{85.0} \tiny{(0.4)} & 66.3 \tiny{(0.3)} & 56.5 \tiny{(0.4)} & 86.3 \tiny{(0.4)} & 76.7 \tiny{(0.5)} & 89.0 \tiny{(0.1)} & 82.1 \tiny{(0.1)} & \textit{80.1} \tiny{(1.4)} & \textit{71.9} \tiny{(1.2)} & 79.3 \tiny{(0.4)} & 71.2 \tiny{(0.3)}  & 1.2E-02 \\
     \hline
     \multicolumn{1}{c|}{\multirow{2}{*}{\textbf{TransGUNet}}} & \textbf{\underline{76.5}} \tiny{(0.5)} & \textbf{\underline{66.2}} \tiny{(0.4)} & \textbf{\underline{47.2}} \tiny{(1.1)} & \textbf{\underline{35.6}} \tiny{(1.2)} & \textbf{\underline{91.7}} \tiny{(0.3)} & \textbf{\underline{85.2}} \tiny{(0.2)} & \textbf{\underline{73.0}} \tiny{(0.2)} & \textit{62.4} \tiny{(0.2)} & 87.4 \tiny{(0.1)} & 78.2 \tiny{(0.4)} & \textbf{\underline{90.0}} \tiny{(0.3)} & \textit{83.1} \tiny{(0.1)} & \textbf{\underline{82.0}} \tiny{(0.2)} & \textbf{\underline{74.1}} \tiny{(0.3)} & \textbf{\underline{81.3}} \tiny{(0.3)} & \textbf{\underline{73.1}} \tiny{(0.2)} & \multicolumn{1}{c}{\multirow{2}{*}{-}} \\ \cline{2-17}
      & \textbf{+1.6} & \textbf{+1.8} & \textbf{+10.9} & \textbf{+8.4} & \textbf{+0.2} & \textbf{+0.2} & \textbf{+0.8} & \textbf{-0.2} & \textbf{-1.0} & \textbf{-1.7} & \textbf{+0.0} & \textbf{-0.1} & \textbf{+1.9} & \textbf{+2.2} & \textbf{+1.8} & \textbf{+1.5} & \\
    \hline
    \end{tabular}
    \caption{Segmentation results on eight different datasets with \textit{unseen} clinical settings. $(\cdot)$ denotes the standard deviations of multiple experiment results. We also provide the Wilcoxon signed rank test results ($P$-value) compared to our method and other methods.}
    \label{tab:comparison_sota_out_domain}
\end{table*}
\section{Experiment Results}

\subsection{Experiment Settings}

Each model was trained and evaluated on five medical segmentation tasks, including multi-organ, skin cancer, COVID-19 infection, breast tumors, and polyp. For convenience, we denote the \textit{seen} clinical settings (Tab. \ref{tab:comparison_sota_in_domain}) as the test dataset, which has the same distribution with the training dataset. Moreover, we additionally evaluate the domain generalizability of each model on eight external segmentation datasets using different distributions for the training and test datasets, which are referred to as \textit{unseen} clinical settings (Tab. \ref{tab:comparison_sota_out_domain}). Due to the page limit, we present the detailed dataset description and split information in the Appendix (Tab. \ref{tab:seen_clinical_dataset} and \ref{tab:unseen_clinical_dataset}). To evaluate the performance of each model, we selected two metrics, the Dice Score Coefficient (DSC) and mean Intersection over Union (mIoU), which are widely used in medical image segmentation. Additionally, the quantitative results with more various metrics are also available in the Appendix.

We compared the proposed \textbf{TransGUNet (Ours)} with twelve representative medical image segmentation models, including UNet \cite{ronneberger2015u}, UNet++ \cite{zhou2018unet++}, CENet \cite{gu2019net}, TransUNet \cite{chen2021transunet}, MSRFNet \cite{srivastava2021msrf}, DCSAUNet \cite{xu2023dcsau}, M2SNet \cite{zhao2023m}, ViGUNet \cite{jiang2023vig}, PVT-GCAS \cite{rahman2024g}, CFATUNet \cite{wang2024cfatransunet}, MADGNet \cite{nam2024modality}, and GSENet \cite{li2025gse}. In all results, we report the mean performance of three trials for reliability. In all tables, \textbf{\underline{Bold}} and \textit{italic} are the first and second best performance results, respectively. And, the last row in Tab. \ref{tab:comparison_sota_in_domain} and \ref{tab:comparison_sota_out_domain} indicates the performance gap between \textbf{TransGUNet} and other second best method.

\subsection{Implementation Details}
We implemented TransGUNet on a single NVIDIA RTX 3090 Ti in Pytorch 1.8.

\noindent\textbf{Multi-Organ Segmentation.} Following the previous literature \cite{wang2024cfatransunet}, we employ an Adam optimizer with a learning rate of 0.001 for multi-organ segmentation. We optimize each model using a batch size of 24 and train them for 150 epochs. During training, we used flipping with a probability of 50\% and rotation between $-20^{\circ}$ and $20^{\circ}$. Because volumes in \textit{seen} and \textit{unseen} clinical settings have different resolutions, all images were resized to $224 \times 224$. We would like to clarify that we used identical settings in CFATUNet, which is the most recent multi-organ segmentation model. Note that we utilize the same settings on the multi-organ segmentation task to train all models for fair comparison.

\noindent\textbf{Binary Segmentation.} We started with an initial learning rate of $10^{-4}$ using the Adam optimizer and reduced the parameters of each model to $10^{-6}$ using a cosine annealing learning rate scheduler. We optimized each model using a batch size of 16 and trained them for 50, 100, 100, and 200 epochs on polyp, skin cancer, breast tumors, and COVID-19 infection segmentation tasks. In the training process, horizontal and vertical flips were applied with a 50\% probability, along with rotations ranging from $-5^{\circ}$ to $5^{\circ}$, as part of a multi-scale training strategy. This approach is commonly utilized in medical image segmentation models \cite{fan2020pranet, zhao2021automatic, zhao2023m, nam2024modality}. Because images in each dataset have different resolutions, all images were resized to $352 \times 352$. Additionally, we would like to clarify that we used identical settings in M2SNet and MADGNet, which are the most representative medical image segmentation methods.

\begin{figure}[t]
    \centering
    \includegraphics[width=0.48\textwidth]{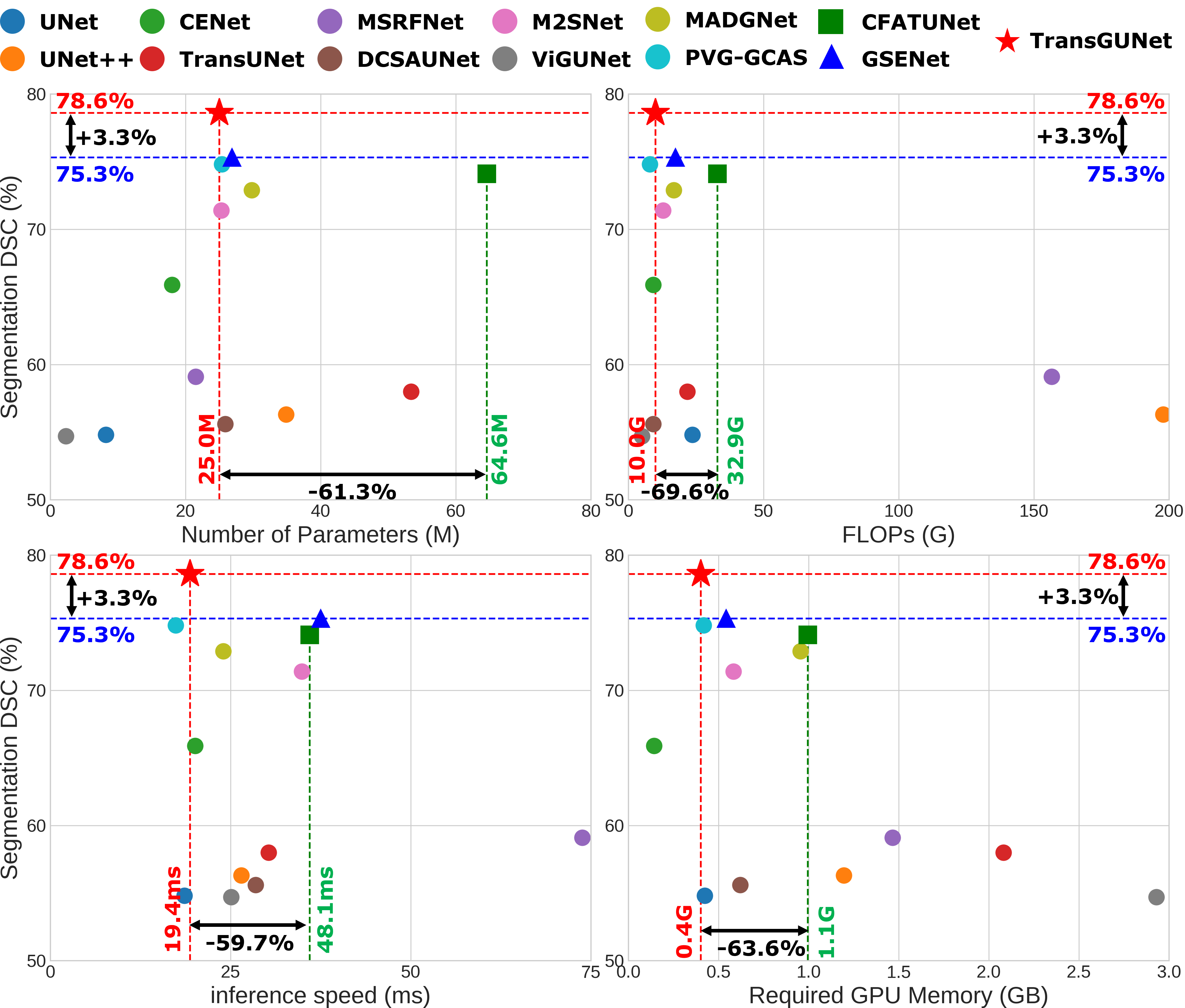}
        \caption{Comparison of parameters (M), FLOPs (G), inference speed (ms), and required GPU memory (G) vs segmentation performance (DSC) on average for all \textit{unseen} datasets.} \vspace{-0.50cm}
    \label{fig:EfficiencyAnalysis}
\end{figure}

\noindent\textbf{Hyperparameters of TransGUNet.} Key hyperparameters for TransGUNet on all datasets were set to $C_{r} = 64$ for efficiency and $(H_{t}, W_{t}) = (\frac{H}{8}, \frac{W}{8}), K = 11, k = 3$ in ACS-GNN and $M = 64$ in EFS-based spatial attention. In the Appendix, we provide the experiment results on various hyperparameter settings (Tab. \ref{tab:ablation_backbone_networks},
\ref{tab:ablation_target_resolution},
\ref{tab:ablation_eca_kernel_size}, \ref{tab:ablation_repetition_time}).

\begin{figure*}[t]
    \centering
    \includegraphics[width=\textwidth]{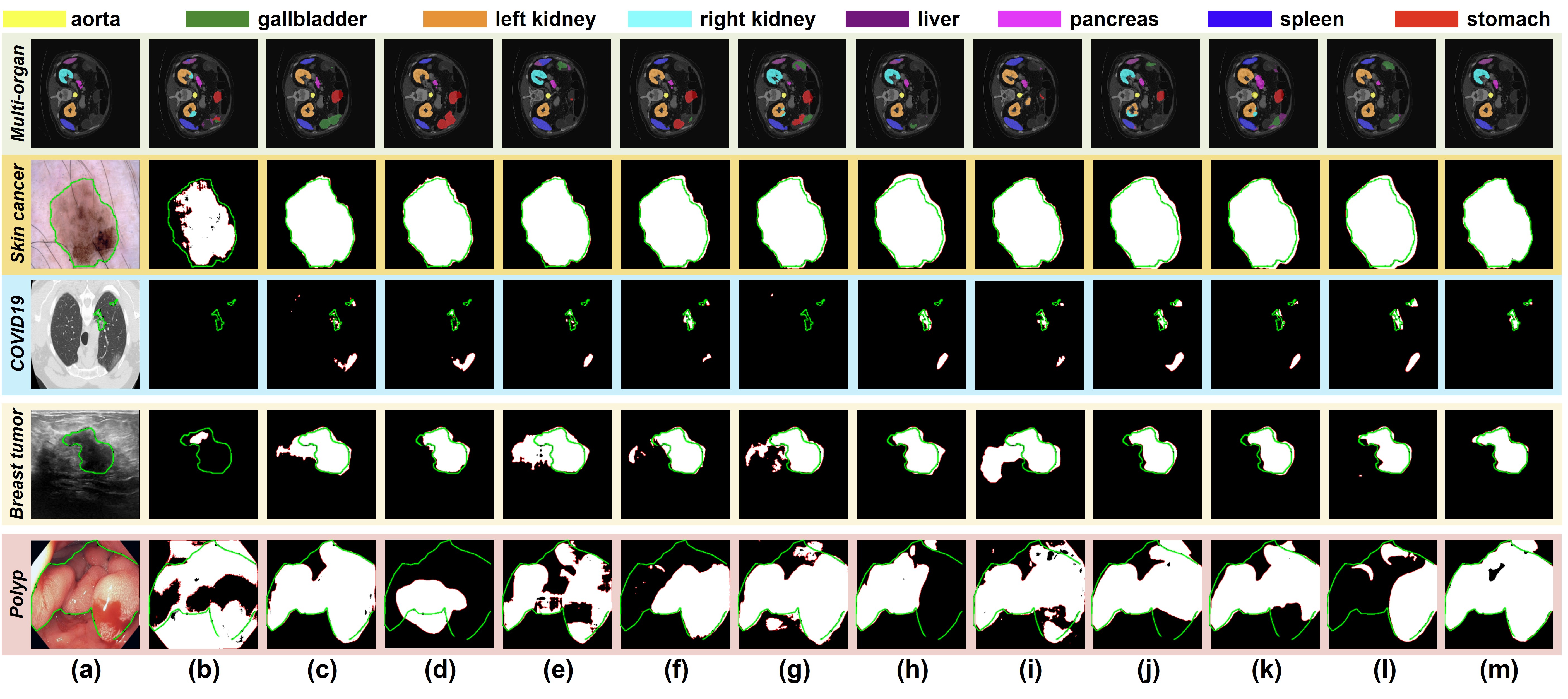}
    \caption{Qualitative comparison of other methods and TransGUNet.  (a) Input images with ground truth. (b) UNet \cite{ronneberger2015u}. (c) UNet++ \cite{zhou2018unet++}. (d) CENet \cite{gu2019net}. (e) TransUNet \cite{chen2021transunet}. (f) MSRFNet \cite{srivastava2021msrf}. (g) DCSAUNet \cite{xu2023dcsau}. (h) ViGUNet \cite{jiang2023vig}. (i) M2SNet \cite{zhao2023m}. (j) PVT-GCASCADE \cite{rahman2024g}. (k) CFATransUNet \cite{wang2024cfatransunet}. (l) MADGNet \cite{nam2024modality}. (m) GSENet \cite{li2025gse}. (n) \textbf{TransGUNet (Ours)}. \textcolor{green}{\textbf{Green}} and \textcolor{red}{\textbf{Red}} lines denote the boundaries of the ground truth and prediction, respectively.} \vspace{-0.25cm}
    \label{fig:QualitativeResults}
\end{figure*}

\subsection{Comparison with State-of-the-art models}

We used the same model as that in Tab. \ref{tab:comparison_sota_in_domain} to evaluate the domain generalizability for unseen clinical settings Tab. \ref{tab:comparison_sota_out_domain}. For convenience, we denote $( \cdot, \cdot )$ as the performance improvement gap between TransGUNet and other models for seen and unseen clinical settings. As listed in Tab. \ref{tab:comparison_sota_in_domain} and \ref{tab:comparison_sota_out_domain}, TransGUNet achieved the highest segmentation performance across all various datasets on average. Compared to UNet++, M2SNet, and CFATransUNet, which focus on enhancing the skip connection framework, TransGUNet exhibited DSC improvements of (9.3\%, 22.3\%), (2.1\%, 7.2\%), and (1.6\%, 4.5\%), respectively. Additionally, compared to PVT-GCASCADE, which uses a single-scale GNN with spatial attention, TransGUNet demonstrated DSC improvement of (1.2\%, 3.8\%). Although MADGNet achieved the state-of-the-art performance in seen clinical settings, TransGUNet exhibited significant DSC improvement of 5.6\% for unseen clinical settings on average. Surprisingly, only TransGUNet achieved a DSC over 45\% on AMOS-MRI (MRI modality) when trained on Synapse (CT modality). These results indicate that employing an attentional cross-scale GNN-based skip connection with non-ambiguous spatial attention is crucial for understanding complex anatomical structures in medical images. Fig. \ref{fig:EfficiencyAnalysis} indicates that TransGUNet contains almost 25.0M parameters with 10.0G FLOPs, which has apparent advantages regarding computational efficiency. We provide detailed number of parameters, FLOPs, and Inference Time (ms) for each model in Appendix (Tab. \ref{tab:efficiency_analysis}).

\begin{table}[t]
    \centering
    \scriptsize
    \setlength\tabcolsep{2pt} 
    \begin{tabular}{c|ccc|cc|cc|c|c}
    \hline
    Setting & \multicolumn{3}{c|}{GNN} & \multicolumn{2}{c|}{\textit{Seen}} & \multicolumn{2}{c|}{\textit{Unseen}} & \multicolumn{1}{c|}{\multirow{2}{*}{Param (M)}} & \multicolumn{1}{c}{\multirow{2}{*}{FLOPs (G)}} \\ \cline{2-8}
    Name & Single & Cross & NA & DSC & mIoU & DSC & mIoU &  & \\
    \hline
    S0 &  &  &  & 90.6 & 84.5 & 82.8 & 75.4 & 24.3M & 8.9G \\
    \hline
    S1 & \cmark &  &  & 92.1 & 87.3 & 82.4 & 74.5 & 24.4M & 9.3G \\
    S2 & \cmark &  & \cmark & \textit{92.3} & \textit{87.6} & 82.7 & 75.4 & 24.4M & 9.3G \\
    \hline
    S3 &  & \cmark &  & 92.2 & \textit{87.6} & \textit{83.5} & \textit{75.8} & 25.0M & 10.0G \\
    \textbf{S4 (Ours)} &  & \cmark & \cmark & \textbf{\underline{92.7}} &\textbf{\underline{88.1}} & \textbf{\underline{84.4}} & \textbf{\underline{76.8}} & 25.0M & 10.0G \\
    \hline
    \end{tabular}
    \caption{Ablation study of ACS-GNN in skip connection on \textit{Seen} and \textit{Unseen} polyp segmentation datasets. \lq Single\rq, \lq Cross\rq, and \lq NA\rq\, denote Single-Scale GNN, Cross-Scale GNN, and Node Attention, respectively.} 
    \label{tab:ablation_on_GNN}
\end{table}

Fig. \ref{fig:QualitativeResults} illustrates the qualitative results of the various models. UNet, CENet, MSRFNet, and DCSAUNet, which do not incorporate global dependencies or reduce the semantic gap between the encoder and decoder, produce noisy and unreliable predictions. Despite the advantage of UNet++ and M2SNet for reducing the semantic gap through nested convolution and subtraction units, respectively, their predictions remain for unreliable in colonoscopy images comprising complex polyp structures or ultrasounds containing severe noise. Although TransUNet uses global dependency, it is not employed in the decoder, resulting in inaccurate predictions. CFATransUNet improves upon the previous methods by reducing the semantic gap and incorporating transformer blocks at the decoding stage. However, it overlooks crucial spatial relationships and misses the essential fine and local details needed to interpret medical images effectively. Although PVT-GCASCADE mitigates these deficiencies to a certain extent, it fails to consider cross-scale interactions, reliable spatial attention and the importance of each node in the graph. Furthermore, as illustrated in Fig. \ref{fig:GraphVisualization}, TransGUNet can understand the relationships between semantically similar patches regardless of their distance. Consequently, despite the severe noise, lesions of various sizes, and complex anatomical structures in various modalities, TransGUNet produces reliable predictions due to the dual utilization of ACS-GNN and EFS-based spatial attention. More various qualitative results are available in the Appendix (Fig. \ref{fig:Sup_QualitativeResults_Dermatoscopy}, \ref{fig:Sup_QualitativeResults_Radiology}, \ref{fig:Sup_QualitativeResults_Ultrasound}, \ref{fig:Sup_QualitativeResults_Colonoscopy}).

\begin{figure}[t]
    \centering
    \includegraphics[width=0.48\textwidth]{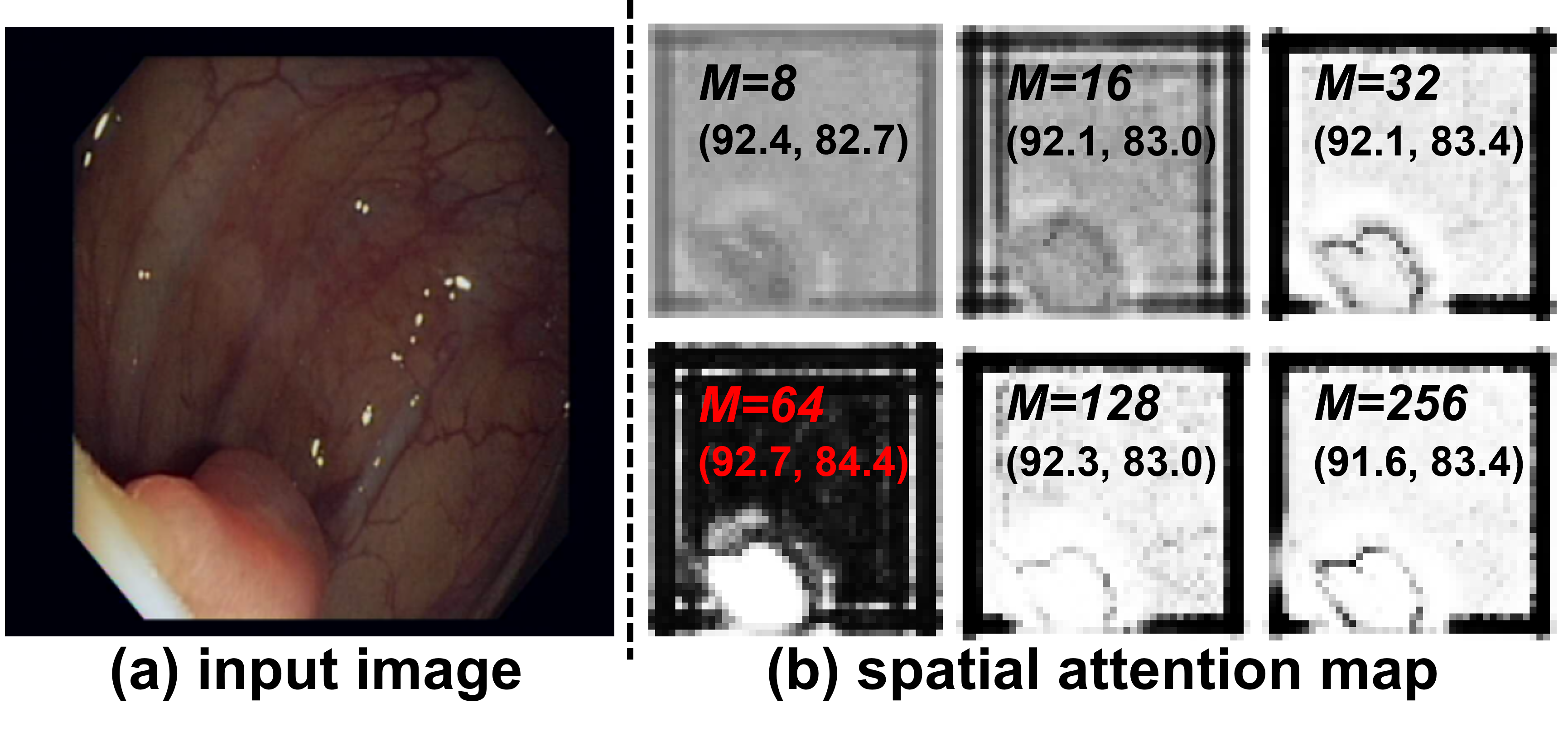}
    \caption{Comparison of spatial attention map quality according to various $M = \{ 8, 16, 32, 64, 128, 256 \}$. (a) Input image, (b) Spatial attention map according to various $M$.} \vspace{-0.25cm}
    \label{fig:USA_Ablation}
\end{figure}

\subsection{Ablation Study on TransGUNet}
We conducted ablation studies on polyp segmentation task to demonstrate the effectiveness of ACS-GNN and EFS-based spatial attention. We want to clarify that the experiment settings for the ablation study are identical to the main experiments for fair comparison.

\noindent \textbf{Ablation Study on ACS-GNN.} As listed in Tab. \ref{tab:ablation_on_GNN}, our approach (S4) exhibited the best performance across all settings. S0 represents TransGUNet without an ACS-GNN. The most notable result is that the application of skip connections in GNN, whether single-scale (S1) or cross-scale (S3), improves the performance on the seen dataset. However, on the unseen dataset, the single-scale GNN exhibits a 0.4\% performance decrease whereas the cross-scale GNN exhibits a 0.7\% performance improvement. These experimental results explain the reason behind TransGUNet outperforming PVT-GCASCADE. Additionally, NA improved the performance of both single-scale (S2) and cross-scale (S4) GNNs, with a negligible increase in the number of parameters and FLOPs owing to the ECA-style attention mechanism. Consequently, TransGUNet, which employs ACS-GNN, performs significantly better across various modalities and clinical settings.

\noindent \textbf{Ablation Study on EFS-based Spatial Attention.} As mentioned in the implementation details section, because we set $C_{r} = 64$, the total number of channels in the skip connection equals $4C_{r} = 256$. Therefore, when $4C_{r} = M$, where $M$ is the number of selected channels, all the feature maps are used to produce a spatial attention map. In Fig. \ref{fig:USA_Ablation}, $(\cdot, \cdot)$ denotes DSC on seen and unseen clinical setting of polyp segmentation. Fig. \ref{fig:USA_Ablation} shows that $M = 256$ results in worse performance than the cases utilizing the EFS-based spatial attention ($M = \{ 8, 16, 32, 64, 128\}$). These results demonstrate that our feature selection approach improves the quality of spatial attention, an aspect not previously addressed. If we select fewer channels ($M = \{ 8, 16, 32 \}$), uncertainty in the spatial attention map increases due to a lack of information.

\begin{table}[t]
    \centering
    \scriptsize
    \setlength\tabcolsep{4.0pt} 
    \begin{tabular}{c|cc|cc|c|c}
    \hline
    Target Resolution & \multicolumn{2}{c|}{\textit{Seen}}  & \multicolumn{2}{c|}{\textit{Unseen}} & \multicolumn{1}{c|}{\multirow{2}{*}{Param (M)}}  & \multicolumn{1}{c}{\multirow{2}{*}{FLOPs (G)}} \\ \cline{2-5}
    $(H_{t}, W_{t})$  & DSC & mIoU & DSC & mIoU & & \\ 
    \hline
    $(H / 4, W / 4)$                 & \textit{87.1} & \textit{80.5} & \textit{78.5} & \textbf{\underline{69.7}} & 25.0M & 13.6G \\
    $(H / 8, W / 8)$ \textbf{(Ours)} & \textbf{\underline{87.3}} & \textbf{\underline{80.6}} & \textbf{\underline{78.6}} & \textbf{\underline{69.7}} & 25.0M & 10.0G \\
    $(H / 16, W / 16)$               & 86.3 & 79.6 & 77.2 & \textit{68.2} & 25.0M & 9.2G \\
    $(H / 32, W / 32)$               & 86.2 & 79.4 & 76.6 & 67.7 & 25.0M & 9.0G \\
    \hline
    \end{tabular}
    \caption{Quantitative results for each \textit{Seen} and \textit{Unseen} datasets according to various target resolution $(H_{t}, W_{t})$.} \vspace{-0.25cm}
    \label{tab:ablation_target_resolution}
\end{table}

\noindent \textbf{Ablation Study on Target Resolution in ACS-GNN.} In this section, we conduct an ablation study to compare the performance and efficiency according to various target resolution $(H_{t}, W_{t}) = \{ (\frac{H}{4}, \frac{W}{4}), (\frac{H}{8}, \frac{W}{8}), (\frac{H}{16}, \frac{W}{16}), (\frac{H}{32}, \frac{W}{32}) \}$. Each resolution correspond to the resolutions of the feature maps $\mathbf{f}_{i} \in \mathbb{R}^{C_{i} \times \frac{H}{2^{i + 1}} \times \frac{W}{2^{i + 1}}}$ extracted from the Transformer encoder for $i = 1, 2, 3, 4$. We used $(\frac{H}{8}, \frac{W}{8})$ as a target resolution $(H_{t}, W_{t})$. As listed in Tab. \ref{tab:ablation_target_resolution}, the experimental results indicate that the resolution $(\frac{H}{8}, \frac{W}{8})$ achieved the best performance. Although $(\frac{H}{4}, \frac{W}{4})$ also demonstrated high performance, the efficiency significantly decreased due to the increased FLOPs associated with the higher resolution of the input feature maps and the inherent characteristics of GNNs. This indicates that while higher resolutions can capture more detailed information, they come at the cost of increased computational resources. 
\section{Conclusion}

In this study, we proposed TransGUNet, an innovative medical image segmentation model that integrates ACS-GNN and EFS-based spatial attention. The proposed model effectively addresses the limitations of the existing models by reducing the semantic gap between the encoder and decoder, preserving crucial information without ambiguous spatial attention, and leveraging global dependencies. Through extensive experiments on six seen and eight unseen datasets, TransGUNet demonstrated superior performance and higher efficiency than the state-of-the-art models. The results of ablation study demonstrate that incorporating GNNs into skip connection engineering significantly enhances the model's ability to capture and utilize complex anatomical features. Additionally, EFS ensures that only the most informative features are considered, thereby improving the quality of spatial attention maps. Consequently, TransGUNet represents a significant advancement in medical image segmentation, offering a robust, efficient, and accurate model that can be employed in various medical applications. In future work, we will focus on optimizing the memory efficiency and explore its deployment in real-world healthcare settings.

{
    \small
    \bibliographystyle{ieeenat_fullname}
    \bibliography{main}

\begin{thebibliography}{80}
\providecommand{\natexlab}[1]{#1}
\providecommand{\url}[1]{\texttt{#1}}
\expandafter\ifx\csname urlstyle\endcsname\relax
  \providecommand{\doi}[1]{doi: #1}\else
  \providecommand{\doi}{doi: \begingroup \urlstyle{rm}\Url}\fi

\bibitem[COV()]{COVID19_2}
Covid19 dataset.
\newblock \url{https://www.kaggle.com/datasets/piyushsamant11/pidata-new-names}.

\bibitem[Al-Dhabyani et~al.(2020)Al-Dhabyani, Gomaa, Khaled, and Fahmy]{al2020dataset}
Walid Al-Dhabyani, Mohammed Gomaa, Hussien Khaled, and Aly Fahmy.
\newblock Dataset of breast ultrasound images.
\newblock \emph{Data in brief}, 28:\penalty0 104863, 2020.

\bibitem[Bennett et~al.(2015)Bennett, Zhoubing, Juan, Martin, Thomas, and Arno]{Synapse_dataset}
Landman Bennett, Xu Zhoubing, Igelsias Juan, Styner Martin, Langerak Thomas, and Klein Arno.
\newblock Segmentation outside the cranial vault challenge.
\newblock In \emph{Medical Image Computing and Computer-Assisted Intervention--MICCAI 2015: 18th International Conference, Munich, Germany, October 5-9, 2015, Proceedings, Part III 18}. Springer, 2015.

\bibitem[Bernal et~al.(2015)Bernal, S{\'a}nchez, Fern{\'a}ndez-Esparrach, Gil, Rodr{\'\i}guez, and Vilari{\~n}o]{bernal2015wm}
Jorge Bernal, F~Javier S{\'a}nchez, Gloria Fern{\'a}ndez-Esparrach, Debora Gil, Cristina Rodr{\'\i}guez, and Fernando Vilari{\~n}o.
\newblock Wm-dova maps for accurate polyp highlighting in colonoscopy: Validation vs. saliency maps from physicians.
\newblock \emph{Computerized medical imaging and graphics}, 43:\penalty0 99--111, 2015.

\bibitem[Cao et~al.(2022)Cao, Wang, Chen, Jiang, Zhang, Tian, and Wang]{cao2022swin}
Hu Cao, Yueyue Wang, Joy Chen, Dongsheng Jiang, Xiaopeng Zhang, Qi Tian, and Manning Wang.
\newblock Swin-unet: Unet-like pure transformer for medical image segmentation.
\newblock In \emph{European conference on computer vision}, pages 205--218. Springer, 2022.

\bibitem[Celaya et~al.(2023)Celaya, Riviere, and Fuentes]{celaya2023generalized}
Adrian Celaya, Beatrice Riviere, and David Fuentes.
\newblock A generalized surface loss for reducing the hausdorff distance in medical imaging segmentation.
\newblock \emph{arXiv preprint arXiv:2302.03868}, 2023.

\bibitem[Chen et~al.(2021{\natexlab{a}})Chen, Lu, Yu, Luo, Adeli, Wang, Lu, Yuille, and Zhou]{chen2021transunet}
Jieneng Chen, Yongyi Lu, Qihang Yu, Xiangde Luo, Ehsan Adeli, Yan Wang, Le Lu, Alan~L Yuille, and Yuyin Zhou.
\newblock Transunet: Transformers make strong encoders for medical image segmentation.
\newblock \emph{arXiv preprint arXiv:2102.04306}, 2021{\natexlab{a}}.

\bibitem[Chen et~al.(2021{\natexlab{b}})Chen, Long, Ren, Sun, and Pu]{chen2021lesion}
Kecheng Chen, Kun Long, Yazhou Ren, Jiayu Sun, and Xiaorong Pu.
\newblock Lesion-inspired denoising network: Connecting medical image denoising and lesion detection.
\newblock In \emph{Proceedings of the 29th ACM International Conference on Multimedia}, pages 3283--3292, 2021{\natexlab{b}}.

\bibitem[Coates et~al.(2015)Coates, Winer, Goldhirsch, Gelber, Gnant, Piccart-Gebhart, Th{\"u}rlimann, Senn, Members, Andr{\'e}, et~al.]{coates2015tailoring}
Alan~S Coates, Eric~P Winer, Aron Goldhirsch, Richard~D Gelber, Michael Gnant, M Piccart-Gebhart, Beat Th{\"u}rlimann, H-J Senn, Panel Members, Fabrice Andr{\'e}, et~al.
\newblock Tailoring therapies—improving the management of early breast cancer: St gallen international expert consensus on the primary therapy of early breast cancer 2015.
\newblock \emph{Annals of oncology}, 26\penalty0 (8):\penalty0 1533--1546, 2015.

\bibitem[Dong et~al.()Dong, Wang, Fan, Li, Fu, and Shao]{dong2108polyp}
Bo Dong, Wenhai Wang, Deng-Ping Fan, Jinpeng Li, Huazhu Fu, and Ling Shao.
\newblock Polyp-pvt: Polyp segmentation with pyramid vision transformers. arxiv 2021.
\newblock \emph{arXiv preprint arXiv:2108.06932}.

\bibitem[Dosovitskiy et~al.(2021)Dosovitskiy, Beyer, Kolesnikov, Weissenborn, Zhai, Unterthiner, Dehghani, Minderer, Heigold, Gelly, Uszkoreit, and Houlsby]{dosovitskiy2021an}
Alexey Dosovitskiy, Lucas Beyer, Alexander Kolesnikov, Dirk Weissenborn, Xiaohua Zhai, Thomas Unterthiner, Mostafa Dehghani, Matthias Minderer, Georg Heigold, Sylvain Gelly, Jakob Uszkoreit, and Neil Houlsby.
\newblock An image is worth 16x16 words: Transformers for image recognition at scale.
\newblock In \emph{International Conference on Learning Representations}, 2021.

\bibitem[Duan and Chen(2023)]{duan20233d}
Fan Duan and Li Chen.
\newblock 3d dental mesh segmentation using semantics-based feature learning with graph-transformer.
\newblock In \emph{International Conference on Medical Image Computing and Computer-Assisted Intervention}, pages 456--465. Springer, 2023.

\bibitem[Fan et~al.(2017)Fan, Cheng, Liu, Li, and Borji]{fan2017structure}
Deng-Ping Fan, Ming-Ming Cheng, Yun Liu, Tao Li, and Ali Borji.
\newblock Structure-measure: A new way to evaluate foreground maps.
\newblock In \emph{Proceedings of the IEEE international conference on computer vision}, pages 4548--4557, 2017.

\bibitem[Fan et~al.(2018)Fan, Gong, Cao, Ren, Cheng, and Borji]{fan2018enhanced}
Deng-Ping Fan, Cheng Gong, Yang Cao, Bo Ren, Ming-Ming Cheng, and Ali Borji.
\newblock Enhanced-alignment measure for binary foreground map evaluation.
\newblock \emph{arXiv preprint arXiv:1805.10421}, 2018.

\bibitem[Fan et~al.(2020)Fan, Ji, Zhou, Chen, Fu, Shen, and Shao]{fan2020pranet}
Deng-Ping Fan, Ge-Peng Ji, Tao Zhou, Geng Chen, Huazhu Fu, Jianbing Shen, and Ling Shao.
\newblock Pranet: Parallel reverse attention network for polyp segmentation.
\newblock In \emph{International conference on medical image computing and computer-assisted intervention}, pages 263--273. Springer, 2020.

\bibitem[Gao et~al.(2019)Gao, Cheng, Zhao, Zhang, Yang, and Torr]{gao2019res2net}
Shang-Hua Gao, Ming-Ming Cheng, Kai Zhao, Xin-Yu Zhang, Ming-Hsuan Yang, and Philip Torr.
\newblock Res2net: A new multi-scale backbone architecture.
\newblock \emph{IEEE transactions on pattern analysis and machine intelligence}, 43\penalty0 (2):\penalty0 652--662, 2019.

\bibitem[Gao et~al.(2021)Gao, Zhou, and Metaxas]{gao2021utnet}
Yunhe Gao, Mu Zhou, and Dimitris~N Metaxas.
\newblock Utnet: a hybrid transformer architecture for medical image segmentation.
\newblock In \emph{Medical Image Computing and Computer Assisted Intervention--MICCAI 2021: 24th International Conference, Strasbourg, France, September 27--October 1, 2021, Proceedings, Part III 24}, pages 61--71. Springer, 2021.

\bibitem[Gaube et~al.(2021)Gaube, Suresh, Raue, Merritt, Berkowitz, Lermer, Coughlin, Guttag, Colak, and Ghassemi]{gaube2021ai}
Susanne Gaube, Harini Suresh, Martina Raue, Alexander Merritt, Seth~J Berkowitz, Eva Lermer, Joseph~F Coughlin, John~V Guttag, Errol Colak, and Marzyeh Ghassemi.
\newblock Do as ai say: susceptibility in deployment of clinical decision-aids.
\newblock \emph{NPJ digital medicine}, 4\penalty0 (1):\penalty0 31, 2021.

\bibitem[Gu et~al.(2019)Gu, Cheng, Fu, Zhou, Hao, Zhao, Zhang, Gao, and Liu]{gu2019net}
Zaiwang Gu, Jun Cheng, Huazhu Fu, Kang Zhou, Huaying Hao, Yitian Zhao, Tianyang Zhang, Shenghua Gao, and Jiang Liu.
\newblock Ce-net: Context encoder network for 2d medical image segmentation.
\newblock \emph{IEEE transactions on medical imaging}, 38\penalty0 (10):\penalty0 2281--2292, 2019.

\bibitem[Gutman et~al.(2016)Gutman, Codella, Celebi, Helba, Marchetti, Mishra, and Halpern]{gutman2016skin}
David Gutman, Noel~CF Codella, Emre Celebi, Brian Helba, Michael Marchetti, Nabin Mishra, and Allan Halpern.
\newblock Skin lesion analysis toward melanoma detection: A challenge at the international symposium on biomedical imaging (isbi) 2016, hosted by the international skin imaging collaboration (isic).
\newblock \emph{arXiv preprint arXiv:1605.01397}, 2016.

\bibitem[Han et~al.(2022)Han, Wang, Guo, Tang, and Wu]{han2022vision}
Kai Han, Yunhe Wang, Jianyuan Guo, Yehui Tang, and Enhua Wu.
\newblock Vision gnn: An image is worth graph of nodes.
\newblock \emph{Advances in neural information processing systems}, 35:\penalty0 8291--8303, 2022.

\bibitem[Han et~al.(2023)Han, Wang, Kundu, Ding, and Wang]{han2023vision}
Yan Han, Peihao Wang, Souvik Kundu, Ying Ding, and Zhangyang Wang.
\newblock Vision hgnn: An image is more than a graph of nodes.
\newblock In \emph{Proceedings of the IEEE/CVF International Conference on Computer Vision}, pages 19878--19888, 2023.

\bibitem[Haralick et~al.(1987)Haralick, Sternberg, and Zhuang]{haralick1987image}
Robert~M Haralick, Stanley~R Sternberg, and Xinhua Zhuang.
\newblock Image analysis using mathematical morphology.
\newblock \emph{IEEE transactions on pattern analysis and machine intelligence}, \penalty0 (4):\penalty0 532--550, 1987.

\bibitem[Hatamizadeh et~al.(2022)Hatamizadeh, Tang, Nath, Yang, Myronenko, Landman, Roth, and Xu]{hatamizadeh2022unetr}
Ali Hatamizadeh, Yucheng Tang, Vishwesh Nath, Dong Yang, Andriy Myronenko, Bennett Landman, Holger~R Roth, and Daguang Xu.
\newblock Unetr: Transformers for 3d medical image segmentation.
\newblock In \emph{Proceedings of the IEEE/CVF winter conference on applications of computer vision}, pages 574--584, 2022.

\bibitem[He et~al.(2016)He, Zhang, Ren, and Sun]{he2016deep}
Kaiming He, Xiangyu Zhang, Shaoqing Ren, and Jian Sun.
\newblock Deep residual learning for image recognition.
\newblock In \emph{Proceedings of the IEEE conference on computer vision and pattern recognition}, pages 770--778, 2016.

\bibitem[Heidari et~al.(2023)Heidari, Kazerouni, Soltany, Azad, Aghdam, Cohen-Adad, and Merhof]{heidari2023hiformer}
Moein Heidari, Amirhossein Kazerouni, Milad Soltany, Reza Azad, Ehsan~Khodapanah Aghdam, Julien Cohen-Adad, and Dorit Merhof.
\newblock Hiformer: Hierarchical multi-scale representations using transformers for medical image segmentation.
\newblock In \emph{Proceedings of the IEEE/CVF Winter Conference on Applications of Computer Vision}, pages 6202--6212, 2023.

\bibitem[Jha et~al.(2020)Jha, Smedsrud, Riegler, Halvorsen, de~Lange, Johansen, and Johansen]{jha2020kvasir}
Debesh Jha, Pia~H Smedsrud, Michael~A Riegler, P{\aa}l Halvorsen, Thomas de Lange, Dag Johansen, and H{\aa}vard~D Johansen.
\newblock Kvasir-seg: A segmented polyp dataset.
\newblock In \emph{MultiMedia Modeling: 26th International Conference, MMM 2020, Daejeon, South Korea, January 5--8, 2020, Proceedings, Part II 26}, pages 451--462. Springer, 2020.

\bibitem[Ji et~al.(2022)Ji, Bai, Ge, Yang, Zhu, Zhang, Li, Zhanng, Ma, Wan, et~al.]{ji2022amos}
Yuanfeng Ji, Haotian Bai, Chongjian Ge, Jie Yang, Ye Zhu, Ruimao Zhang, Zhen Li, Lingyan Zhanng, Wanling Ma, Xiang Wan, et~al.
\newblock Amos: A large-scale abdominal multi-organ benchmark for versatile medical image segmentation.
\newblock \emph{Advances in neural information processing systems}, 35:\penalty0 36722--36732, 2022.

\bibitem[Jiang et~al.(2023)Jiang, Chen, Tian, and Liu]{jiang2023vig}
Juntao Jiang, Xiyu Chen, Guanzhong Tian, and Yong Liu.
\newblock Vig-unet: vision graph neural networks for medical image segmentation.
\newblock In \emph{2023 IEEE 20th International Symposium on Biomedical Imaging (ISBI)}, pages 1--5. IEEE, 2023.

\bibitem[Jun et~al.(2020)Jun, Cheng, Yixin, Xingle, Jiantao, Ziqi, Minqing, Xin, Xueyuan, Shucheng, Hao, Sen, Xiaoyu, Ziwei, Chen, Lu, Yuntao, Qiongjie, Guoqiang, and Jian]{ma_jun_2020_3757476}
Ma Jun, Ge Cheng, Wang Yixin, An Xingle, Gao Jiantao, Yu Ziqi, Zhang Minqing, Liu Xin, Deng Xueyuan, Cao Shucheng, Wei Hao, Mei Sen, Yang Xiaoyu, Nie Ziwei, Li Chen, Tian Lu, Zhu Yuntao, Zhu Qiongjie, Dong Guoqiang, and He Jian.
\newblock {COVID-19 CT Lung and Infection Segmentation Dataset}.
\newblock 2020.

\bibitem[Kanopoulos et~al.(1988)Kanopoulos, Vasanthavada, and Baker]{kanopoulos1988design}
Nick Kanopoulos, Nagesh Vasanthavada, and Robert~L Baker.
\newblock Design of an image edge detection filter using the sobel operator.
\newblock \emph{IEEE Journal of solid-state circuits}, 23\penalty0 (2):\penalty0 358--367, 1988.

\bibitem[Kass et~al.(1988)Kass, Witkin, and Terzopoulos]{kass1988snakes}
Michael Kass, Andrew Witkin, and Demetri Terzopoulos.
\newblock Snakes: Active contour models.
\newblock \emph{International journal of computer vision}, 1\penalty0 (4):\penalty0 321--331, 1988.

\bibitem[Kipf and Welling(2016)]{kipf2016semi}
Thomas~N Kipf and Max Welling.
\newblock Semi-supervised classification with graph convolutional networks.
\newblock \emph{arXiv preprint arXiv:1609.02907}, 2016.

\bibitem[Li et~al.(2019)Li, Muller, Thabet, and Ghanem]{li2019deepgcns}
Guohao Li, Matthias Muller, Ali Thabet, and Bernard Ghanem.
\newblock Deepgcns: Can gcns go as deep as cnns?
\newblock In \emph{Proceedings of the IEEE/CVF international conference on computer vision}, pages 9267--9276, 2019.

\bibitem[Li et~al.(2018)Li, Han, and Wu]{li2018deeper}
Qimai Li, Zhichao Han, and Xiao-Ming Wu.
\newblock Deeper insights into graph convolutional networks for semi-supervised learning.
\newblock In \emph{Proceedings of the AAAI conference on artificial intelligence}, 2018.

\bibitem[Li et~al.(2025)Li, Fu, Wang, Zhang, Ye, and Ma]{li2025gse}
Xiang Li, Chong Fu, Qun Wang, Wenchao Zhang, Chen Ye, and Tao Ma.
\newblock Gse-nets: Global structure enhancement decoder for thyroid nodule segmentation.
\newblock \emph{Biomedical Signal Processing and Control}, 102:\penalty0 107340, 2025.

\bibitem[Liu et~al.(2024)Liu, Yao, Liu, Chang, Chen, Wang, and Wei]{liu2024cafe}
Guoqi Liu, Sheng Yao, Dong Liu, Baofang Chang, Zongyu Chen, Jiajia Wang, and Jiangqi Wei.
\newblock Cafe-net: Cross-attention and feature exploration network for polyp segmentation.
\newblock \emph{Expert Systems with Applications}, 238:\penalty0 121754, 2024.

\bibitem[Mahmud et~al.(2021)Mahmud, Rahman, Fattah, and Kung]{mahmud2021covsegnet}
Tanvir Mahmud, Md~Awsafur Rahman, Shaikh~Anowarul Fattah, and Sun-Yuan Kung.
\newblock Covsegnet: A multi encoder--decoder architecture for improved lesion segmentation of covid-19 chest ct scans.
\newblock \emph{IEEE Transactions on Artificial Intelligence}, 2\penalty0 (3):\penalty0 283--297, 2021.

\bibitem[Majaj et~al.(2015)Majaj, Hong, Solomon, and DiCarlo]{majaj2015simple}
Najib~J Majaj, Ha Hong, Ethan~A Solomon, and James~J DiCarlo.
\newblock Simple learned weighted sums of inferior temporal neuronal firing rates accurately predict human core object recognition performance.
\newblock \emph{Journal of Neuroscience}, 35\penalty0 (39):\penalty0 13402--13418, 2015.

\bibitem[Margolin et~al.(2014)Margolin, Zelnik-Manor, and Tal]{margolin2014evaluate}
Ran Margolin, Lihi Zelnik-Manor, and Ayellet Tal.
\newblock How to evaluate foreground maps?
\newblock In \emph{Proceedings of the IEEE conference on computer vision and pattern recognition}, pages 248--255, 2014.

\bibitem[Marr(2010)]{marr2010vision}
David Marr.
\newblock \emph{Vision: A computational investigation into the human representation and processing of visual information}.
\newblock MIT press, 2010.

\bibitem[Mendon{\c{c}}a et~al.(2013)Mendon{\c{c}}a, Ferreira, Marques, Marcal, and Rozeira]{mendoncca2013ph}
Teresa Mendon{\c{c}}a, Pedro~M Ferreira, Jorge~S Marques, Andr{\'e}~RS Marcal, and Jorge Rozeira.
\newblock Ph 2-a dermoscopic image database for research and benchmarking.
\newblock In \emph{2013 35th annual international conference of the IEEE engineering in medicine and biology society (EMBC)}, pages 5437--5440. IEEE, 2013.

\bibitem[Milletari et~al.(2016)Milletari, Navab, and Ahmadi]{milletari2016v}
Fausto Milletari, Nassir Navab, and Seyed-Ahmad Ahmadi.
\newblock V-net: Fully convolutional neural networks for volumetric medical image segmentation.
\newblock In \emph{2016 fourth international conference on 3D vision (3DV)}, pages 565--571. Ieee, 2016.

\bibitem[Nam et~al.(2023)Nam, Park, Syazwany, Jung, Im, and Lee]{nam2023m3fpolypsegnet}
Ju-Hyeon Nam, Seo-Hyeong Park, Nur~Suriza Syazwany, Yerim Jung, Yu-Han Im, and Sang-Chul Lee.
\newblock M3fpolypsegnet: Segmentation network with multi-frequency feature fusion for polyp localization in colonoscopy images.
\newblock In \emph{2023 IEEE International Conference on Image Processing (ICIP)}, pages 1530--1534. IEEE, 2023.

\bibitem[Nam et~al.(2024)Nam, Syazwany, Kim, and Lee]{nam2024modality}
Ju-Hyeon Nam, Nur~Suriza Syazwany, Su~Jung Kim, and Sang-Chul Lee.
\newblock Modality-agnostic domain generalizable medical image segmentation by multi-frequency in multi-scale attention.
\newblock In \emph{Proceedings of the IEEE/CVF Conference on Computer Vision and Pattern Recognition}, pages 11480--11491, 2024.

\bibitem[Oono and Suzuki(2020)]{oono2020graph}
Kenta Oono and Taiji Suzuki.
\newblock Graph neural networks exponentially lose expressive power for node classification.
\newblock In \emph{International Conference on Learning Representations}, 2020.

\bibitem[Otsu(1979)]{otsu1979threshold}
Nobuyuki Otsu.
\newblock A threshold selection method from gray-level histograms.
\newblock \emph{IEEE transactions on systems, man, and cybernetics}, 9\penalty0 (1):\penalty0 62--66, 1979.

\bibitem[Palmer(1999)]{palmer1999vision}
Stephen~E Palmer.
\newblock \emph{Vision science: Photons to phenomenology}.
\newblock MIT press, 1999.

\bibitem[Posner et~al.(1990)Posner, Petersen, et~al.]{posner1990attention}
Michael~I Posner, Steven~E Petersen, et~al.
\newblock The attention system of the human brain.
\newblock \emph{Annual review of neuroscience}, 13\penalty0 (1):\penalty0 25--42, 1990.

\bibitem[Rahman and Marculescu(2024)]{rahman2024g}
Md~Mostafijur Rahman and Radu Marculescu.
\newblock G-cascade: Efficient cascaded graph convolutional decoding for 2d medical image segmentation.
\newblock In \emph{Proceedings of the IEEE/CVF Winter Conference on Applications of Computer Vision}, pages 7728--7737, 2024.

\bibitem[Riccio et~al.(2018)Riccio, Brancati, Frucci, and Gragnaniello]{riccio2018new}
Daniel Riccio, Nadia Brancati, Maria Frucci, and Diego Gragnaniello.
\newblock A new unsupervised approach for segmenting and counting cells in high-throughput microscopy image sets.
\newblock \emph{IEEE journal of biomedical and health informatics}, 23\penalty0 (1):\penalty0 437--448, 2018.

\bibitem[Ronneberger et~al.(2015)Ronneberger, Fischer, and Brox]{ronneberger2015u}
Olaf Ronneberger, Philipp Fischer, and Thomas Brox.
\newblock U-net: Convolutional networks for biomedical image segmentation.
\newblock In \emph{Medical Image Computing and Computer-Assisted Intervention--MICCAI 2015: 18th International Conference, Munich, Germany, October 5-9, 2015, Proceedings, Part III 18}, pages 234--241. Springer, 2015.

\bibitem[Shawn et~al.(2024)Shawn, Chyrikov, Lanet, Chen, Zhao, and Chajo]{shawn2024ct}
Helena Shawn, Thompson Chyrikov, Jacob Lanet, Lam-chi Chen, Jim Zhao, and Christina Chajo.
\newblock A ct image denoising method with residual encoder-decoder network.
\newblock \emph{arXiv preprint arXiv:2404.01553}, 2024.

\bibitem[Silva et~al.(2014)Silva, Histace, Romain, Dray, and Granado]{silva2014toward}
Juan Silva, Aymeric Histace, Olivier Romain, Xavier Dray, and Bertrand Granado.
\newblock Toward embedded detection of polyps in wce images for early diagnosis of colorectal cancer.
\newblock \emph{International journal of computer assisted radiology and surgery}, 9:\penalty0 283--293, 2014.

\bibitem[Srivastava et~al.(2021)Srivastava, Jha, Chanda, Pal, Johansen, Johansen, Riegler, Ali, and Halvorsen]{srivastava2021msrf}
Abhishek Srivastava, Debesh Jha, Sukalpa Chanda, Umapada Pal, H{\aa}vard~D Johansen, Dag Johansen, Michael~A Riegler, Sharib Ali, and P{\aa}l Halvorsen.
\newblock Msrf-net: a multi-scale residual fusion network for biomedical image segmentation.
\newblock \emph{IEEE Journal of Biomedical and Health Informatics}, 26\penalty0 (5):\penalty0 2252--2263, 2021.

\bibitem[Tajbakhsh et~al.(2015)Tajbakhsh, Gurudu, and Liang]{tajbakhsh2015automated}
Nima Tajbakhsh, Suryakanth~R Gurudu, and Jianming Liang.
\newblock Automated polyp detection in colonoscopy videos using shape and context information.
\newblock \emph{IEEE transactions on medical imaging}, 35\penalty0 (2):\penalty0 630--644, 2015.

\bibitem[Tizhoosh(2005)]{tizhoosh2005image}
Hamid~R Tizhoosh.
\newblock Image thresholding using type ii fuzzy sets.
\newblock \emph{Pattern recognition}, 38\penalty0 (12):\penalty0 2363--2372, 2005.

\bibitem[Tragakis et~al.(2023)Tragakis, Kaul, Murray-Smith, and Husmeier]{tragakis2023fully}
Athanasios Tragakis, Chaitanya Kaul, Roderick Murray-Smith, and Dirk Husmeier.
\newblock The fully convolutional transformer for medical image segmentation.
\newblock In \emph{Proceedings of the IEEE/CVF Winter Conference on Applications of Computer Vision}, pages 3660--3669, 2023.

\bibitem[Treue(2001)]{treue2001neural}
Stefan Treue.
\newblock Neural correlates of attention in primate visual cortex.
\newblock \emph{Trends in neurosciences}, 24\penalty0 (5):\penalty0 295--300, 2001.

\bibitem[V{\'a}zquez et~al.(2017)V{\'a}zquez, Bernal, S{\'a}nchez, Fern{\'a}ndez-Esparrach, L{\'o}pez, Romero, Drozdzal, Courville, et~al.]{vazquez2017benchmark}
David V{\'a}zquez, Jorge Bernal, F~Javier S{\'a}nchez, Gloria Fern{\'a}ndez-Esparrach, Antonio~M L{\'o}pez, Adriana Romero, Michal Drozdzal, Aaron Courville, et~al.
\newblock A benchmark for endoluminal scene segmentation of colonoscopy images.
\newblock \emph{Journal of healthcare engineering}, 2017, 2017.

\bibitem[Vu et~al.(2019)Vu, Jain, Bucher, Cord, and P{\'e}rez]{vu2019advent}
Tuan-Hung Vu, Himalaya Jain, Maxime Bucher, Matthieu Cord, and Patrick P{\'e}rez.
\newblock Advent: Adversarial entropy minimization for domain adaptation in semantic segmentation.
\newblock In \emph{Proceedings of the IEEE/CVF conference on computer vision and pattern recognition}, pages 2517--2526, 2019.

\bibitem[Wang et~al.(2024{\natexlab{a}})Wang, Pan, Aboah, Zhang, Keles, Torigian, Turkbey, Krupinski, Udupa, and Bagci]{wang2024gazegnn}
Bin Wang, Hongyi Pan, Armstrong Aboah, Zheyuan Zhang, Elif Keles, Drew Torigian, Baris Turkbey, Elizabeth Krupinski, Jayaram Udupa, and Ulas Bagci.
\newblock Gazegnn: A gaze-guided graph neural network for chest x-ray classification.
\newblock In \emph{Proceedings of the IEEE/CVF Winter Conference on Applications of Computer Vision}, pages 2194--2203, 2024{\natexlab{a}}.

\bibitem[Wang et~al.(2024{\natexlab{b}})Wang, Wang, Wang, Wei, Feng, Wu, Yao, and Zhang]{wang2024cfatransunet}
Cheng Wang, Le Wang, Nuoqi Wang, Xiaoling Wei, Ting Feng, Minfeng Wu, Qi Yao, and Rongjun Zhang.
\newblock Cfatransunet: Channel-wise cross fusion attention and transformer for 2d medical image segmentation.
\newblock \emph{Computers in Biology and Medicine}, 168:\penalty0 107803, 2024{\natexlab{b}}.

\bibitem[Wang et~al.(2022{\natexlab{a}})Wang, Cao, Wang, and Zaiane]{wang2022uctransnet}
Haonan Wang, Peng Cao, Jiaqi Wang, and Osmar~R Zaiane.
\newblock Uctransnet: rethinking the skip connections in u-net from a channel-wise perspective with transformer.
\newblock In \emph{Proceedings of the AAAI conference on artificial intelligence}, pages 2441--2449, 2022{\natexlab{a}}.

\bibitem[Wang et~al.(2020)Wang, Wu, Zhu, Li, Zuo, and Hu]{wang2020eca}
Qilong Wang, Banggu Wu, Pengfei Zhu, Peihua Li, Wangmeng Zuo, and Qinghua Hu.
\newblock Eca-net: Efficient channel attention for deep convolutional neural networks.
\newblock In \emph{Proceedings of the IEEE/CVF conference on computer vision and pattern recognition}, pages 11534--11542, 2020.

\bibitem[Wang et~al.(2019)Wang, Yu, Li, Yang, Fu, and Heng]{wang2019boundary}
Shujun Wang, Lequan Yu, Kang Li, Xin Yang, Chi-Wing Fu, and Pheng-Ann Heng.
\newblock Boundary and entropy-driven adversarial learning for fundus image segmentation.
\newblock In \emph{Medical Image Computing and Computer Assisted Intervention--MICCAI 2019: 22nd International Conference, Shenzhen, China, October 13--17, 2019, Proceedings, Part I 22}, pages 102--110. Springer, 2019.

\bibitem[Wang et~al.(2022{\natexlab{b}})Wang, Xie, Li, Fan, Song, Liang, Lu, Luo, and Shao]{wang2022pvt}
Wenhai Wang, Enze Xie, Xiang Li, Deng-Ping Fan, Kaitao Song, Ding Liang, Tong Lu, Ping Luo, and Ling Shao.
\newblock Pvt v2: Improved baselines with pyramid vision transformer.
\newblock \emph{Computational Visual Media}, 8\penalty0 (3):\penalty0 415--424, 2022{\natexlab{b}}.

\bibitem[Wang et~al.(2023)Wang, Yu, Chen, Hu, and Peng]{wang2023dynamic}
Yuan Wang, Kun Yu, Chen Chen, Xiyuan Hu, and Silong Peng.
\newblock Dynamic graph learning with content-guided spatial-frequency relation reasoning for deepfake detection.
\newblock In \emph{Proceedings of the IEEE/CVF Conference on Computer Vision and Pattern Recognition}, pages 7278--7287, 2023.

\bibitem[Wang et~al.(2024{\natexlab{c}})Wang, Guo, Zhao, Zhang, Zhao, Fang, Wang, Lu, Yu, and Tian]{wang2024multi}
Zhichao Wang, Lin Guo, Shuchang Zhao, Shiqing Zhang, Xiaoming Zhao, Jiangxiong Fang, Guoyu Wang, Hongsheng Lu, Jun Yu, and Qi Tian.
\newblock Multi-scale group agent attention-based graph convolutional decoding networks for 2d medical image segmentation.
\newblock \emph{IEEE Journal of Biomedical and Health Informatics}, 2024{\natexlab{c}}.

\bibitem[Wu et~al.(2022)Wu, Liu, Zhan, and Cheng]{wu2022p2t}
Yu-Huan Wu, Yun Liu, Xin Zhan, and Ming-Ming Cheng.
\newblock P2t: Pyramid pooling transformer for scene understanding.
\newblock \emph{IEEE transactions on pattern analysis and machine intelligence}, 2022.

\bibitem[Xu et~al.(2023{\natexlab{a}})Xu, Ma, Na, and Duan]{xu2023dcsau}
Qing Xu, Zhicheng Ma, HE Na, and Wenting Duan.
\newblock Dcsau-net: A deeper and more compact split-attention u-net for medical image segmentation.
\newblock \emph{Computers in Biology and Medicine}, 154:\penalty0 106626, 2023{\natexlab{a}}.

\bibitem[Xu et~al.(2023{\natexlab{b}})Xu, Duan, Zhang, Zhang, Sun, and Tian]{xu2023graph}
Shanshan Xu, Lianhong Duan, Yang Zhang, Zhicheng Zhang, Tiansheng Sun, and Lixia Tian.
\newblock Graph-and transformer-guided boundary aware network for medical image segmentation.
\newblock \emph{Computer Methods and Programs in Biomedicine}, 242:\penalty0 107849, 2023{\natexlab{b}}.

\bibitem[Ying et~al.(2018)Ying, He, Chen, Eksombatchai, Hamilton, and Leskovec]{ying2018graph}
Rex Ying, Ruining He, Kaifeng Chen, Pong Eksombatchai, William~L Hamilton, and Jure Leskovec.
\newblock Graph convolutional neural networks for web-scale recommender systems.
\newblock In \emph{Proceedings of the 24th ACM SIGKDD international conference on knowledge discovery \& data mining}, pages 974--983, 2018.

\bibitem[Zhang et~al.(2022{\natexlab{a}})Zhang, Wu, Zhang, Zhu, Lin, Zhang, Sun, He, Mueller, Manmatha, et~al.]{zhang2022resnest}
Hang Zhang, Chongruo Wu, Zhongyue Zhang, Yi Zhu, Haibin Lin, Zhi Zhang, Yue Sun, Tong He, Jonas Mueller, R Manmatha, et~al.
\newblock Resnest: Split-attention networks.
\newblock In \emph{Proceedings of the IEEE/CVF conference on computer vision and pattern recognition}, pages 2736--2746, 2022{\natexlab{a}}.

\bibitem[Zhang et~al.(2025)Zhang, Ye, Chen, Yu, and Cheng]{zhang2025transgraphnet}
Ju Zhang, Zhiyi Ye, Mingyang Chen, Jiahao Yu, and Yun Cheng.
\newblock Transgraphnet: A novel network for medical image segmentation based on transformer and graph convolution.
\newblock \emph{Biomedical Signal Processing and Control}, 104:\penalty0 107510, 2025.

\bibitem[Zhang et~al.(2022{\natexlab{b}})Zhang, Fu, Zheng, Zhang, Zhao, and Sham]{zhang2022hsnet}
Wenchao Zhang, Chong Fu, Yu Zheng, Fangyuan Zhang, Yanli Zhao, and Chiu-Wing Sham.
\newblock Hsnet: A hybrid semantic network for polyp segmentation.
\newblock \emph{Computers in biology and medicine}, 150:\penalty0 106173, 2022{\natexlab{b}}.

\bibitem[Zhao et~al.(2021)Zhao, Zhang, and Lu]{zhao2021automatic}
Xiaoqi Zhao, Lihe Zhang, and Huchuan Lu.
\newblock Automatic polyp segmentation via multi-scale subtraction network.
\newblock In \emph{Medical Image Computing and Computer Assisted Intervention--MICCAI 2021: 24th International Conference, Strasbourg, France, September 27--October 1, 2021, Proceedings, Part I 24}, pages 120--130. Springer, 2021.

\bibitem[Zhao et~al.(2023)Zhao, Jia, Pang, Lv, Tian, Zhang, Sun, and Lu]{zhao2023m}
Xiaoqi Zhao, Hongpeng Jia, Youwei Pang, Long Lv, Feng Tian, Lihe Zhang, Weibing Sun, and Huchuan Lu.
\newblock $\text{M}^{2}\text{SNet}$: Multi-scale in multi-scale subtraction network for medical image segmentation.
\newblock \emph{arXiv preprint arXiv:2303.10894}, 2023.

\bibitem[Zhou et~al.(2018)Zhou, Rahman~Siddiquee, Tajbakhsh, and Liang]{zhou2018unet++}
Zongwei Zhou, Md~Mahfuzur Rahman~Siddiquee, Nima Tajbakhsh, and Jianming Liang.
\newblock Unet++: A nested u-net architecture for medical image segmentation.
\newblock In \emph{Deep Learning in Medical Image Analysis and Multimodal Learning for Clinical Decision Support: 4th International Workshop, DLMIA 2018, and 8th International Workshop, ML-CDS 2018, Held in Conjunction with MICCAI 2018, Granada, Spain, September 20, 2018, Proceedings 4}, pages 3--11. Springer, 2018.

\bibitem[Zhuang et~al.(2019)Zhuang, Li, Joseph~Raj, Mahesh, and Qiu]{zhuang2019rdau}
Zhemin Zhuang, Nan Li, Alex~Noel Joseph~Raj, Vijayalakshmi~GV Mahesh, and Shunmin Qiu.
\newblock An rdau-net model for lesion segmentation in breast ultrasound images.
\newblock \emph{PloS one}, 14\penalty0 (8):\penalty0 e0221535, 2019.

\end{thebibliography}
}
\clearpage
\setcounter{page}{1}
\maketitlesupplementary

\section{Dataset Descriptions}
\label{appendix_dataset_descriptions}

\begin{table}[h]
    \centering
    \scriptsize
    \setlength\tabcolsep{2pt} 
    \begin{tabular}{c|c|cccc}
    \hline
    Segmentation Task &  Dataset & Resolutions       & Train & Test \\
    \hline
    Multi-organ Segmentation      & Synapse   & 512 $\times$ 512        & 18 Scans & 12 Scans \\ 
    \hline
    Skin Cancer Segmentation      & ISIC2018   & Variable          & 1868     & 261 \\ 
    \hline
    COVID19 Infection Segmentation & COVID19-1   & 512 $\times$ 512  & 643      & 383 \\
    \hline
    Breast Cancer Segmentation    & BUSI    & Variable          & 324      & 161 \\
    \hline
    \multicolumn{1}{c|}{\multirow{2}{*}{Polyp Segmentation}} & CVC-ClinicDB    & 384 $\times$ 288          & 490  & 62  \\
      & Kvasir-SEG & Variable           & 800 & 100 \\
    \hline
    \end{tabular}
    \caption{Details of the medical segmentation \textit{seen} clinical settings used in our experiments.}
    \label{tab:seen_clinical_dataset}
\end{table}

\begin{table}[h]
    \centering
    \scriptsize
    \setlength\tabcolsep{3pt} 
    \begin{tabular}{c|c|ccc}
    \hline
    Segmentation Task &  Dataset  & Resolutions & Test \\
    \hline
    \multicolumn{1}{c|}{\multirow{2}{*}{Multi-organ Segmentation}} & AMOS-CT & Variable & 100 Scans \\
     & AMOS-MRI & Variable & 20 Scans \\
    \hline
    Skin Cancer Segmentation     & PH2   & 767 $\times$ 576  & 200  \\
    \hline
    COVID19 Infection Segmentation & COVID19-2   & 512 $\times$ 512  & 2535 \\
    \hline
    Breast Cancer Segmentation    & STU    & Variable          & 42	  \\
    \hline
    \multicolumn{1}{c|}{\multirow{3}{*}{Polyp Segmentation}} & CVC-300    & 574 $\times$ 500  & 60   \\
      & CVC-ColonDB    & 574 $\times$ 500  & 380  \\
      & ETIS    & 1255 $\times$ 966 & 196  \\
    \hline
    \end{tabular}
    \caption{Details of the medical segmentation \textit{unseen} clinical settings used in our experiments. Note that these datasets are used only for testing.}
    \label{tab:unseen_clinical_dataset}
\end{table}

In this section, we describe the training and testing datasets used in this paper. Tab. \ref{tab:seen_clinical_dataset} and Tab. \ref{tab:unseen_clinical_dataset} summarize the seen and unseen clinical setting datasets, respectively. We want to clarify that Tab. \ref{tab:unseen_clinical_dataset} datasets are used only to evaluated the generalizability of each model.

\begin{itemize}
    \item \textit{Multi-organ Segmentation:} The Synapse multi-organ segmentation dataset \cite{Synapse_dataset} is a widely recognized benchmark in the medical imaging community, specifically designed for the task of abdominal organ segmentation. The dataset consists of 30 abdominal CT scans, encompassing a total of 3,779 axial contrast-enhanced images. Each image is annotated by medical experts to identify and delineate eight different abdominal organs: the aorta, gallbladder, spleen, left kidney, right kidney, liver, pancreas, and stomach. The AMOS dataset \cite{ji2022amos} is an advanced and comprehensive benchmark dataset designed for the multi-organ segmentation. The dataset includes both Computed Tomography (CT) and Magnetic Resonance Imaging (MRI) scans, providing a diverse range of imaging data. We apply same preprocessing steps as the Synapse to ensure consistency. For evaluation, we separated into AMOS-CT (100 Scans) and AMOS-MRI (20 Scans), and the labels were aligned with those used in the Synapse. We want to clarify that AMOS-CT/MRI datasets are used only to evaluated the generalizability of each model. \\
    
    \item \textit{Breast Ultrasound Segmentation:} The BUSI dataset \cite{al2020dataset} comprises 780 images from 600 female patients, including 133 normal cases, 437 benign cases, and 210 malignant tumors. In contrast, the STU dataset \cite{zhuang2019rdau} includes only 42 breast ultrasound images collected by Shantou University. Due to the limited number of images in the STU dataset, it is primarily used to evaluate the generalizability of models across different datasets. \\

    \item \textit{Skin Lesion Segmentation:} The ISIC 2018 dataset \cite{gutman2016skin} comprises 2,594 images of varying sizes. We randomly selected 1,868 images for training and 261 images for testing. Additionally, the PH2 dataset \cite{mendoncca2013ph} was used to evaluate the domain generalizability of each model. \\

    \item \textit{COVID19 Infection Segmentation:} The COVID19-1 dataset \cite{ma_jun_2020_3757476} contains 1,277 high-quality CT images. For our experiments, we randomly divided the dataset into 643 training images and 383 test images. To assess the domain generalizability of our models, we utilized the COVID19-2 dataset\footnote{https://www.kaggle.com/datasets/piyushsamant11/pidata-new-names} solely for testing purposes. \\
     
    \item \textit{Polyp Segmentation:} To train and evaluate our proposed model, we utilized five benchmark datasets: CVC-ColonDB \cite{tajbakhsh2015automated}, ETIS \cite{silva2014toward}, Kvasir \cite{jha2020kvasir}, CVC-300 \cite{vazquez2017benchmark}, and CVC-ClinicDB \cite{bernal2015wm}. For training, we adopted the same dataset as the latest image polyp segmentation method, comprising 800 samples from Kvasir and 490 samples from CVC-ClinicDB. The remaining images from these datasets, along with the other three datasets, were used exclusively for testing.
\end{itemize}

\begin{figure*}[t]
    \centering
    \includegraphics[width=\textwidth]{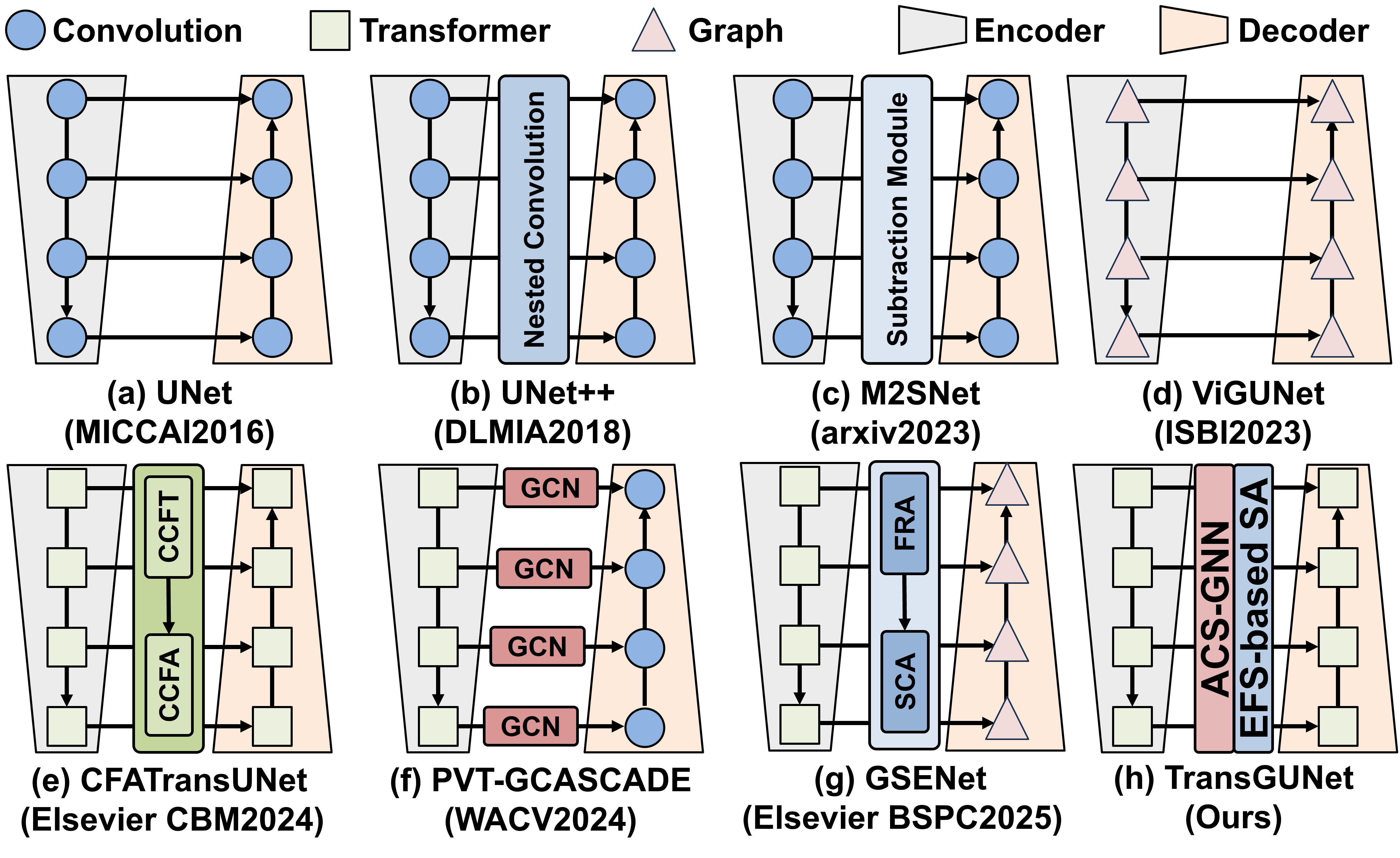}
    \caption{Comparison of skip connection frameworks scheme. Note that our unique approach (\textbf{TransGUNet}) incorporates \textit{ACS-GNN} with \textit{EFS-based spatial attention}.}
    \label{fig:SkipConnectionEngineeringScheme}
\end{figure*}

\section{Intuitiveness and Design Principle of TransGUNet}
We want to clarify that the design of TransGUNet is carefully considered rather than an ad-hoc decision, as follows: a) Medical images contain diverse anatomical structures, making it essential to flexibly capture both local details and global context. To address this, we transform cross-scale feature maps into a graph and apply efficient node-level attention. This design choice improves the model’s generalization across unseen clinical settings, as demonstrated in the "Ablation Study on ACS-GNN" section. b) Medical images segmentation models often produce uninformative feature maps (Figure 7 in the Appendix) due to various noise and irrelevant background artifacts, which can degrade the quality of spatial attention maps and reduce segmentation performance. To address this, we incorporate EFS-based spatial attention, prioritizing low-entropy feature maps to enhance the quality of spatial attention map. This novel approach also enhances generalization ability across unseen clinical settings, as shown in the "Ablation Study on EFS-based Attention" section.

\section{Technical Novelty of TransGUNet}
\noindent \textbf{PVT-GCASCADE (WACV2024) vs TransGUNet.} Like our approach, PVT-GCASCADE utilizes GNN; however, it does not consider cross-scale information, which is limited to medical images with more diverse lesion sizes. Additionally, unlike PVT-GCASCADE, it reduces the influence of uninformative feature maps with UFS to enable reliable spatial attention.

\noindent \textbf{CFATransGNet (CBM2024) vs TransGUNet.} TransGUNet focuses on efficiently reducing the semantic gap between the encoder and decoder using the GNN, which is not commonly used in skip connections. This suggests the potential for further development of GNN-based skip connection engineering beyond Transformer-based skip connection engineering (CFATransUNet).

\noindent \textbf{MADGNet (CVPR2024) vs TransGUNet.} While MADGNet employs Multi-Frequency and Multi-Scale Attention mechanisms to enhance feature extraction, TransGUNet's use of ACS-GNN allows for more robust integration of cross-scale features and adaptive learning of node importance. This novel approach not only improves segmentation accuracy but also addresses the limitations of traditional convolutional methods by leveraging the strengths of GNNs in handling diverse and noisy medical data.

\begin{table}[t]
    \centering
    \scriptsize
    \setlength\tabcolsep{5.0pt} 
    \begin{tabular}{c|ccc}
    \hline
    Method & Parameters (M) & FLOPs (G) & Inference Time (ms) \\
    \hline
    UNet        & 8.2 & 23.7 & 10.1 \\ 
    UNet++      & 34.9 & 197.8 & 22.9	\\
    CENet       & 18.0 & 9.2 & 10.5 \\ 
    TransUNet   & 53.4 & 21.8 & 93.4 \\
    nnUNet      & - & - & - \\
    MSRFNet     & 21.5 & 156.6 & 73.8 \\
    DCSAUNet    & 25.9 & 9.2 & 24.3 \\
    M2SNet      & 25.3 & 12.8 & 32.1 \\
    ViGUNet     & - & - & - \\
    PVT-GCASCADE & 25.4 & 7.9 & 17.4 \\ 
    CFATransUNet & 64.6 & 32.9 & 36.0 \\
    MADGNet     & 29.8 & 16.8 & 24.0 \\
    \hline
    \textbf{TransGUNet (Ours)} & 25.0 & 10.0 & 19.4 \\
    \hline
    \end{tabular}
    \caption{The number of parameters (M), FLOPs (G), and Inference Time (ms) of different models.}
    \label{tab:efficiency_analysis}
\end{table}

\section{Broader Impact in Artificial Intelligence}

TransGUNet’s superior performance in medical image segmentation has the potential to reliable medical diagnostics and treatment planning. By providing accurate and reliable segmentation of complex anatomical structures, it enables healthcare professionals to make more informed decisions, leading to improved patient outcomes. The model’s effective integration of Graph Neural Networks (GNNs) and Transformer architectures demonstrates a hybrid approach that addresses the limitations of traditional convolutional neural networks (CNNs), enhancing robustness and adaptability across various AI tasks. Furthermore, TransGUNet highlights the importance of developing efficient AI models that do not compromise on performance. Techniques like entropy-driven feature selection (EFS) and attentional cross-scale GNNs optimize resource use, setting a new benchmark for efficiency in AI-driven medical applications. This balance between high performance and computational efficiency can guide future research in creating powerful and resource-conscious AI models.

\section{More Detailed Ablation Study on TransGUNet}
In this section, we perform a more detailed ablation study on TransGUNet.

\subsection{Ablation Study on Backbone in TransGUNet}

\begin{table}[h]
    \centering
    \scriptsize
    \setlength\tabcolsep{2.0pt} 
    \begin{tabular}{c|c|cc|cc|c|c}
    \hline
    Network & \multicolumn{1}{c|}{\multirow{2}{*}{Backbone}} & \multicolumn{2}{c|}{\textit{Seen}}  & \multicolumn{2}{c|}{\textit{Unseen}} & \multicolumn{1}{c|}{\multirow{2}{*}{Param (M)}}  & \multicolumn{1}{c}{\multirow{2}{*}{FLOPs (G)}} \\ \cline{3-6}
    Type    & & DSC & mIoU & DSC & mIoU & & \\
    \hline
    \multicolumn{1}{c|}{\multirow{3}{*}{CNN}} & ResNet50 & 85.9 & 79.1 & 67.3 & 60.5 & 25.1M & 18.0G \\
     & Res2Net50 & 86.0 & 79.1 & 70.6 & 62.8 & 25.2M & 18.8G \\
     & ResNeSt50 & \textit{86.6} & 79.7 & 73.3 & 64.9 & 26.9M & 20.4G \\
     \hline
    \multicolumn{1}{c|}{\multirow{3}{*}{Transformer}} & ViT-B-16 & 83.6 & 75.9 & 64.7 & 56.3 & 53.4M & 18.4G \\
     & PVT-v2-b2 & \textit{86.6} & \textit{80.0} & \textit{77.0} & \textit{68.3} & 26.3M & 10.5G \\
     & \textbf{P2T-Small \tiny{(Ours)}} & \textbf{\underline{87.3}} & \textbf{\underline{80.6}} & \textbf{\underline{78.6}} & \textbf{\underline{69.7}} & 25.0M & 10.0G \\
     \hline
    \end{tabular}
    \caption{Quantitative results for each \textit{Seen} and \textit{Unseen} datasets according to backbone network.}
    \label{tab:ablation_backbone_networks}
\end{table}

In this section, we conduct an ablation study to evaluate the impact of different backbone models on the performance of TransGUNet. This experiment uses several popular CNN and Transformer architectures, including ResNet50 \cite{he2016deep}, Res2Net \cite{gao2019res2net}, ResNeSt50 \cite{zhang2022resnest}, ViT-B-16 \cite{dosovitskiy2021an}, PVT-v2-b2 \cite{wang2022pvt}, and P2T-Small \cite{wu2022p2t}. Notably, only the backbone network was changed, while all other architectural settings remained consistent with those in the main experiment. We reported the \textit{mean} performance for each \textit{seen} and \textit{unseen} clinical settings in Tab. \ref{tab:ablation_backbone_networks}. The datasets used in the \textit{seen} and \textit{unseen} clinical settings are the same as Tab. \ref{tab:comparison_sota_in_domain} and Tab. \ref{tab:comparison_sota_out_domain}, respectively. For convenience, we denote $( \cdot, \cdot )$ as the performance improvement gap between TransGUNet and other models for seen and unseen clinical settings.

ResNeSt50, as utilized in MADGNet, achieves comparable or even superior performance. Furthermore, using PVT-v2-b2, as implemented in PVT-GCASCADE, yields higher performance in both seen and unseen clinical settings. Specifically, TransGUNet with PVT-v2-b2 outperforms PVT-GCASCADE by \textbf{\underline{(0.4\%, 2.2\%)}} and \textbf{\underline{(0.7\%, 2.2\%)}} in DSC and mIoU, respectively, demonstrating its robustness and effectiveness according to the backbone type.

\subsection{Ablation Study on ECA Kernel Size in ACS-GNN}

\begin{table}[h]
    \centering
    \scriptsize
    \setlength\tabcolsep{4.0pt} 
    \begin{tabular}{c|cc|cc|c|c}
    \hline
    \multicolumn{1}{c|}{\multirow{2}{*}{ECA Kernel Size $k$}} & \multicolumn{2}{c|}{\textit{Seen}}  & \multicolumn{2}{c|}{\textit{Unseen}} & \multicolumn{1}{c|}{\multirow{2}{*}{Param (M)}}  & \multicolumn{1}{c}{\multirow{2}{*}{FLOPs (G)}} \\ \cline{2-5}
       & DSC & mIoU & DSC & mIoU & & \\ 
    \hline
    $k = 3$ \textbf{(Ours)}                 & \textbf{\underline{87.3}} & \textbf{\underline{80.6}} & \textbf{\underline{78.6}} & \textbf{\underline{69.7}} & 25.0M & 10.0G \\
    $k = 5$ & 87.5 & 80.7 & 78.3 & 69.6 & 25.0M & 10.0G \\
    $k = 7$ & 87.4 & 80.5 & 78.4 & 69.5 & 25.0M & 10.0G \\
    \hline
    \end{tabular}
    \caption{Quantitative results for each \textit{Seen} and \textit{Unseen} datasets according to various ECA kernel size $k$.}
    \label{tab:ablation_eca_kernel_size}
\end{table}

In this section, we conduct an ablation study to compare the performance according to various ECA kernel size $k = \{ 3, 5, 7 \}$. In the main manuscript, we used $k = 3$ as the ECA kernel size. We reported the \textit{mean} performance for each \textit{seen} and \textit{unseen} clinical settings in Tab. \ref{tab:ablation_eca_kernel_size}. The datasets used in the \textit{seen} and \textit{unseen} clinical settings are the same as Tab. \ref{tab:comparison_sota_in_domain} and Tab. \ref{tab:comparison_sota_out_domain}, respectively. 

As listed in Tab. \ref{tab:ablation_eca_kernel_size}, we observed that while increasing the kernel size for $k = 3$ to $k = 7$, it does not provide meaningful statistical improvement. Consequently, we choose $k = 3$ as our model basic configuration in ECA kernel size.

\subsection{Ablation Study on Repetition Time of ACS-GNN with EFS-based Spatial Attention}

\begin{table}[h]
    \centering
    \scriptsize
    \setlength\tabcolsep{2.5pt} 
    \begin{tabular}{c|cc|cc|c|c}
    \hline
    \multicolumn{1}{c|}{\multirow{2}{*}{Repetition Time $G$}} & \multicolumn{2}{c|}{\textit{Seen}}  & \multicolumn{2}{c|}{\textit{Unseen}} & \multicolumn{1}{c|}{\multirow{2}{*}{Param (M)}}  & \multicolumn{1}{c}{\multirow{2}{*}{FLOPs (G)}} \\ \cline{2-5}
     & DSC & mIoU & DSC & mIoU & & \\ 
    \hline
     $G = 3$ & 86.7 & 80.0 & 77.4 & 68.6 & 26.6M & 12.4G \\
     $G = 5$ & 86.4 & 79.7 & 77.2 & 68.3 & 28.1M & 14.7G \\
     $G = 7$ & 86.3 & 80.0 & 77.1 & 68.3 & 29.6M & 17.0G \\
     $G = 9$ & 86.3 & 79.7 & 77.1 & 68.2 & 31.1M & 19.4G \\
    \hline
    $G = 1$ \textbf{(Ours)} & \textbf{\underline{87.3}} & \textbf{\underline{80.6}} & \textbf{\underline{78.6}} & \textbf{\underline{69.7}} & 25.0M & 10.0G \\
    \hline
    \end{tabular}
    \caption{Quantitative results for each \textit{Seen} and \textit{Unseen} datasets according to repetition time $G$ of ACS-GNN with EFS-based spatial attention.}
    \label{tab:ablation_repetition_time}
\end{table}

In this section, we conduct an ablation study to compare the performance according to repetition time $G \in \{ 1, 3, 5, 7, 9 \}$ of ACS-GNN with EFS-based spatial attention. In the main manuscript, we used $G = 1$ as the default repetition time. We reported the \textit{mean} performance for each \textit{seen} and \textit{unseen} clinical settings in Tab. \ref{tab:ablation_repetition_time}. The datasets used in the \textit{seen} and \textit{unseen} clinical settings are the same as Tab. \ref{tab:comparison_sota_in_domain} and Tab. \ref{tab:comparison_sota_out_domain}, respectively. 

The ablation study reveals that repeating the ACS-GNN module with EFS-based spatial attention leads to overfitting. Despite the increased capacity of the model, performance decreased on both seen and unseen datasets. This result indicates that the model is overfitting to the training data.

\section{Metrics Descriptions}
\label{appendix_metric_descriptions}

In this section, we describe the metrics used in this paper. For convenience, we denote $TP, FP$, and $FN$ as the number of samples of true positive, false positive, and false negative between two binary masks $A$ and $B$.  

\begin{itemize}
    \item The \textit{Mean Dice Similarity Coefficient (DSC)} \cite{milletari2016v} measures the similarity between two samples and is widely used to assess the performance of segmentation tasks, such as image segmentation or object detection. \textbf{\underline{Higher is better}}. For given two binary masks $A$ and $B$, DSC is defined as follows:
    \begin{equation}
        \textbf{DSC}(A, B) = \frac{2 \times | A \cap B |}{| A \cup B |} = \frac{2 \times TP}{2 \times TP + FP + FN}
    \end{equation}

    \item The \textit{Mean Intersection over Union (mIoU)} measures the ratio of the intersection area to the union area between predicted and ground truth masks in segmentation tasks. \textbf{\underline{Higher is better}}. For given two binary masks $A$ and $B$, mIoU is defined as follows:
    \begin{equation}
        \textbf{mIoU}(A, B) = \frac{A \cap B}{A \cup B} = \frac{TP}{TP + FP + FN}
    \end{equation}

    \item The \textit{Mean Weighted F-Measure} $F_{\beta}^{\omega}$ \cite{margolin2014evaluate} is a metric that combines weighted precision $Precision^{\omega}$ (Measure of exactness) and weighted recall $Recall^{\omega}$ (Measure of completeness) into a single value by calculating the harmonic mean. $\beta$ signifies the effectiveness of detection with respect to a user who attaches $\beta$ times as much importance to $Recall^{\omega}$ as to $Precision^{\omega}$. \textbf{\underline{Higher is better}}. $F_{\beta}^{\omega}$ is defined as follows:
    \begin{equation}
        F_{\beta}^{\omega} = (1 + \beta^{2}) \cdot \frac{Precision^{\omega} \cdot Recall^{\omega}}{\beta^{2} \cdot Precision^{\omega} + Recall^{\omega}}
    \end{equation}

    \item The \textit{Mean S-Measure} $S_{\alpha}$ \cite{fan2017structure} is used to evaluate the quality of image segmentation, specifically focusing on the structural similarity between the region-aware $S_{o}$ and object-aware $S_{r}$. \textbf{\underline{Higher is better}}. $S_{\alpha}$ is defined as follows:
    \begin{equation}
        S_{\alpha} = \gamma S_{o} + (1 - \gamma) S_{r}
    \end{equation}
    
    \item \textit{Mean E-Measure} $E_{\phi}^{max}$ \cite{fan2018enhanced} assesses the edge accuracy in edge detection or segmentation tasks. It evaluates how well the predicted edges align with the ground truth edges using foreground map $FM$. \textbf{\underline{Higher is better}}. $E_{\phi}^{max}$ is defined as follows: \\
    \begin{equation}
        E_{\phi}^{max} = \frac{1}{H \times W} \sum_{h = 1}^{H} \sum_{w = 1}^{W} \phi FM(h, w)
    \end{equation}

    \item \textit{Mean Absolute Error (MAE)} calculates the average absolute differences between predicted and ground truth values. \textbf{\underline{Lower is better}}. For given two binary masks $A$ and $B$, MAE is defined as follows: \\
    \begin{equation}
        \textbf{MAE}(A, B) = \frac{1}{H \times W} \sum_{h = 1}^{H} \sum_{w = 1}^{W} \left| A(h, w) - B(h, w) \right|
    \end{equation}

    \item \textit{Hausdorff Distance 95 (HD95)} \cite{celaya2023generalized} is the 95th percentile of the Hausdorff Distance, which means it excludes the top 5\% of the most extreme distances for more robust to noise and outliers. It is calculated by first computing all pairwise distances between the points in the predicted and ground truth segmentations, sorting these distances, and then taking the 95th percentile value. Note that the Hausdorff distance ranges from 0, indicating no difference when two sets are identical, to infinity, as the maximum distance between two sets can grow indefinitely. \textbf{\underline{Lower is better}}. For given two binary masks $A$ and $B$ and euclidean distance $d(\cdot, \cdot)$, HD is defined as follows:
    \begin{dmath}
        \textbf{HD}(A, B) = \textbf{max} \left( \textbf{max}_{a \in A} \textbf{min}_{b \in B} d(a, b), \textbf{max}_{b \in B} \textbf{min}_{a \in A} d(a, b) \right)
    \end{dmath}
\end{itemize}

\section{More Qualtative and Quantitative Results}
\begin{figure*}[hbtp]
    \centering
    \includegraphics[width=\textwidth]{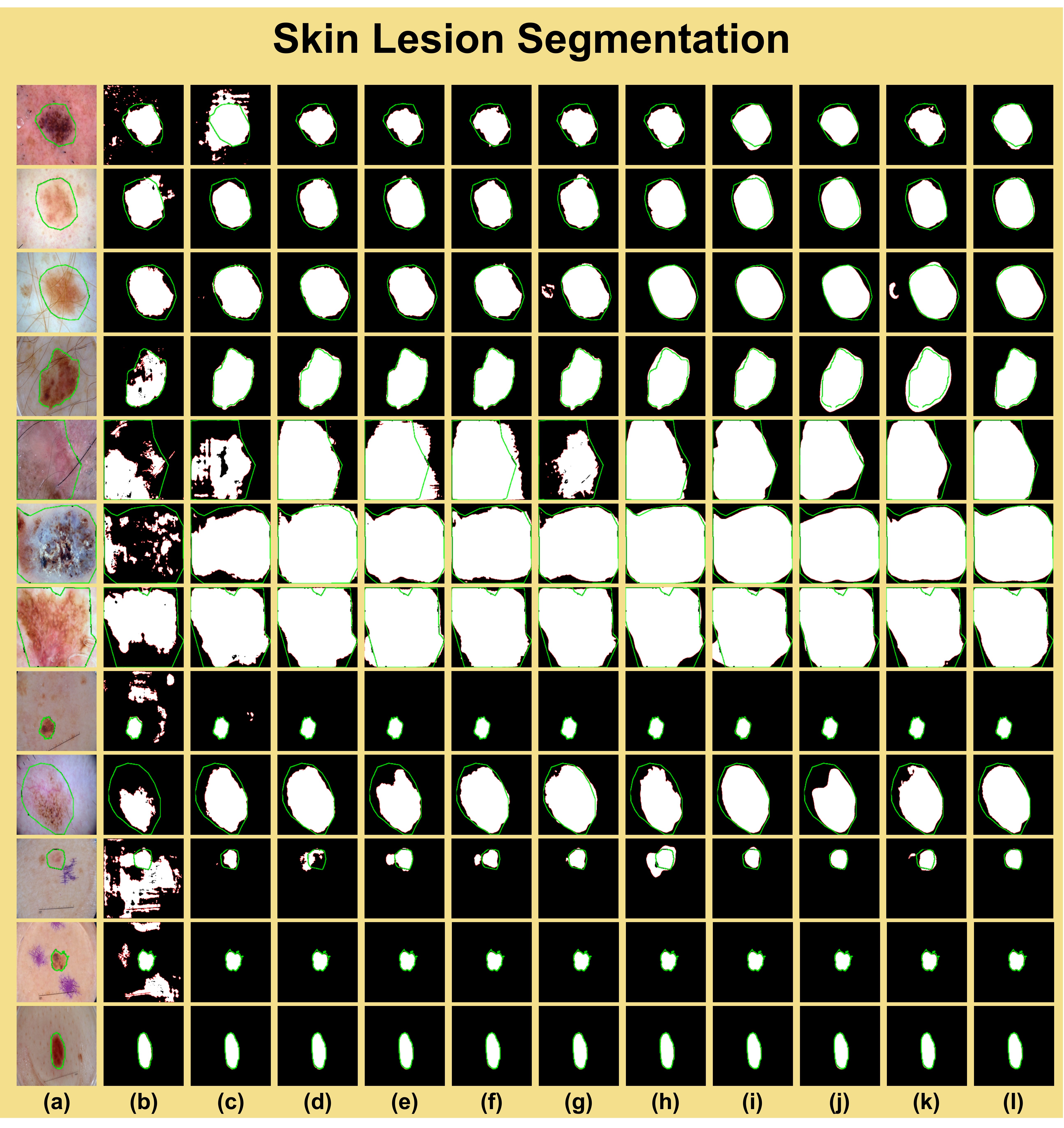}
    \caption{Qualitative comparison of other methods and \textbf{TransGUNet} on \textbf{\underline{Skin Lesion Segmentation}} \cite{gutman2016skin, mendoncca2013ph}. (a) Input images with ground truth. (b) UNet. (c) UNet++. (d) CENet. (e) TransUNet. (f) MSRFNet. (g) DCSAUNet. (h) M2SNet. (i) PVT-GCASCADE. (j) CFATransUNet. (k) MADGNet. (l) \textbf{TransGUNet (Ours)}. \textbf{Green} and \textbf{Red} lines denote the boundaries of the ground truth and prediction, respectively.}
    \label{fig:Sup_QualitativeResults_Dermatoscopy}
\end{figure*}

\begin{figure*}[hbtp]
    \centering
    \includegraphics[width=\textwidth]{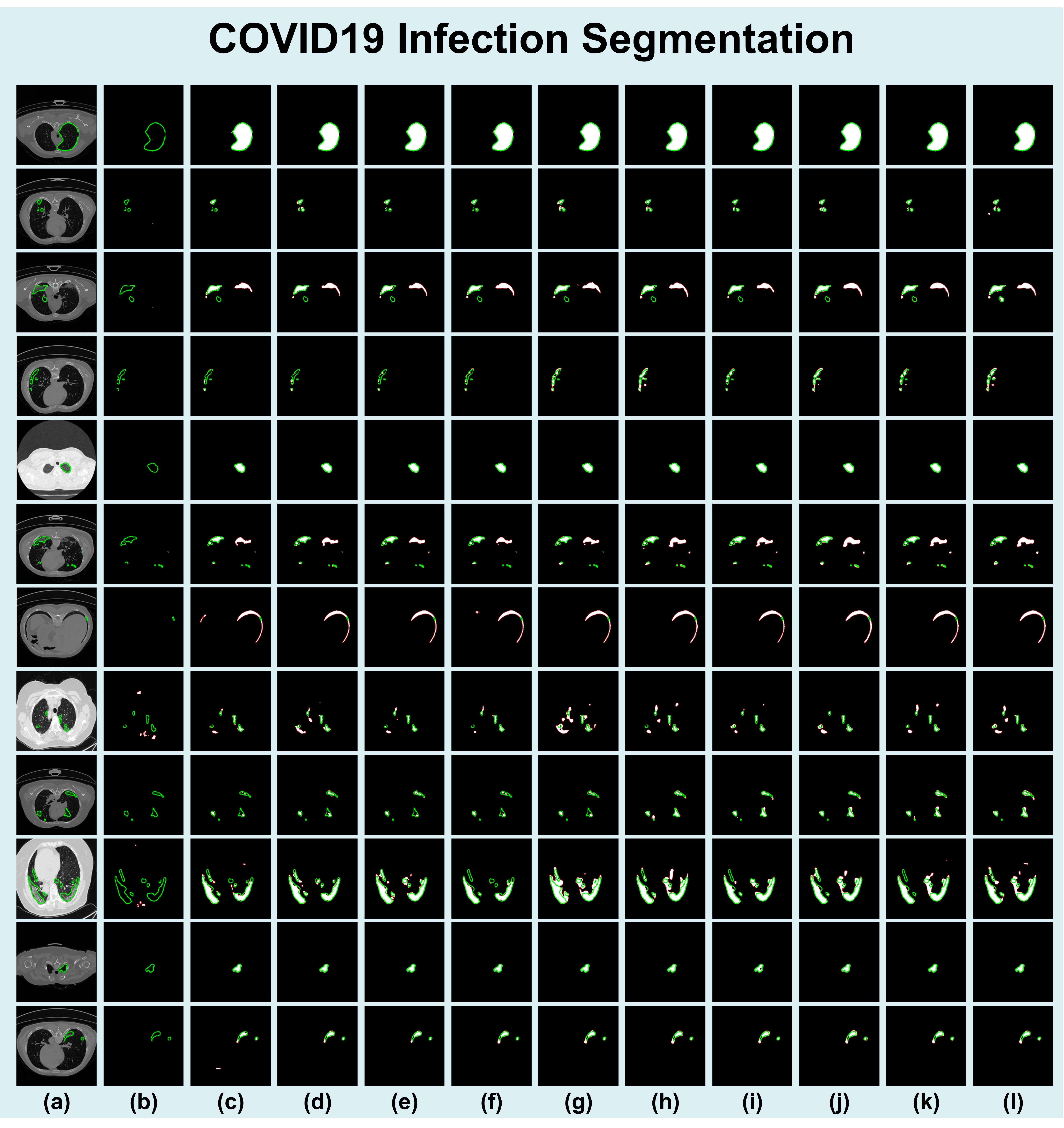}
    \caption{Qualitative comparison of other methods and \textbf{TransGUNet} on \textbf{\underline{COVID19 Infection Segmentation}} \cite{ma_jun_2020_3757476}. (a) Input images with ground truth. (b) UNet. (c) UNet++. (d) CENet. (e) TransUNet. (f) MSRFNet. (g) DCSAUNet. (h) M2SNet. (i) PVT-GCASCADE. (j) CFATransUNet. (k) MADGNet. (l) \textbf{TransGUNet (Ours)}. \textbf{Green} and \textbf{Red} lines denote the boundaries of the ground truth and prediction, respectively.}
    \label{fig:Sup_QualitativeResults_Radiology}
\end{figure*}

\begin{figure*}[hbtp]
    \centering
    \includegraphics[width=\textwidth]{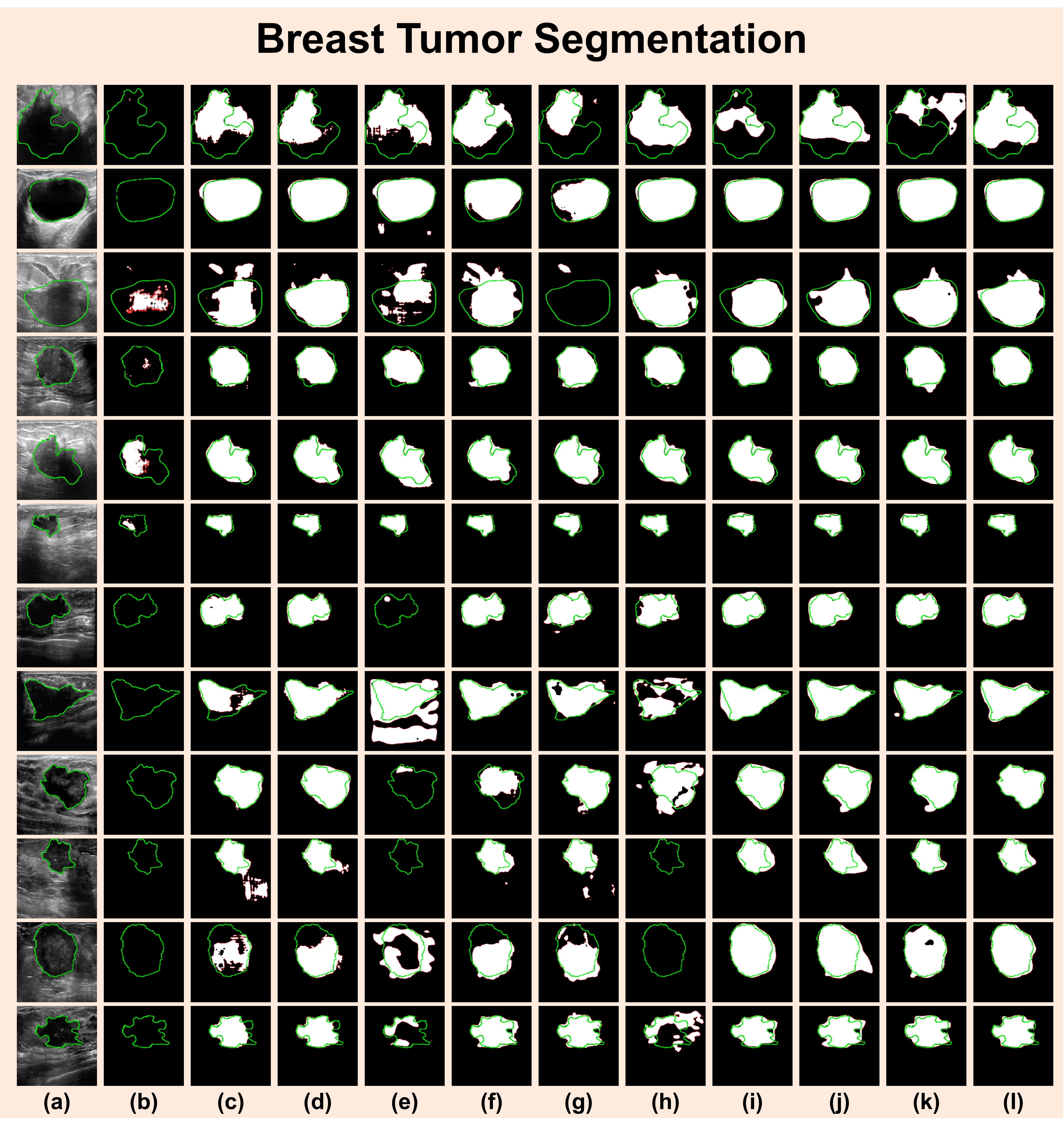}
    \caption{Qualitative comparison of other methods and \textbf{TransGUNet} on \textbf{\underline{Breast Tumor Segmentation}} \cite{al2020dataset, zhuang2019rdau}. (a) Input images with ground truth. (b) UNet. (c) UNet++. (d) CENet. (e) TransUNet. (f) MSRFNet. (g) DCSAUNet. (h) M2SNet. (i) PVT-GCASCADE. (j) CFATransUNet. (k) MADGNet. (l) \textbf{TransGUNet (Ours)}. \textbf{Green} and \textbf{Red} lines denote the boundaries of the ground truth and prediction, respectively.}
    \label{fig:Sup_QualitativeResults_Ultrasound}
\end{figure*}

\begin{figure*}[hbtp]
    \centering
    \includegraphics[width=\textwidth]{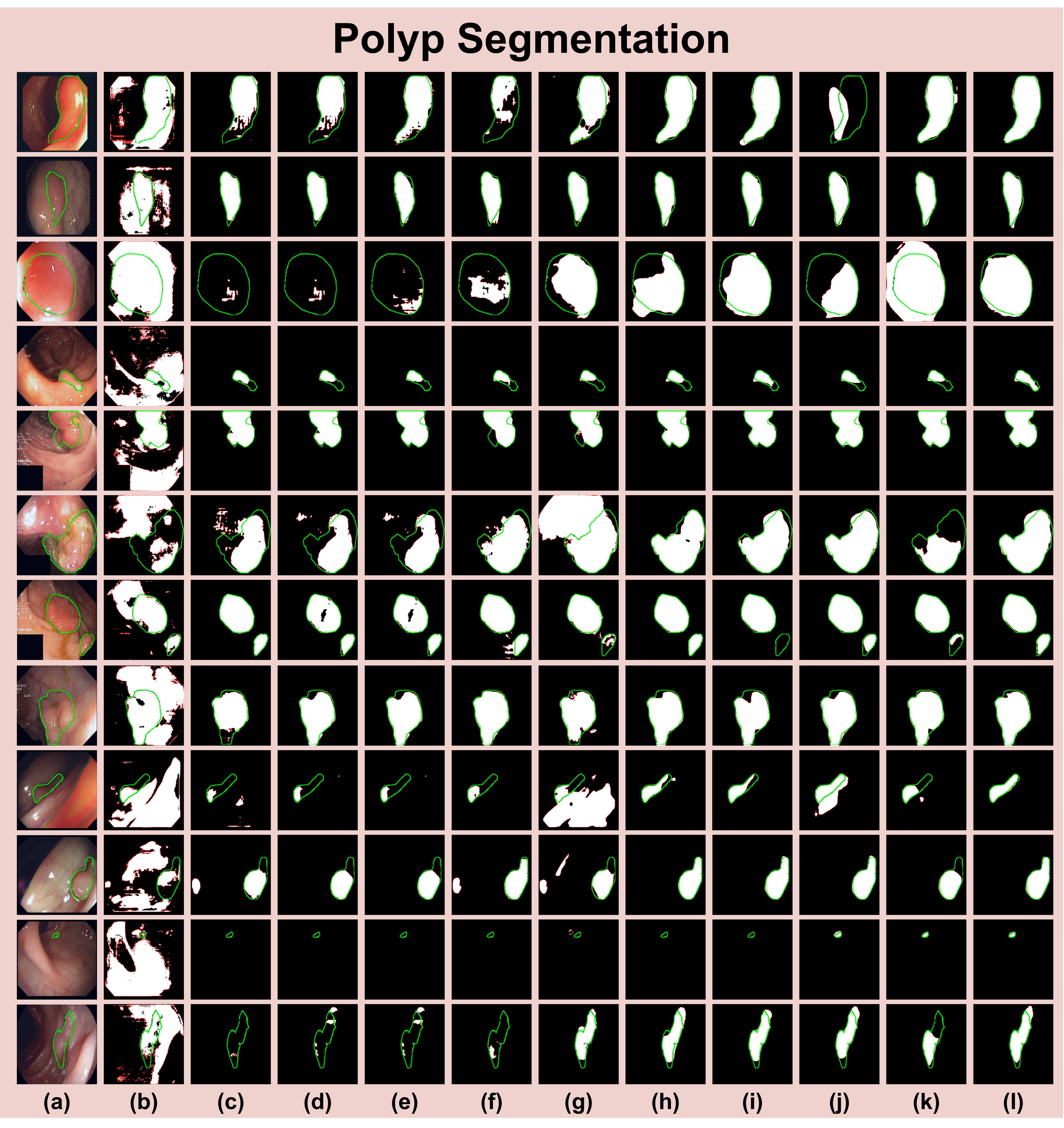}
    \caption{Qualitative comparison of other methods and \textbf{TransGUNet} on \textbf{\underline{Polyp Segmentation}} \cite{bernal2015wm, jha2020kvasir, vazquez2017benchmark, tajbakhsh2015automated, silva2014toward}. (a) Input images with ground truth. (b) UNet. (c) UNet++. (d) CENet. (e) TransUNet. (f) MSRFNet. (g) DCSAUNet. (h) M2SNet. (i) PVT-GCASCADE. (j) CFATransUNet. (k) MADGNet. (l) \textbf{TransGUNet (Ours)}. \textbf{Green} and \textbf{Red} lines denote the boundaries of the ground truth and prediction, respectively.}
    \label{fig:Sup_QualitativeResults_Colonoscopy}
\end{figure*}

In this section, we provide the quantitative results with various metrics in Tab. \ref{tab:comparison_sota_dermatoscopy_other_metrics}, \ref{tab:comparison_sota_radiology_other_metrics}, \ref{tab:comparison_sota_ultrasound_other_metrics}, \ref{tab:comparison_sota_colonoscopy_other_metrics} for binary segmentation. Additionally, we also provide the quantitative results in Tab. \ref{tab:comparison_sota_multiorgan_dsc_metrics}, \ref{tab:comparison_sota_multiorgan_mIoU_metrics}, \ref{tab:comparison_sota_multiorgan_HD95_metrics} with each organ for multi-organ segmentation. For all tables, \textbf{\underline{Bold}} and \textit{italic} are the first and second best performance results, respectively. We also present more various qualitative results on datasets in Fig. \ref{fig:Sup_QualitativeResults_Dermatoscopy}, \ref{fig:Sup_QualitativeResults_Radiology}, \ref{fig:Sup_QualitativeResults_Ultrasound}, \ref{fig:Sup_QualitativeResults_Colonoscopy}.

\begin{table}[h]
    \centering
    \scriptsize
    \setlength\tabcolsep{2pt} 
    \renewcommand{\arraystretch}{0.9} 
    \begin{tabular}{c|cccccc}
    \hline
    \multicolumn{1}{c|}{\multirow{2}{*}{Method}} & \multicolumn{6}{c}{ISIC2018 $\Rightarrow$ ISIC2018} \\ \cline{2-7}
     & DSC \scriptsize{$\uparrow$} & mIoU \scriptsize{$\uparrow$} & $F_{\beta}^{w}$ \scriptsize{$\uparrow$}  & $S_{\alpha}$ \scriptsize{$\uparrow$} & $E_{\phi}^{max}$ \scriptsize{$\uparrow$} & MAE \scriptsize{$\downarrow$} \\
     \hline
     UNet \tiny{(MICCAI2016)}       & 86.9 & 80.2 & 87.9 & 80.4 & 91.3 & 4.7 \\
     UNet++ \tiny{(DLMIA2018)}      & 87.8 & 80.5 & 86.5 & 80.5 & 92.0 & 4.5 \\
     CENet \tiny{(TMI2019)}         & 89.1 & 82.1 & 88.1 & 81.3 & 93.0 & 4.3 \\
     TransUNet \tiny{(arxiv2021)}   & 87.3 & 81.2 & 88.6 & 80.8 & 91.9 & 4.2 \\
     MSRFNet \tiny{(BHI2022)}       & 88.2 & 81.3 & 86.9 & 80.7 & 92.0 & 4.7 \\
     DCSAUNet \tiny{(CBM2023)}      & 89.0 & 82.0 & 87.8 & 81.4 & 92.9 & 4.4 \\
     M2SNet \tiny{(arxiv2023)}      & 89.2 & 83.4 & \textit{90.0} & 82.0 & 93.8 & 3.7 \\
     PVT-GCASCADE \tiny{(WACV2024)} & \textit{90.3} & 83.5 & 88.9 & 82.0 & 93.9 & 3.7 \\
     CFATransUNet \tiny{(CBM2024)}  & 90.1 & 83.5 & 89.0 & \textit{82.1} & \textit{94.1} & \textit{3.5} \\
     MADGNet \tiny{(CVPR2024)}      & 90.2 & \textit{83.7} & 89.2 & 82.0 & \textit{94.1} & 3.6 \\
     \hline
     \multicolumn{1}{c|}{\multirow{2}{*}{\textbf{TransGUNet \tiny{(Ours)}}}}     & \textbf{\underline{91.1}} & \textbf{\underline{84.8}} & \textbf{\underline{90.1}} & \textbf{\underline{82.6}} & \textbf{\underline{94.4}} & \textbf{\underline{3.4}} \\ \cline{2-7}
     & \textbf{+0.8} & \textbf{+1.1} & \textbf{+0.1} & \textbf{+0.5} & \textbf{+0.3} & \textbf{+0.2} \\
    \hline
    \multicolumn{1}{c|}{\multirow{2}{*}{Method}} & \multicolumn{6}{c}{ISIC2018 $\Rightarrow$ PH2} \\ \cline{2-7}
     & DSC \scriptsize{$\uparrow$} & mIoU \scriptsize{$\uparrow$} & $F_{\beta}^{w}$ \scriptsize{$\uparrow$}  & $S_{\alpha}$ \scriptsize{$\uparrow$} & $E_{\phi}^{max}$ \scriptsize{$\uparrow$} & MAE \scriptsize{$\downarrow$} \\
     \hline
     UNet \tiny{(MICCAI2016)}       & 90.3 & 83.5 & 88.4 & 74.8 & 90.8 & 6.9 \\
     UNet++ \tiny{(DLMIA2018)}      & 88.0 & 80.1 & 85.7 & 73.2 & 89.2 & 7.9 \\
     CENet \tiny{(TMI2019)}         & 90.5 & 83.3 & 87.3 & \textbf{\underline{78.1}} & 91.5 & 6.0 \\
     TransUNet \tiny{(arxiv2021)}   & 89.5 & 82.1 & 86.9 & 74.3 & 90.3 & 6.7 \\
     MSRFNet \tiny{(BHI2022)}       & 90.5 & 83.5 & 87.5 & 75.0 & 91.4 & 6.0 \\
     DCSAUNet \tiny{(CBM2023)}      & 89.0 & 81.5 & 85.7 & 74.0 & 90.2 & 6.9 \\
     M2SNet \tiny{(arxiv2023)}      & 90.7 & 83.5 & 87.6 & 75.5 & 92.0 & 5.9 \\
     PVT-GCASCADE \tiny{(WACV2024)} & \textit{91.5} & 84.9 & 88.6 & 76.3 & 92.7 & 5.3 \\
     CFATransUNet \tiny{(CBM2024)}  & \textit{91.5} & \textit{85.0} & \textit{88.7} & 76.3 & 92.6 & 5.3 \\
     MADGNet \tiny{(CVPR2024)}      & 91.3 & 84.6 & 88.4 & 76.2 & \textit{92.8} & \textit{5.1} \\
     \hline
     \multicolumn{1}{c|}{\multirow{2}{*}{\textbf{TransGUNet \tiny{(Ours)}}}}     & \textbf{\underline{91.7}} & \textbf{\underline{85.2}} & \textbf{\underline{88.9}} & \textit{76.6} & \textbf{\underline{93.1}} & \textbf{\underline{5.0}} \\ \cline{2-7}
     & \textbf{+0.2} & \textbf{+0.2} & \textbf{+0.2} & \textbf{-1.5} & \textbf{+0.3} & \textbf{+0.1} \\
    \hline
    \end{tabular}
    \caption{Segmentation results on \textbf{Skin Lesion Segmentation}. We train each model on ISIC2018 \cite{gutman2016skin} train dataset and evaluate on ISIC2018 \cite{gutman2016skin} and PH2 \cite{mendoncca2013ph} test datasets.}
    \label{tab:comparison_sota_dermatoscopy_other_metrics}
\end{table}

\begin{table}[h]
    \centering
    \scriptsize
    \setlength\tabcolsep{2pt} 
    \renewcommand{\arraystretch}{0.9} 
    \begin{tabular}{c|cccccc}
    \hline
    \multicolumn{1}{c|}{\multirow{2}{*}{Method}} & \multicolumn{6}{c}{COVID19-1 $\Rightarrow$ COVID19-1} \\ \cline{2-7}
     & DSC \scriptsize{$\uparrow$} & mIoU \scriptsize{$\uparrow$} & $F_{\beta}^{w}$ \scriptsize{$\uparrow$}  & $S_{\alpha}$ \scriptsize{$\uparrow$} & $E_{\phi}^{max}$ \scriptsize{$\uparrow$} & MAE \scriptsize{$\downarrow$} \\
     \hline
     UNet \tiny{(MICCAI2016)}       & 47.7 & 38.6 & 36.1 & 69.6 & 62.7 & 2.1 \\
     UNet++ \tiny{(DLMIA2018)}      & 65.6 & 57.1 & 54.4 & 78.8 & 73.2 & 1.3 \\
     CENet \tiny{(TMI2019)}         & 76.3 & 69.2 & 64.4 & 83.2 & 76.6 & 0.6 \\
     TransUNet \tiny{(arxiv2021)}   & 75.6 & 68.8 & 63.4 & 82.7 & 75.5 & 0.7 \\
     MSRFNet \tiny{(BHI2022)}       & 75.2 & 68.0 & 63.4 & 82.7 & 76.3 & 0.8 \\
     DCSAUNet \tiny{(CBM2023)}      & 75.3 & 68.2 & 63.1 & 83.0 & 77.3 & 0.7 \\
     M2SNet \tiny{(arxiv2023)}      & 81.7 & 74.7 & \textit{68.3} & 85.7 & 80.1 & 0.6 \\
     PVT-GCASCADE \tiny{(WACV2024)} & 82.3 & 74.8 & 68.1 & 85.8 & 80.1 & \textit{0.5} \\
     CFATransUNet \tiny{(CBM2024)}  & 80.4 & 73.6 & \textit{68.3} & 84.8 & 79.2 & \textit{0.5} \\
     MADGNet \tiny{(CVPR2024)}      & \textit{83.7} & \textit{76.8} & \textbf{\underline{70.2}} & \textit{86.3} & \textbf{\underline{81.5}} & \textit{0.5} \\
     \hline
     \multicolumn{1}{c|}{\multirow{2}{*}{\textbf{TransGUNet \tiny{(Ours)}}}}     & \textbf{\underline{84.0}} & \textbf{\underline{77.0}} & \textbf{\underline{70.2}} & \textbf{\underline{86.6}} & \textit{81.2} & \textbf{\underline{0.4}} \\ \cline{2-7}
     & \textbf{+0.3} & \textbf{+0.2} & \textbf{+0.0} & \textbf{+0.3} & \textbf{-0.3} & \textbf{+0.1} \\
    \hline
    \multicolumn{1}{c|}{\multirow{2}{*}{Method}} & \multicolumn{6}{c}{COVID19-1 $\Rightarrow$ COVID19-2} \\ \cline{2-7}
     & DSC \scriptsize{$\uparrow$} & mIoU \scriptsize{$\uparrow$} & $F_{\beta}^{w}$ \scriptsize{$\uparrow$}  & $S_{\alpha}$ \scriptsize{$\uparrow$} & $E_{\phi}^{max}$ \scriptsize{$\uparrow$} & MAE \scriptsize{$\downarrow$} \\
     \hline
     UNet \tiny{(MICCAI2016)}       & 47.1 & 37.7 & 46.7 & 68.7 & 68.6 & 1.0 \\
     UNet++ \tiny{(DLMIA2018)}      & 50.5 & 40.9 & 50.6 & 69.8 & 75.7 & 1.0 \\
     CENet \tiny{(TMI2019)}         & 60.1 & 49.9 & 61.1 & 73.4 & 80.1 & 1.1 \\
     TransUNet \tiny{(arxiv2021)}   & 56.9 & 48.0 & 58.0 & 72.5 & 79.7 & \textbf{\underline{0.8}} \\
     MSRFNet \tiny{(BHI2022)}       & 58.3 & 48.4 & 59.1 & 72.7 & 79.8 & 1.0 \\
     DCSAUNet \tiny{(CBM2023)}      & 52.4 & 44.0 & 52.0 & 71.3 & 76.3 & 1.0 \\
     M2SNet \tiny{(arxiv2023)}      & 68.6 & 58.9 & 68.5 & 76.9 & 86.1 & 1.1 \\
     PVT-GCASCADE \tiny{(WACV2024)} & 71.0 & 60.4 & 70.0 & 77.8 & 87.9 & 1.2 \\
     CFATransUNet \tiny{(CBM2024)}  & 65.7 & 56.2 & 67.0 & 75.1 & 83.0 & 1.2 \\
     MADGNet \tiny{(CVPR2024)}      & \textit{72.2} & \textbf{\underline{62.6}} & \textbf{\underline{72.3}} & \textit{78.2} & \textit{88.1} & 1.0 \\
     \hline
     \multicolumn{1}{c|}{\multirow{2}{*}{\textbf{TransGUNet \tiny{(Ours)}}}}     & \textbf{\underline{73.0}} & \textit{62.4} & \textit{72.0} & \textbf{\underline{78.7}} & \textbf{\underline{89.5}} & \textit{0.9} \\ \cline{2-7}
     & \textbf{+0.8} & \textbf{-0.2} & \textbf{-0.3} & \textbf{+0.5} & \textbf{+1.4} & \textbf{-0.1} \\
    \hline
    \end{tabular}
    \caption{Segmentation results on \textbf{COVID19 Infection Segmentation}. We train each model on COVID19-1 \cite{ma_jun_2020_3757476} train dataset and evaluate on COVID19-1 \cite{ma_jun_2020_3757476} and COVID19-2 test datasets.}
    \label{tab:comparison_sota_radiology_other_metrics}
\end{table}

\begin{table}[h]
    \centering
    \scriptsize
    \setlength\tabcolsep{2pt} 
    \renewcommand{\arraystretch}{0.9} 
    \begin{tabular}{c|cccccc}
    \hline
    \multicolumn{1}{c|}{\multirow{2}{*}{Method}} & \multicolumn{6}{c}{BUSI $\Rightarrow$ BUSI} \\ \cline{2-7}
     & DSC \scriptsize{$\uparrow$} & mIoU \scriptsize{$\uparrow$} & $F_{\beta}^{w}$ \scriptsize{$\uparrow$}  & $S_{\alpha}$ \scriptsize{$\uparrow$} & $E_{\phi}^{max}$ \scriptsize{$\uparrow$} & MAE \scriptsize{$\downarrow$} \\
     \hline
     UNet \tiny{(MICCAI2016)}       & 69.5 & 60.2 & 67.2 & 76.9 & 83.2 & 4.8 \\
     UNet++ \tiny{(DLMIA2018)}      & 71.3 & 62.3 & 68.9 & 78.1 & 84.4 & 4.8 \\
     CENet \tiny{(TMI2019)}         & 79.7 & 71.5 & 78.1 & 82.8 & 91.1 & 3.9 \\
     TransUNet \tiny{(arxiv2021)}   & 75.5 & 68.4 & 73.8 & 79.8 & 88.6 & 4.2 \\
     MSRFNet \tiny{(BHI2022)}       & 76.6 & 68.1 & 75.1 & 80.9 & 88.5 & 4.2 \\
     DCSAUNet \tiny{(CBM2023)}      & 73.7 & 65.0 & 71.5 & 79.6 & 86.0 & 4.6 \\
     M2SNet \tiny{(arxiv2023)}      & 80.4 & 72.5 & 78.7 & 83.0 & 91.2 & 4.1 \\
     PVT-GCASCADE \tiny{(WACV2024)} & \textit{82.0} & \textit{73.6} & \textit{80.2} & 83.7 & \textit{92.1} & 3.8 \\
     CFATransUNet \tiny{(CBM2024)}  & 80.6 & 72.8 & 79.5 & 83.1 & 91.1 & 4.1 \\
     MADGNet \tiny{(CVPR2024)}      & 81.3 & 73.4 & 79.5 & \textit{83.8} & 91.7 & \textbf{\underline{3.6}} \\
     \hline
     \multicolumn{1}{c|}{\multirow{2}{*}{\textbf{TransGUNet \tiny{(Ours)}}}}     & \textbf{\underline{82.7}} & \textbf{\underline{74.7}} & \textbf{\underline{81.0}} & \textbf{\underline{84.3}} & \textbf{\underline{93.0}} & \textit{3.7} \\ \cline{2-7}
     & \textbf{+1.0} & \textbf{+1.1} & \textbf{+0.8} & \textbf{+0.5} & \textbf{+0.9} & \textbf{-0.1} \\
    \hline
    \multicolumn{1}{c|}{\multirow{2}{*}{Method}} & \multicolumn{6}{c}{BUSI $\Rightarrow$ STU} \\ \cline{2-7}
     & DSC \scriptsize{$\uparrow$} & mIoU \scriptsize{$\uparrow$} & $F_{\beta}^{w}$ \scriptsize{$\uparrow$}  & $S_{\alpha}$ \scriptsize{$\uparrow$} & $E_{\phi}^{max}$ \scriptsize{$\uparrow$} & MAE \scriptsize{$\downarrow$} \\
     \hline
     UNet \tiny{(MICCAI2016)}       & 71.6 & 61.6 & 71.6 & 76.1 & 82.4 & 5.2 \\
     UNet++ \tiny{(DLMIA2018)}      & 77.0 & 68.0 & 76.3 & 79.8 & 86.7 & 4.4 \\
     CENet \tiny{(TMI2019)}         & 86.0 & 77.2 & 84.2 & 84.5 & 93.7 & 2.8 \\
     TransUNet \tiny{(arxiv2021)}   & 41.4 & 32.1 & 40.8 & 60.2 & 58.1 & 9.7 \\
     MSRFNet \tiny{(BHI2022)}       & 84.0 & 75.2 & 75.1 & 83.5 & 92.2 & 3.1 \\
     DCSAUNet \tiny{(CBM2023)}      & 86.1 & 76.5 & 82.7 & 84.9 & 94.7 & 3.2 \\
     M2SNet \tiny{(arxiv2023)}      & 79.4 & 69.3 & 76.4 & 81.3 & 90.7 & 4.3 \\
     PVT-GCASCADE \tiny{(WACV2024)} & 86.4 & 76.6 & 82.2 & 84.3 & 84.2 & 3.1 \\
     CFATransUNet \tiny{(CBM2024)}  & \textit{87.9} & \textit{79.2} & \textit{85.3} & \textit{85.7} & \textit{95.4} & \textbf{\underline{2.6}} \\
     MADGNet \tiny{(CVPR2024)}      & \textbf{\underline{88.4}} & \textbf{\underline{79.9}} & \textbf{\underline{86.4}} & \textbf{\underline{86.2}} & \textbf{\underline{95.9}} & \textbf{\underline{2.6}} \\
     \hline
     \multicolumn{1}{c|}{\multirow{2}{*}{\textbf{TransGUNet \tiny{(Ours)}}}}     & 87.4 & 78.2 & 84.1 & 85.4 & 94.9 & \textit{2.7} \\ \cline{2-7}
     & \textbf{-1.0} & \textbf{-1.7} & \textbf{-2.3} & \textbf{-0.8} & \textbf{-1.0} & \textbf{-0.1} \\
    \hline
    \end{tabular}
    \caption{Segmentation results on \textbf{Breast Tumor Segmentation}. We train each model on BUSI \cite{al2020dataset} train dataset and evaluate on BUSI \cite{al2020dataset} and STU \cite{zhuang2019rdau} test datasets.}
    \label{tab:comparison_sota_ultrasound_other_metrics}
\end{table}

\begin{table}[h]
    \centering
    \scriptsize
    \setlength\tabcolsep{2pt} 
    \renewcommand{\arraystretch}{0.75} 
    \begin{tabular}{c|cccccc}
    \hline
    \multicolumn{1}{c|}{\multirow{2}{*}{Method}} & \multicolumn{6}{c}{CVC-ClinicDB + Kvasir-SEG $\Rightarrow$ CVC-ClinicDB} \\ \cline{2-7}
     & DSC \scriptsize{$\uparrow$} & mIoU \scriptsize{$\uparrow$} & $F_{\beta}^{w}$ \scriptsize{$\uparrow$}  & $S_{\alpha}$ \scriptsize{$\uparrow$} & $E_{\phi}^{max}$ \scriptsize{$\uparrow$} & MAE \scriptsize{$\downarrow$} \\
     \hline
     UNet \tiny{(MICCAI2016)}       & 76.5 & 69.1 & 75.1 & 83.0 & 86.4 & 2.7 \\
     UNet++ \tiny{(DLMIA2018)}      & 79.7 & 73.6 & 79.4 & 85.1 & 88.3 & 2.2 \\
     CENet \tiny{(TMI2019)}         & 89.4 & 84.0 & 89.1 & 89.8 & 96.0 & 1.1 \\
     TransUNet \tiny{(arxiv2021)}   & 87.4 & 82.4 & 87.2 & 88.5 & 95.2 & 1.3 \\
     MSRFNet \tiny{(BHI2022)}       & 83.2 & 76.5 & 81.9 & 86.4 & 91.3 & 1.7 \\
     DCSAUNet \tiny{(CBM2023)}      & 80.6 & 73.7 & 79.6 & 84.9 & 89.9 & 2.4 \\
     M2SNet \tiny{(arxiv2023)}      & \textit{92.8} & \textit{88.2} & \textit{92.3} & 91.4 & \textit{97.7} & \textbf{\underline{0.7}} \\
     PVT-GCASCADE \tiny{(WACV2024)} & 92.2 & 87.6 & 91.6 & \textit{91.8} & 97.0 & \textit{0.8} \\
     CFATransUNet \tiny{(CBM2024)}  & 91.0 & 86.2 & 90.8 & 90.7 & 97.0 & \textit{0.8} \\
     MADGNet \tiny{(CVPR2024)}      & \textbf{\underline{93.9}} & \textbf{\underline{89.5}} & \textbf{\underline{93.6}} & \textbf{\underline{92.2}} & \textbf{\underline{98.5}} & \textbf{\underline{0.7}} \\
     \hline
     \multicolumn{1}{c|}{\multirow{2}{*}{\textbf{TransGUNet \tiny{(Ours)}}}}     & 92.3 & 87.7 & 91.8 & 91.6 & 97.1 & \textbf{\underline{0.7}} \\ \cline{2-7}
     & \textbf{-1.6} & \textbf{-1.8} & \textbf{-1.8} & \textbf{-0.6} & \textbf{-1.4} & \textbf{+0.0} \\
    \hline
    \multicolumn{1}{c|}{\multirow{2}{*}{Method}} & \multicolumn{6}{c}{CVC-ClinicDB + Kvasir-SEG $\Rightarrow$ Kvasir-SEG} \\ \cline{2-7}
    & DSC \scriptsize{$\uparrow$} & mIoU \scriptsize{$\uparrow$} & $F_{\beta}^{w}$ \scriptsize{$\uparrow$}  & $S_{\alpha}$ \scriptsize{$\uparrow$} & $E_{\phi}^{max}$ \scriptsize{$\uparrow$} & MAE \scriptsize{$\downarrow$} \\
    \hline
     UNet \tiny{(MICCAI2016)}       & 80.5 & 72.6 & 78.2 & 79.9 & 88.2 & 5.2 \\
     UNet++ \tiny{(DLMIA2018)}      & 84.3 & 77.4 & 83.1 & 82.1 & 90.5 & 4.6 \\
     CENet \tiny{(TMI2019)}         & 89.5 & 83.9 & 88.9 & 85.3 & 94.1 & 3.0 \\
     TransUNet \tiny{(arxiv2021)}   & 86.4 & 80.1 & 85.4 & 83.0 & 92.1 & 4.0 \\
     MSRFNet \tiny{(BHI2022)}       & 86.1 & 79.3 & 84.9 & 82.8 & 92.0 & 4.0 \\
     DCSAUNet \tiny{(CBM2023)}      & 82.6 & 75.2 & 80.7 & 81.3 & 90.1 & 4.9 \\
     M2SNet \tiny{(arxiv2023)}      & 90.2 & 85.1 & 89.4 & 85.6 & 94.6 & 2.7 \\
     PVT-GCASCADE \tiny{(WACV2024)} & 91.6 & 86.8 & 91.0 & 86.4 & \textit{96.3} & 2.4 \\
     CFATransUNet \tiny{(CBM2024)}  & \textit{92.1} & \textit{87.2} & \textit{91.7} & \textit{86.6} & 96.0 & \textit{2.2} \\
     MADGNet \tiny{(CVPR2024)}      & 90.7 & 85.3 & 89.9 & 85.6 & 94.7 & 3.1 \\
     \hline
     \multicolumn{1}{c|}{\multirow{2}{*}{\textbf{TransGUNet \tiny{(Ours)}}}}     & \textbf{\underline{93.1}} & \textbf{\underline{88.4}} & \textbf{\underline{92.6}} & \textbf{\underline{87.1}} & \textbf{\underline{96.6}} & \textbf{\underline{2.0}} \\ \cline{2-7}
     & \textbf{+1.0} & \textbf{+1.2} & \textbf{+0.9} & \textbf{+0.5} & \textbf{+0.3} & \textbf{+0.2} \\
    \hline
    \multicolumn{1}{c|}{\multirow{2}{*}{Method}} & \multicolumn{6}{c}{CVC-ClinicDB + Kvasir-SEG $\Rightarrow$ CVC-300} \\ \cline{2-7}
    & DSC \scriptsize{$\uparrow$} & mIoU \scriptsize{$\uparrow$} & $F_{\beta}^{w}$ \scriptsize{$\uparrow$}  & $S_{\alpha}$ \scriptsize{$\uparrow$} & $E_{\phi}^{max}$ \scriptsize{$\uparrow$} & MAE \scriptsize{$\downarrow$} \\
    \hline
     UNet \tiny{(MICCAI2016)}       & 66.1 & 58.5 & 65.0 & 79.7 & 80.0 & 1.7 \\
     UNet++ \tiny{(DLMIA2018)}      & 64.4 & 58.4 & 63.7 & 79.5 & 77.4 & 1.8 \\
     CENet \tiny{(TMI2019)}         & 85.4 & 78.2 & 84.2 & 90.2 & 94.0 & 0.8 \\
     TransUNet \tiny{(arxiv2021)}   & 85.0 & 77.3 & 83.1 & 89.4 & 95.2 & 1.1 \\
     MSRFNet \tiny{(BHI2022)}       & 72.3 & 65.4 & 71.2 & 83.5 & 84.6 & 1.4 \\
     DCSAUNet \tiny{(CBM2023)}      & 68.9 & 59.8 & 66.3 & 81.1 & 83.8 & 2.0 \\
     M2SNet \tiny{(arxiv2023)}      & \textit{89.8} & \textbf{\underline{83.2}} & \textbf{\underline{88.3}} & \textit{93.0} & \textbf{\underline{97.0}} & \textbf{\underline{0.6}} \\
     PVT-GCASCADE \tiny{(WACV2024)} & 88.2 & 81.0 & 85.9 & 92.0 & 95.6 & 0.9 \\
     CFATransUNet \tiny{(CBM2024)}  & 89.1 & 82.4 & 87.4 & 92.5 & 96.6 & \textit{0.7} \\
     MADGNet \tiny{(CVPR2024)}      & 87.4 & 79.9 & 84.5 & 92.0 & 94.7 & 0.9 \\
     \hline
     \multicolumn{1}{c|}{\multirow{2}{*}{\textbf{TransGUNet \tiny{(Ours)}}}}     & \textbf{\underline{90.0}} & \textit{83.1} & \textit{88.0} & \textbf{\underline{93.2}} & \textit{96.8} & \textit{0.7} \\ \cline{2-7}
     & \textbf{+0.2} & \textbf{-0.1} & \textbf{-0.3} & \textbf{+0.2} & \textbf{+0.2} & \textbf{-0.1} \\
    \hline
    \multicolumn{1}{c|}{\multirow{2}{*}{Method}} & \multicolumn{6}{c}{CVC-ClinicDB + Kvasir-SEG $\Rightarrow$ CVC-ColonDB} \\ \cline{2-7}
    & DSC \scriptsize{$\uparrow$} & mIoU \scriptsize{$\uparrow$} & $F_{\beta}^{w}$ \scriptsize{$\uparrow$}  & $S_{\alpha}$ \scriptsize{$\uparrow$} & $E_{\phi}^{max}$ \scriptsize{$\uparrow$} & MAE \scriptsize{$\downarrow$} \\
    \hline
     UNet \tiny{(MICCAI2016)}       & 56.8 & 49.0 & 55.9 & 72.6 & 73.9 & 5.1 \\
     UNet++ \tiny{(DLMIA2018)}      & 57.5 & 50.2 & 56.6 & 73.3 & 73.9 & 5.0 \\
     CENet \tiny{(TMI2019)}         & 65.9 & 59.2 & 65.8 & 77.7 & 79.5 & 4.0 \\
     TransUNet \tiny{(arxiv2021)}   & 63.7 & 58.4 & 62.8 & 75.8 & 79.3 & 4.8 \\
     MSRFNet \tiny{(BHI2022)}       & 61.5 & 54.8 & 60.8 & 75.4 & 76.1 & 4.5 \\
     DCSAUNet \tiny{(CBM2023)}      & 57.8 & 49.3 & 54.9 & 73.3 & 76.0 & 5.8 \\
     M2SNet \tiny{(arxiv2023)}      & 75.8 & 68.5 & 73.7 & 84.2 & 86.9 & 3.8 \\
     PVT-GCASCADE \tiny{(WACV2024)} & \textit{79.5} & \textit{71.6} & \textit{77.7} & \textit{84.5} & \textit{89.6} & 3.8 \\
     CFATransUNet \tiny{(CBM2024)}  & 78.0 & 70.3 & 77.3 & 83.7 & 88.7 & 3.5 \\
     MADGNet \tiny{(CVPR2024)}      & 77.5 & 69.7 & 76.2 & 83.3 & 88.0 & \textit{3.2} \\
     \hline
     \multicolumn{1}{c|}{\multirow{2}{*}{\textbf{TransGUNet \tiny{(Ours)}}}}     & \textbf{\underline{82.0}} & \textbf{\underline{74.1}} & \textbf{\underline{80.5}} & \textbf{\underline{85.6}} & \textbf{\underline{92.0}} & \textbf{\underline{3.0}} \\ \cline{2-7}
     & \textbf{+2.5} & \textbf{+2.5} & \textbf{+2.8} & \textbf{+1.1} & \textbf{+2.4} & \textbf{+0.2} \\
    \hline
    \multicolumn{1}{c|}{\multirow{2}{*}{Method}} & \multicolumn{6}{c}{CVC-ClinicDB + Kvasir-SEG $\Rightarrow$ ETIS} \\ \cline{2-7}
    & DSC \scriptsize{$\uparrow$} & mIoU \scriptsize{$\uparrow$} & $F_{\beta}^{w}$ \scriptsize{$\uparrow$}  & $S_{\alpha}$ \scriptsize{$\uparrow$} & $E_{\phi}^{max}$ \scriptsize{$\uparrow$} & MAE \scriptsize{$\downarrow$} \\
    \hline
     UNet \tiny{(MICCAI2016)}       & 41.6 & 35.4 & 39.5 & 67.2 & 61.7 & 2.7 \\
     UNet++ \tiny{(DLMIA2018)}      & 39.1 & 34.0 & 38.3 & 65.8 & 59.3 & 2.7 \\
     CENet \tiny{(TMI2019)}         & 57.0 & 51.4 & 56.0 & 74.9 & 73.8 & 2.2 \\
     TransUNet \tiny{(arxiv2021)}   & 50.1 & 44.0 & 48.8 & 70.7 & 68.7 & 2.6 \\
     MSRFNet \tiny{(BHI2022)}       & 38.3 & 33.7 & 36.9 & 66.0 & 58.4 & 3.6 \\
     DCSAUNet \tiny{(CBM2023)}      & 43.0 & 36.1 & 40.5 & 67.9 & 69.3 & 4.1 \\
     M2SNet \tiny{(arxiv2023)}      & 74.9 & 67.8 & 71.2 & 84.6 & 87.2 & 1.7 \\
     PVT-GCASCADE \tiny{(WACV2024)} & \textit{79.5} & \textit{71.6} & \textit{76.6} & \textit{86.3} & \textit{90.0} & 1.7 \\
     CFATransUNet \tiny{(CBM2024)}  & 77.0 & 69.5 & 75.0 & 84.4 & 88.6 & \textit{1.5} \\
     MADGNet \tiny{(CVPR2024)}      & 77.0 & 69.7 & 75.3 & 84.6 & 88.4 & 1.6 \\
     \hline
     \multicolumn{1}{c|}{\multirow{2}{*}{\textbf{TransGUNet \tiny{(Ours)}}}}     & \textbf{\underline{81.3}} & \textbf{\underline{73.1}} & \textbf{\underline{76.8}} & \textbf{\underline{87.6}} & \textbf{\underline{91.5}} & \textbf{\underline{1.4}} \\ \cline{2-7}
     & \textbf{+1.8} & \textbf{+1.5} & \textbf{+0.2} & \textbf{+1.3} & \textbf{+1.5} & \textbf{+0.1} \\
     \hline
    \end{tabular}
    \caption{Segmentation results on \textbf{Polyp Segmentation}. We train each model on CVC-ClinicDB \cite{bernal2015wm} + Kvasir-SEG \cite{jha2020kvasir} train dataset and evaluate on CVC-ClinicDB \cite{bernal2015wm}, Kvasir-SEG \cite{jha2020kvasir}, CVC-300 \cite{vazquez2017benchmark}, CVC-ColonDB \cite{tajbakhsh2015automated}, and ETIS \cite{silva2014toward} test datasets.}
    \label{tab:comparison_sota_colonoscopy_other_metrics}
\end{table}

\begin{table*}[t]
    \centering
    \scriptsize
    \setlength\tabcolsep{9pt} 
    \renewcommand{\arraystretch}{0.6} 
    \begin{tabular}{c|c|cccccccc}
    \hline
    \multicolumn{1}{c|}{\multirow{2}{*}{Method}} & \multicolumn{9}{c}{Synapse $\Rightarrow$ Synapse} \\ \cline{2-10}
     & Average & Aorta & Gallbladder & Left Kidney & Right Kidney & Liver & Pancreas & Spleen & Stomach \\
    \hline
    UNet \tiny{(MICCAI2016)}     & 69.8 & 84.3 & 44.1 & 73.0 & 71.9 & 91.8 & 46.0 & 79.2 & 68.0 \\
    UNet++ \tiny{(DLMIA2018)}    & 79.3 & 87.4 & 64.8 & 81.2 & 77.9 & 94.3 & 60.8 & \textbf{\underline{89.6}} & 78.7 \\
    CENet \tiny{(TMI2019)}       & 75.2 & 81.3 & 55.1 & 80.3 & 77.1 & 94.0 & 47.8 & 87.2 & 78.4 \\
    TransUNet \tiny{(arxiv2021)} & 77.5 & \textbf{\underline{87.2}} & 63.1 & 81.9 & 77.0 & 94.1 & 55.9 & 85.1 & 75.6 \\
    MSRFNet \tiny{(BHI2022)}     & 77.2 & 87.6 & 58.3 & 82.8 & 73.6 & 94.6 & 57.3 & 88.3 & 75.4 \\
    DCSAUNet \tiny{(CBM2023)}    & 71.0 & 81.4 & 51.9 & 75.1 & 68.8 & 92.7 & 45.4 & 84.6 & 68.0 \\
    M2SNet \tiny{(arxiv2023)}    & 77.1 & 84.9 & 55.6 & 78.2 & 76.1 & 94.9 & 57.4 & 89.1 & 80.3 \\
    PVT-GCASCADE \tiny{(WACV2024)} & 78.1 & 85.1 & 57.0 & 82.2 & 79.3 & 94.9 & 55.2 & 88.7 & 82.7 \\
    CFATransUNet \tiny{(CBM2024)} & \textit{80.5} & 85.8 & \textbf{\underline{65.9}} & \textit{85.8} & \textit{81.7} & \textit{95.2} & 59.4 & 89.0 & 81.4 \\
    MADGNet \tiny{(CVPR2024)}    & 79.3 & 85.2 & 59.9 & 85.4 & 78.0 & 94.5 & \textbf{\underline{60.5}} & 88.2 & \textit{83.0} \\
    \hline
    \multicolumn{1}{c|}{\multirow{2}{*}{\textbf{TransGUNet \tiny{(Ours)}}}} & \textbf{\underline{80.9}} & \textit{86.6} & \textit{61.0} & \textbf{\underline{87.0}} & \textbf{\underline{83.8}} & \textbf{\underline{95.3}} & \textit{59.7} & \textit{89.4} & \textbf{\underline{84.6}} \\ \cline{2-10}
    & \textbf{+0.4} & \textbf{-0.6} & \textbf{-4.9} & \textbf{+1.2} & \textbf{+2.1} & \textbf{+0.1} & \textbf{-0.8} & \textbf{-0.2} & \textbf{+1.6} \\
    \hline
    \multicolumn{1}{c|}{\multirow{2}{*}{Method}} & \multicolumn{9}{c}{Synapse $\Rightarrow$ AMOS-CT} \\ \cline{2-10}
     & Average & Aorta & Gallbladder & Left Kidney & Right Kidney & Liver & Pancreas & Spleen & Stomach \\
    \hline
    UNet \tiny{(MICCAI2016)}     & 56.3 & 62.8 & 43.2 & 50.3 & 55.3 & 84.6 & 26.8 & 68.9 & 58.8 \\
    UNet++ \tiny{(DLMIA2018)}    & 67.5 & 67.0 & 62.7 & 65.2 & 65.9 & 88.2 & 42.3 & 78.1 & 70.4 \\
    CENet \tiny{(TMI2019)}       & 67.9 & 72.4 & 52.0 & 69.5 & 72.3 & 89.3 & 40.8 & 78.9 & 68.2 \\
    TransUNet \tiny{(arxiv2021)} & 68.3 & 75.4 & 60.4 & 65.3 & 68.5 & 90.3 & 38.2 & 78.6 & 69.8 \\
    MSRFNet \tiny{(BHI2022)}     & 61.8 & 70.7 & 54.1 & 57.8 & 54.6 & 87.4 & 31.7 & 75.8 & 62.1 \\
    DCSAUNet \tiny{(CBM2023)}    & 45.7 & 43.4 & 30.6 & 32.5 & 39.5 & 85.1 & 18.7 & 67.6 & 48.0 \\
    M2SNet \tiny{(arxiv2023)}    & 69.6 & 74.4 & 60.9 & 66.7 & 69.5 & 90.8 & 41.8 & 79.6 & 73.3 \\
    PVT-GCASCADE \tiny{(WACV2024)} & 69.3 & 70.5 & 56.6 & 64.6 & 71.9 & \textit{91.7} & 42.0 & 82.0 & 75.4 \\
    CFATransUNet \tiny{(CBM2024)} & 68.0 & 75.0 & 51.8 & 67.5 & 74.5 & 88.2 & 43.6 & 78.2 & 65.5 \\
    MADGNet \tiny{(CVPR2024)}    & \textit{74.9} & \textit{79.4} & \textit{63.3} & \textit{77.6} & \textit{75.6} & 90.7 & \textbf{\underline{51.9}} & \textit{83.2} & \textit{77.3} \\
    \hline
    \multicolumn{1}{c|}{\multirow{2}{*}{\textbf{TransGUNet \tiny{(Ours)}}}} & \textbf{\underline{76.5}} & \textbf{\underline{81.0}} & \textbf{\underline{63.4}} & \textbf{\underline{79.1}} & \textbf{\underline{82.3}} & \textbf{\underline{91.9}} & \textit{51.4} & \textbf{\underline{84.7}} & \textbf{\underline{78.3}} \\ \cline{2-10}
    & \textbf{+1.6} & \textbf{+1.6} & \textbf{+0.1} & \textbf{+1.5} & \textbf{+6.7} & \textbf{+0.2} & \textbf{-0.5} & \textbf{+1.5} & \textbf{+1.0} \\
    \hline
    \multicolumn{1}{c|}{\multirow{2}{*}{Method}} & \multicolumn{9}{c}{Synapse $\Rightarrow$ AMOS-MRI} \\ \cline{2-10}
     & Average & Aorta & Gallbladder & Left Kidney & Right Kidney & Liver & Pancreas & Spleen & Stomach \\
    \hline
    UNet \tiny{(MICCAI2016)}     & 8.3 & 6.1 & 5.6 & 3.0 & 14.0 & 26.1 & 1.2 & 4.9 & 5.1 \\
    UNet++ \tiny{(DLMIA2018)}    & 6.0 & 9.7 & 6.7 & 7.9 & 2.7 & 7.1 & 1.5 & 4.6 & 7.6 \\
    CENet \tiny{(TMI2019)}       & 14.5 & 13.4 & 6.2 & 21.8 & 34.7 & 14.6 & \textit{5.7} & 9.5 & 9.8 \\
    TransUNet \tiny{(arxiv2021)} & 9.1 & 10.2 & 8.0 & 18.9 & 21.1 & 7.5 & 1.9 & 2.7 & 2.8 \\
    MSRFNet \tiny{(BHI2022)}     & 6.5 & 2.6 & 6.9 & 3.7 & 3.0 & 26.9 & 1.4 & 2.4 & 4.9 \\
    DCSAUNet \tiny{(CBM2023)}    & 1.7 & 0.9 & 1.7 & 0.0 & 0.8 & 7.8 & 0.0 & 1.8 & 0.4 \\
    M2SNet \tiny{(arxiv2023)}    & 22.0 & 29.1 & 8.4 & 35.0 & 34.5 & 29.0 & 5.1 & 18.5 & 16.7 \\
    PVT-GCASCADE \tiny{(WACV2024)} & 32.8 & 26.8 & 11.6 & \textit{44.1} & \textit{47.8} & \textit{64.3} & 3.9 & 38.3 & 25.5 \\
    CFATransUNet \tiny{(CBM2024)} & \textit{35.8} & \textit{30.0} & \textit{16.9} & 41.8 & 47.7 & 58.6 & 5.4 & \textit{57.0} & \textit{29.3} \\
    MADGNet \tiny{(CVPR2024)}    & 14.8 & 19.5 & 6.5 & 32.3 & 16.0 & 11.5 & 2.3 & 13.4 & 17.2 \\
    \hline
    \multicolumn{1}{c|}{\multirow{2}{*}{\textbf{TransGUNet \tiny{(Ours)}}}} & \textbf{\underline{47.2}} & \textbf{\underline{52.9}} & \textbf{\underline{20.5}} & \textbf{\underline{51.5}} & \textbf{\underline{59.9}} & \textbf{\underline{73.3}} & \textbf{\underline{17.0}} & \textbf{\underline{63.7}} & \textbf{\underline{38.8}} \\ \cline{2-10}
    & \textbf{+11.4} & \textbf{+22.9} & \textbf{+3.9} & \textbf{+7.4} & \textbf{+12.1} & \textbf{+9.0} & \textbf{+11.3} & \textbf{+6.7} & \textbf{+9.5} \\
    \hline
    \end{tabular}
    \caption{Segmentation results on \textbf{Multi-organ Segmentation} with \textit{DSC}. We train each model on Synapse \cite{Synapse_dataset} train dataset and evaluate on Synapse \cite{Synapse_dataset} and AMOS-CT/MRI \cite{ji2022amos} test datasets.}
    \label{tab:comparison_sota_multiorgan_dsc_metrics}
\end{table*}

\begin{table*}[t]
    \centering
    \scriptsize
    \setlength\tabcolsep{9pt} 
    \renewcommand{\arraystretch}{0.6} 
    \begin{tabular}{c|c|cccccccc}
    \hline
    \multicolumn{1}{c|}{\multirow{2}{*}{Method}} & \multicolumn{9}{c}{Synapse $\Rightarrow$ Synapse} \\ \cline{2-10}
     & Average & Aorta & Gallbladder & Left Kidney & Right Kidney & Liver & Pancreas & Spleen & Stomach \\
    \hline
    UNet \tiny{(MICCAI2016)}     & 58.9 & 73.1 & 35.3 & 62.5 & 60.8 & 85.1 & 32.4 & 68.2 & 54.1 \\
    UNet++ \tiny{(DLMIA2018)}    & 69.8 & \textbf{\underline{78.0}} & \textbf{\underline{53.3}} & 72.6 & 69.2 & 89.5 & \textit{46.4} & \textbf{\underline{82.7}} & 67.0 \\
    CENet \tiny{(TMI2019)}       & 64.6 & 68.9 & 43.4 & 71.1 & 67.2 & 88.9 & 33.1 & 78.4 & 65.9 \\
    TransUNet \tiny{(arxiv2021)} & 67.2 & 71.8 & 46.3 & 73.2 & 67.6 & 89.4 & \textbf{\underline{48.0}} & 80.3 & 60.7 \\
    MSRFNet \tiny{(BHI2022)}     & 67.6 & 78.1 & 47.7 & 74.1 & 63.8 & 89.9 & 43.2 & 80.9 & 62.8 \\
    DCSAUNet \tiny{(CBM2023)}    & 59.8 & 68.8 & 41.8 & 64.4 & 57.4 & 86.5 & 32.2 & 73.9 & 53.2 \\
    M2SNet \tiny{(arxiv2023)}    & 67.5 & 74.1 & 45.5 & 70.4 & 67.9 & 90.3 & 42.0 & 81.2 & 68.7 \\
    PVT-GCASCADE \tiny{(WACV2024)} & 68.9 & 75.0 & 47.5 & 74.0 & 70.1 & 90.4 & 39.9 & 80.5 & 73.5 \\
    CFATransUNet \tiny{(CBM2024)} & \textit{70.4} & 76.2 & 47.7 & 75.4 & \textit{73.0} & \textit{90.9} & \textbf{\underline{48.0}} & 80.8 & 71.3 \\
    MADGNet \tiny{(CVPR2024)}    & 69.8 & 73.0 & \textit{48.7} & \textit{75.8} & 71.9 & 90.4 & 45.7 & 80.2 & \textit{72.9} \\
    \hline
    \multicolumn{1}{c|}{\multirow{2}{*}{\textbf{TransGUNet \tiny{(Ours)}}}} & \textbf{\underline{71.4}} & \textit{76.4} & \textit{48.7} & \textbf{\underline{79.4}} & \textbf{\underline{76.0}} & \textbf{\underline{91.1}} & 43.9 & \textit{81.6} & \textbf{\underline{74.0}} \\ \cline{2-10}
    & \textbf{+0.9} & \textbf{-1.6} & \textbf{-4.6} & \textbf{+3.6} & \textbf{+3.0} & \textbf{+0.2} & \textbf{-4.1} & \textbf{-1.1} & \textbf{+1.1} \\
    \hline
    \multicolumn{1}{c|}{\multirow{2}{*}{Method}} & \multicolumn{9}{c}{Synapse $\Rightarrow$ AMOS-CT} \\ \cline{2-10}
     & Average & Aorta & Gallbladder & Left Kidney & Right Kidney & Liver & Pancreas & Spleen & Stomach \\
    \hline
    UNet \tiny{(MICCAI2016)}     & 44.8 & 49.3 & 33.2 & 39.0 & 42.3 & 74.9 & 17.2 & 57.0 & 45.5 \\
    UNet++ \tiny{(DLMIA2018)}    & 56.6 & 54.4 & \textit{51.9} & 54.4 & 54.7 & 80.2 & 30.5 & 68.3 & 58.1 \\
    CENet \tiny{(TMI2019)}       & 56.5 & 59.7 & 40.0 & 58.5 & 61.1 & 81.9 & 27.3 & 68.7 & 54.7 \\
    TransUNet \tiny{(arxiv2021)} & 57.7 & 63.6 & 49.0 & 55.0 & 57.4 & 83.3 & 26.9 & 69.2 & 57.5 \\
    MSRFNet \tiny{(BHI2022)}     & 51.3 & 58.4 & 43.7 & 47.3 & 44.2 & 79.1 & 22.5 & 65.4 & 49.7 \\
    DCSAUNet \tiny{(CBM2023)}    & 36.3 & 32.4 & 22.9 & 25.0 & 31.4 & 75.4 & 11.8 & 56.4 & 34.9 \\
    M2SNet \tiny{(arxiv2023)}    & 58.5 & 62.0 & 49.0 & 55.7 & 58.2 & 84.2 & 28.6 & 69.9 & 60.6 \\
    PVT-GCASCADE \tiny{(WACV2024)} & 58.5 & 57.5 & 45.0 & 54.3 & 60.9 & \textit{85.6} & 28.9 & 72.8 & 62.9 \\
    CFATransUNet \tiny{(CBM2024)} & 56.7 & 62.4 & 40.9 & 56.0 & 63.5 & 80.2 & 30.4 & 67.7 & 52.5 \\
    MADGNet \tiny{(CVPR2024)}    & \textit{64.4} & \textit{68.2} & \textbf{\underline{52.0}} & \textit{68.3} & \textit{65.5} & 84.1 & \textbf{\underline{37.6}} & \textit{74.2} & \textit{65.6} \\
    \hline
    \multicolumn{1}{c|}{\multirow{2}{*}{\textbf{TransGUNet \tiny{(Ours)}}}} & \textbf{\underline{66.2}} & \textbf{\underline{69.8}} & 50.6 & \textbf{\underline{70.1}} & \textbf{\underline{73.3}} & \textbf{\underline{86.0}} & \textit{36.8} & \textbf{\underline{76.1}} & \textbf{\underline{66.5}} \\ \cline{2-10}
    & \textbf{+1.7} & \textbf{+1.6} & \textbf{-1.4} & \textbf{+1.8} & \textbf{+7.8} & \textbf{+0.4} & \textbf{-0.8} & \textbf{+1.9} & \textbf{+0.9} \\
    \hline
    \multicolumn{1}{c|}{\multirow{2}{*}{Method}} & \multicolumn{9}{c}{Synapse $\Rightarrow$ AMOS-MRI} \\ \cline{2-10}
     & Average & Aorta & Gallbladder & Left Kidney & Right Kidney & Liver & Pancreas & Spleen & Stomach \\
    \hline
    UNet \tiny{(MICCAI2016)}     & 5.3 & 3.5 & 5.3 & 1.7 & 8.8 & 16.6 & 0.6 & 3.2 & 2.8 \\
    UNet++ \tiny{(DLMIA2018)}    & 3.8 & 5.9 & 6.0 & 5.0 & 1.6 & 3.8 & 0.8 & 2.9 & 4.2 \\
    CENet \tiny{(TMI2019)}       & 9.0 & 8.1 & 5.6 & 14.1 & 22.3 & 8.0 & \textit{3.1} & 5.5 & 5.3 \\
    TransUNet \tiny{(arxiv2021)} & 5.8 & 5.9 & 6.9 & 12.2 & 13.3 & 4.1 & 1.0 & 1.8 & 1.5 \\
    MSRFNet \tiny{(BHI2022)}     & 4.2 & 2.6 & 6.9 & 3.7 & 3.0 & 8.5 & 1.4 & 2.4 & 4.9 \\
    DCSAUNet \tiny{(CBM2023)}    & 1.1 & 0.5 & 1.7 & 0.0 & 0.4 & 4.8 & 0.0 & 1.0 & 0.2 \\
    M2SNet \tiny{(arxiv2023)}    & 14.7 & 20.2 & 7.0 & 24.7 & 24.1 & 17.8 & 2.8 & 11.4 & 9.8 \\
    PVT-GCASCADE \tiny{(WACV2024)} & 24.3 & 17.4 & 9.4 & \textit{33.6} & 35.8 & \textit{50.4} & 2.2 & 29.6 & 15.9 \\
    CFATransUNet \tiny{(CBM2024)} & \textit{25.9} & \textit{21.1} & \textit{13.1} & 31.0 & \textit{36.1} & 42.6 & 2.9 & \textit{41.9} & \textit{18.5} \\
    MADGNet \tiny{(CVPR2024)}    & 9.8 & 12.5 & 5.8 & 22.1 & 10.3 & 6.5 & 1.2 & 9.5 & 10.2 \\
    \hline
    \multicolumn{1}{c|}{\multirow{2}{*}{\textbf{TransGUNet \tiny{(Ours)}}}} & \textbf{\underline{35.6}} & \textbf{\underline{39.0}} & \textbf{\underline{14.7}} & \textbf{\underline{40.5}} & \textbf{\underline{46.1}} & \textbf{\underline{59.5}} & \textbf{\underline{9.5}} & \textbf{\underline{49.3}} & \textbf{\underline{25.8}} \\ \cline{2-10}
    & \textbf{+9.6} & \textbf{+17.9} & \textbf{+1.6} & \textbf{+6.9} & \textbf{+10.0} & \textbf{+9.1} & \textbf{+6.4} & \textbf{+7.4} & \textbf{+7.3} \\
    \hline
    \end{tabular}
    \caption{Segmentation results on \textbf{Multi-organ Segmentation} with \textit{mIoU}. We train each model on Synapse \cite{Synapse_dataset} train dataset and evaluate on Synapse \cite{Synapse_dataset} and AMOS-CT/MRI \cite{ji2022amos} test datasets.}
    \label{tab:comparison_sota_multiorgan_mIoU_metrics}
\end{table*}

\begin{table*}
    \centering
    \scriptsize
    \setlength\tabcolsep{9pt} 
    \renewcommand{\arraystretch}{0.6} 
    \begin{tabular}{c|c|cccccccc}
    \hline
    \multicolumn{1}{c|}{\multirow{2}{*}{Method}} & \multicolumn{9}{c}{Synapse $\Rightarrow$ Synapse} \\ \cline{2-10}
     & Average & Aorta & Gallbladder & Left Kidney & Right Kidney & Liver & Pancreas & Spleen & Stomach \\
    \hline
    UNet \tiny{(MICCAI2016)}     & 43.8 & 15.1 & 46.8 & 45.6 & 90.6 & 28.3 & 21.5 & 76.6 & 26.2 \\
    UNet++ \tiny{(DLMIA2018)}    & 34.9 & 4.3 & 52.5 & 51.2 & 62.2 & 20.4 & 11.4 & 61.5 & 15.6 \\
    CENet \tiny{(TMI2019)}       & 39.8 & 13.5 & 86.5 & 53.8 & 86.8 & 16.4 & 13.8 & 21.6 & 25.9 \\
    TransUNet \tiny{(arxiv2021)} & 34.3 & 10.7 & 53.8 & 37.0 & 76.5 & \textit{15.0} & \textbf{\underline{8.1}} & 55.3 & 17.7 \\
    MSRFNet \tiny{(BHI2022)}     & 38.2 & 8.8 & 56.1 & 59.5 & 66.1 & 26.8 & 13.3 & 36.2 & 38.7 \\
    DCSAUNet \tiny{(CBM2023)}    & 38.6 & 10.4 & 80.0 & 54.9 & 49.4 & 26.0 & 15.7 & 52.4 & 20.2 \\
    M2SNet \tiny{(arxiv2023)}    & 25.4 & \textit{5.7} & \textit{27.7} & 49.6 & 48.0 & 22.4 & 10.9 & \textit{25.7} & 13.5 \\
    PVT-GCASCADE \tiny{(WACV2024)} & \textbf{\underline{18.8}} & \textbf{\underline{5.6}} & \textbf{\underline{20.2}} & 25.5 & \textit{36.0} & 17.8 & 12.9 & \textit{20.8} & \textit{11.6} \\
    CFATransUNet \tiny{(CBM2024)} & \textbf{\underline{18.8}} & \textbf{\underline{5.6}} & 29.0 & \textit{16.0} & \textbf{\underline{34.6}} & 23.1 & 10.8 & \textbf{\underline{20.1}} & \textbf{\underline{10.8}} \\
    MADGNet \tiny{(CVPR2024)}    & 24.9 & \textit{5.7} & 29.8 & 40.4 & 44.1 & \textbf{\underline{14.0}} & 13.2 & 35.0 & 17.0 \\
    \hline
    \multicolumn{1}{c|}{\multirow{2}{*}{\textbf{TransGUNet \tiny{(Ours)}}}} & \textit{23.4} & 7.1 & 33.6 & \textbf{\underline{14.6}} & 48.3 & 22.4 & \textit{10.5} & 39.8 & \textbf{\underline{10.8}} \\ \cline{2-10}
    & \textbf{-4.6} & \textbf{-1.5} & \textbf{-13.4} & \textbf{+1.4} & \textbf{-13.7} & \textbf{-8.4} & \textbf{-2.4} & \textbf{-19.7} & \textbf{+0.0} \\
    \hline
    \multicolumn{1}{c|}{\multirow{2}{*}{Method}} & \multicolumn{9}{c}{Synapse $\Rightarrow$ AMOS-CT} \\ \cline{2-10}
     & Average & Aorta & Gallbladder & Left Kidney & Right Kidney & Liver & Pancreas & Spleen & Stomach \\
    \hline
    UNet \tiny{(MICCAI2016)}     & 72.1 & 39.5 & 83.5 & 94.1 & 132.4 & 36.8 & 24.5 & 129.7 & 36.5 \\
    UNet++ \tiny{(DLMIA2018)}    & 43.9 & 19.9 & 55.0 & 73.4 & \textit{45.7} & 25.5 & \textit{16.9} & 90.5 & 24.0 \\
    CENet \tiny{(TMI2019)}       & 52.2 & 13.5 & 97.5 & 53.3 & 106.2 & 24.1 & 27.0 & 55.5 & 40.1 \\
    TransUNet \tiny{(arxiv2021)} & 39.5 & 15.4 & 46.2 & \textit{46.1} & 102.3 & 19.6 & 21.6 & 40.1 & 25.0 \\
    MSRFNet \tiny{(BHI2022)}     & 47.0 & 20.2 & 31.8 & 96.6 & 49.2 & 22.6 & 26.3 & 90.9 & 38.4 \\
    DCSAUNet \tiny{(CBM2023)}    & 50.5 & 28.7 & 92.1 & 63.1 & 82.9 & 23.4 & 34.6 & 44.7 & 34.5 \\
    M2SNet \tiny{(arxiv2023)}    & 41.2 & \textbf{\underline{10.0}} & \textbf{\underline{30.8}} & 73.7 & 72.9 & 22.0 & 20.3 & 69.8 & 29.7 \\
    PVT-GCASCADE \tiny{(WACV2024)} & \textit{27.5} & 12.8 & 35.9 & 46.5 & \textbf{\underline{40.2}} & \textit{13.9} & 21.9 & \textit{27.7} & \textit{20.9} \\
    CFATransUNet \tiny{(CBM2024)} & 46.3 & 14.9 & 81.0 & 84.3 & 54.1 & 27.1 & 24.7 & 59.0 & 24.9 \\
    MADGNet \tiny{(CVPR2024)}    & 31.2 & \textbf{\underline{10.0}} & 34.7 & 58.5 & 55.7 & 16.8 & 18.6 & 29.2 & 26.2 \\
    \hline
    \multicolumn{1}{c|}{\multirow{2}{*}{\textbf{TransGUNet \tiny{(Ours)}}}} & \textbf{\underline{24.4}} & \textit{12.5} & \textit{31.1} & \textbf{\underline{29.6}} & 46.3 & \textbf{\underline{13.5}} & \textbf{\underline{15.7}} & \textbf{\underline{27.1}} & \textbf{\underline{19.1}} \\ \cline{2-10}
    & \textbf{+1.8} & \textbf{-2.5} & \textbf{-0.3} & \textbf{+16.5} & \textbf{-6.1} & \textbf{+0.4} & \textbf{+1.2} & \textbf{+0.6} & \textbf{+1.8} \\
    \hline
    \multicolumn{1}{c|}{\multirow{2}{*}{Method}} & \multicolumn{9}{c}{Synapse $\Rightarrow$ AMOS-MRI} \\ \cline{2-10}
     & Average & Aorta & Gallbladder & Left Kidney & Right Kidney & Liver & Pancreas & Spleen & Stomach \\
    \hline
    UNet \tiny{(MICCAI2016)}     & 190.3 & 144.3 & 198.4 & 143.0 & 187.3 & 198.4 & 105.9 & 322.0 & 223.2 \\
    UNet++ \tiny{(DLMIA2018)}    & 169.3 & 68.9 & 248.7 & \textit{109.0} & 134.2 & 207.7 & 113.0 & 264.1 & 209.0 \\
    CENet \tiny{(TMI2019)}       & 182.8 & 95.9 & 242.6 & 171.4 & 143.1 & 216.3 & 157.0 & 267.1 & 168.9 \\
    TransUNet \tiny{(arxiv2021)} & 162.3 & 100.6 & 256.8 & 114.3 & 160.9 & 199.3 & 116.4 & 237.4 & 112.8 \\
    MSRFNet \tiny{(BHI2022)}     & 166.3 & 102.4 & 166.1 & 144.0 & 137.0 & 178.6 & 136.9 & 210.9 & 254.4 \\
    DCSAUNet \tiny{(CBM2023)}    & 147.2 & 124.1 & \textbf{\underline{108.7}} & 230.1 & 183.4 & 97.1 & 86.4 & 194.8 & 152.6 \\
    M2SNet \tiny{(arxiv2023)}    & 131.7 & \textit{55.4} & 148.8 & 130.7 & 136.9 & 153.0 & 90.0 & 203.3 & 135.6 \\
    PVT-GCASCADE \tiny{(WACV2024)} & \textit{110.5} & 62.4 & \textit{115.4} & 127.4 & 141.0 & \textbf{\underline{91.0}} & 74.5 & \textit{150.0} & 122.1 \\
    CFATransUNet \tiny{(CBM2024)} & 115.6 & 69.2 & 146.2 & 156.3 & \textit{137.0} & 108.3 & \textbf{\underline{57.4}} & 171.3 & \textbf{\underline{79.2}} \\
    MADGNet \tiny{(CVPR2024)}    & 128.2 & 71.7 & 121.0 & 148.4 & 144.6 & 149.9 & 79.9 & 199.5 & 110.7 \\
    \hline
    \multicolumn{1}{c|}{\multirow{2}{*}{\textbf{TransGUNet \tiny{(Ours)}}}} & \textbf{\underline{98.7}} & \textbf{\underline{45.2}} & 123.4 & \textbf{\underline{105.2}} & \textbf{\underline{136.8}} & \textit{91.2} & \textit{60.7} & \textbf{\underline{142.0}} & \textit{84.9} \\ \cline{2-10}
    & \textbf{+11.8} & \textbf{+10.2} & \textbf{-14.7} & \textbf{+3.8} & \textbf{+0.2} & \textbf{-0.2} & \textbf{-3.3} & \textbf{+8.0} & \textbf{-5.7} \\
    \hline
    \end{tabular}
    \caption{Segmentation results on \textbf{Multi-organ Segmentation} with \textit{HD95}. We train each model on Synapse \cite{Synapse_dataset} train dataset and evaluate on Synapse \cite{Synapse_dataset} and AMOS-CT/MRI \cite{ji2022amos} test datasets.}
    \label{tab:comparison_sota_multiorgan_HD95_metrics}
\end{table*}

\end{document}